%% file: main.tex
\documentclass[10pt,twocolumn,letterpaper]{article}

\usepackage[pagenumbers]{cvpr}   
\input{preamble}

\definecolor{cvprblue}{rgb}{0.21,0.49,0.74}
\usepackage[pagebackref,breaklinks,colorlinks,allcolors=cvprblue]{hyperref}

\title{Structure From Tracking: \\Distilling Structure-Preserving Motion for Video Generation}

\def\@fnsymbol#1{\ensuremath{\ifcase#1\or *\or \dagger\or \ddagger\or
   \mathsection\or \mathparagraph\or \|\or **\or \dagger\dagger
   \or \ddagger\ddagger \else\@ctrerr\fi}}

\author{
Yang Fei$^{1,2\ddagger}$ \quad
George Stoica$^{2,3}$ \quad
Jingyuan Liu$^{4}$ \quad
Qifeng Chen$^{1\dagger}$ \quad \\
Ranjay Krishna$^{2*}$ \quad 
Xiaojuan Wang$^{2*}$ \quad
Benlin Liu$^{2*\dagger}$ \\
$^{1}$HKUST \quad
$^{2}$University of Washington \quad
$^{3}$Georgia Tech \quad
$^{4}$Adobe \\
}

\begin{document}

\maketitle

\begingroup
\renewcommand\thefootnote{\fnsymbol{footnote}}
\setcounter{footnote}{0}
\footnotetext[1]{equal advising.}
\footnotetext[2]{corresponding authors.}
\footnotetext[3]{This work was done in part while Yang Fei was an exchange student at the University of Washington.}
\endgroup

\input{sec/0_abstract}    
\input{sec/1_intro}
\input{sec/2_relat}
\input{sec/3_method}
\input{sec/4_exper}
\input{sec/5_conclusion}
{
    \small
    \bibliographystyle{ieeenat_fullname}
    \bibliography{main}
}
\input{sec/99_supp}

\end{document}

%% file: preamble.tex
 \newcommand{\model}{{\texttt{SAM2VideoX}}}

\newcommand{\KL}{\operatorname{KL}}
\newcommand{\softmax}{\operatorname{S}}
\newcommand{\LGF}{\operatorname{LGF}}



%% file: sec/0_abstract.tex
\begin{abstract}
Reality is a dance between rigid constraints and deformable structures. For video models, that means generating motion that preserves fidelity as well as \textit{structure}.
Despite progress in diffusion models, producing realistic structure-preserving motion remains challenging, especially for articulated and deformable objects such as humans and animals. 
Scaling training data alone, so far, has failed to resolve physically implausible transitions. 
Existing approaches rely on conditioning with noisy motion representations, such as optical flow or skeletons extracted using an external imperfect model.
To address these challenges, we introduce an algorithm to distill structure-preserving motion priors from an autoregressive video tracking model (SAM2) into a bidirectional video diffusion model (CogVideoX). 
With our method, we train \model{}, which contains two innovations: (1) a \textit{bidirectional feature fusion} module that extracts global structure-preserving motion priors from a recurrent model like SAM2; (2) a \textit{Local Gram Flow} loss that aligns how local features move together.
Experiments on VBench and in human studies show that \model{} delivers consistent gains (+2.60\% on VBench, 21–22\% lower FVD, and 71.4\% human preference) over prior baselines.
Specifically, on VBench, we achieve 95.51\%, surpassing REPA (92.91\%) by 2.60\%, and reduce FVD to 360.57, a 21.20\% and 22.46\% improvement over REPA- and LoRA-finetuning, respectively. The project website can be found at~\href{https://sam2videox.github.io/}{https://sam2videox.github.io/}.
\end{abstract}

%% file: sec/1_intro.tex
\section{Introduction}
\label{sec:intro}

\input{figures/teaser.tex}

From Heraclitus to Bergson, philosophy has cast cognition as the apprehension of \textit{becoming} rather than \textit{being}~\cite{heraclitus_fragments,bergson_creative_evolution}. 
While image generation models have excelled at generating what is \textit{there} in high-fidelity images~\cite{sora2024, blattmann2023stable, li2024hunyuan}, our best video generation still struggles to express the dynamics of \textit{change}.  
The central challenge is \emph{structure-preserving} motion: dynamics that maintain part topology and local neighborhoods while allowing constrained deformations.
Without these constraints, generated motions drift; limbs shear, textures tear, and object identity is lost.
Only by achieving this can models move beyond static appearance toward faithful world simulators.

Motion remains a major challenge. This is particularly true for articulated and highly deformable objects, such as humans and animals, where models often suffer from inconsistent or physically implausible transitions in object states. 
Inference-time interventions, such as ControlNet-style~\cite{zhang2023adding} conditioning on explicit motion representations during inference, require knowing the ideal motion apriori. 
Unlike rigid objects whose motion can be well represented by simple dragged trajectories~\citep{deng2024dragvideo}, articulated and deformable entities lack a unified motion representation
A common assumption is that low motion quality stems from insufficient training data, especially for high-quality complex articulated motions.
However, scaling up or augmenting training data only helps marginally; training with motion proxies (optical flow~\citep{chefer2025videojam}, skeletons~\citep{jeong2024track4gen}) for objects~\cite{sun2024ponymation, yang2024omnimotiongpt} still results in physically implausible transitions. Our experiments show that generated videos still produce lions walking without alternating legs and cyclists with static knees.
Scaling up such priors results in noisy training data; motion priors are usually collected using imperfect models (\eg RAFT can generate optical flow priors~\cite{teed2020raft}).

To improve articulated motion, we propose a simple but powerful idea: deriving \textit{structure from tracking}.
Our model, \model{}, distills \emph{structure-preserving} motion priors from a video tracking model into a video diffusion generator.
Previous work has shown that distilling image representations improves image generation fidelity~\cite{zhang2025videorepa}. 
To generalize this insight from static to dynamic generation, we distill video representations to improve video generation.
Specifically, we leverage SAM2~\citep{ravi2024sam}, a state-of-the-art video tracking model trained on large-scale, diverse video data. SAM2 is capable of maintaining object identity across long sequences and through complex occlusions.
To track, SAM2's internal representations have captured how parts move together, how limbs stay connected, and how occlusions resolve over time.
Instead of conditioning generation on explicit control signals like optical flow~\cite{chefer2025videojam} or skeletons~\cite{jeong2024track4gen}, \model{} extracts implicit structural cues directly from SAM2’s internal representations.
By transferring these motion priors, the generator acquires an internal sense of structure and continuity. 
In short, we leverage the structural understanding of a tracker to guide the generation of motion.

However, transferring useful information from SAM2 into a video generation model is technically challenging. We find that directly supervising diffusion models to predict SAM2 output masks yields only limited benefit, as the masks are discrete and boundary-focused; they fail to supervise useful fine-grained motion. 
Also, a direct alignment between feature spaces is hindered by an architectural asymmetry: state-of-the-art video generation models use DiT~\cite{peebles2023scalable} architecture with bidirectional attention to access global context, while SAM2 is inherently recurrent and causal. 
To bridge this gap, we extract a supervision signal by fusing \textit{forward} and \textit{backward} SAM2 features, where backward features are extracted by reversing the order of frames in a training video. 
Together, the forward and backward features represent a better global video context.
We align video diffusion features with this supervisory signal using a \emph{Local Gram Flow} loss.
Although $\ell_2$ loss has worked well for image generation~\cite{zhang2025videorepa}, we find that a local Gram loss captures better motion priors by emphasizing local relational structure.

Across qualitative and quantitative evaluations, \model{} yields more realistic, structurally coherent motion. 
On VBench~\cite{huang2023vbench} (matched dynamic degree) we achieve 95.51\%, surpassing REPA~\cite{yu2024representation} (92.91\%) by 2.60 points. Our FVD is 360.57, a reduction of 21.20\% and 22.46\% versus REPA- and LoRA-finetuning~\cite{hu2022lora}.
Since existing benchmarks are limited in assessing preservation of structure in articulated motion, we further conduct a human evaluation, where 71.4\% of ratings prefer our results.
Qualitatively, \model{} produces videos with the correct number of legs when animals walk, ensuring plausible human limb trajectories during complex activities, and producing accurate human-object interactions. 
These gains indicate that SAM2-guided distillation substantially strengthens structure-preserving motion in video diffusion models without sacrificing visual quality.

%% file: figures/teaser.tex
\begin{figure*}[t]
\centering
\setlength{\fboxsep}{0pt}   
\setlength{\fboxrule}{1pt}  
\def\imW{0.16\linewidth}

\begin{tabular}{c@{\hskip 0pt}c@{\hskip 0pt}c@{\hskip 0pt}c@{\hskip 5pt}c@{\hskip 0pt}c@{\hskip 0pt}c@{\hskip 0pt}}
    \rotatebox{90}{\small CogVideoX} &
    \includegraphics[width=\imW]{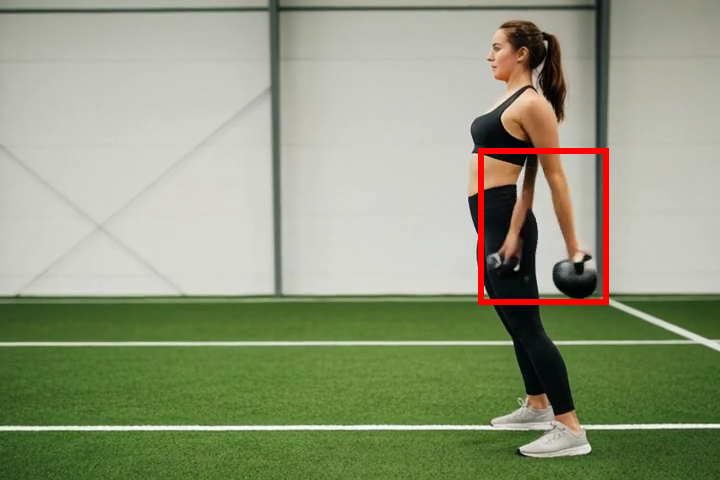} &
    \includegraphics[width=\imW]{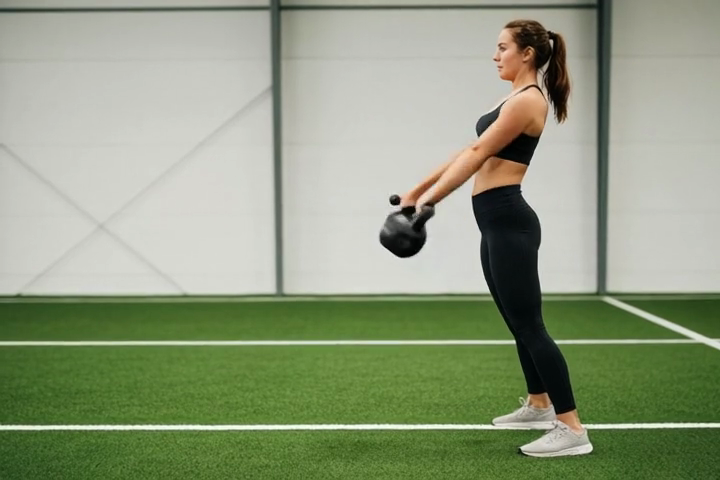} &
    \includegraphics[width=\imW]{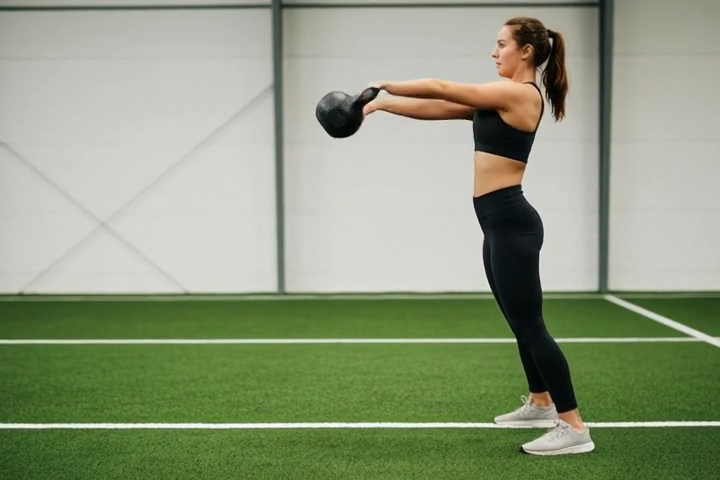} &
    
    \includegraphics[width=\imW]{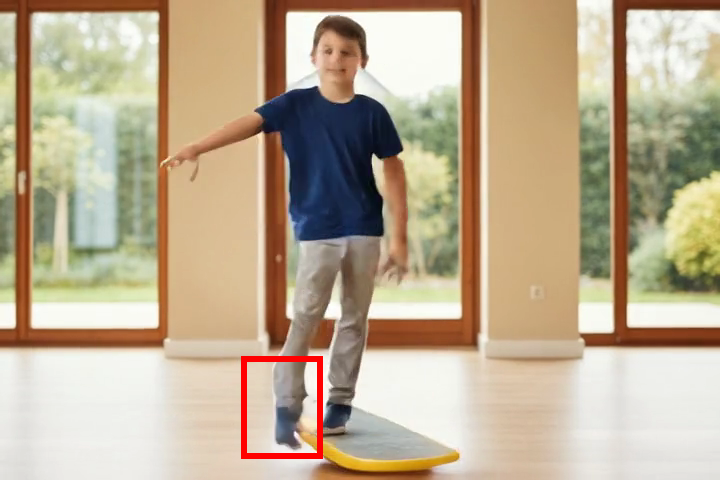} &
    \includegraphics[width=\imW]{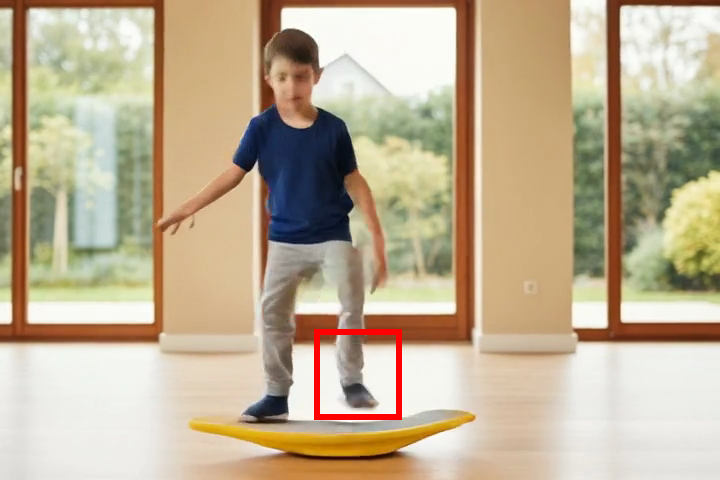} &
    \includegraphics[width=\imW]{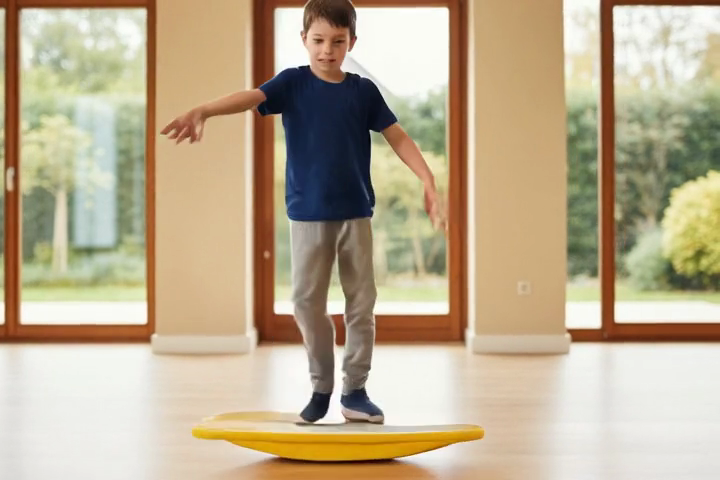} \\

    \rotatebox{90}{\small HunyuanVid} &
    \includegraphics[width=\imW]{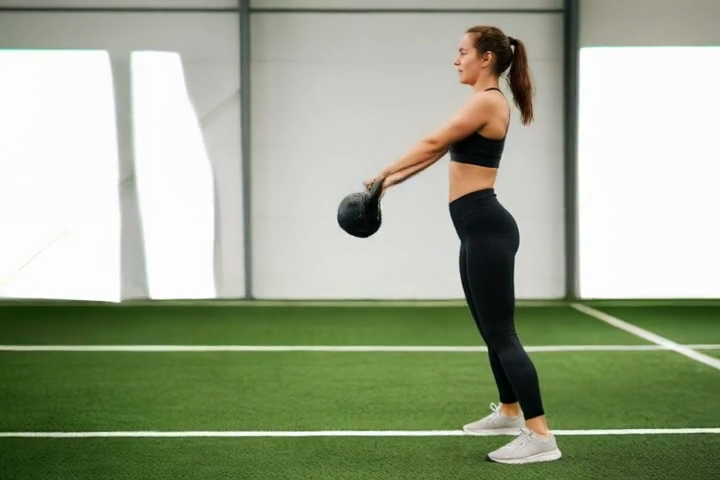} &
    \includegraphics[width=\imW]{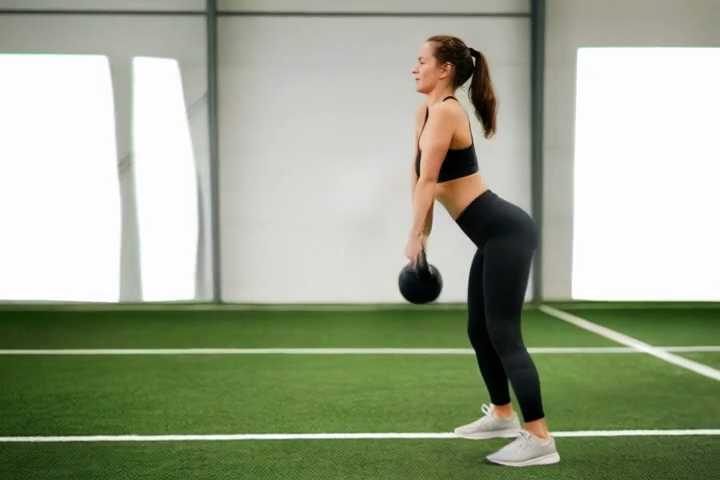} &
    \includegraphics[width=\imW]{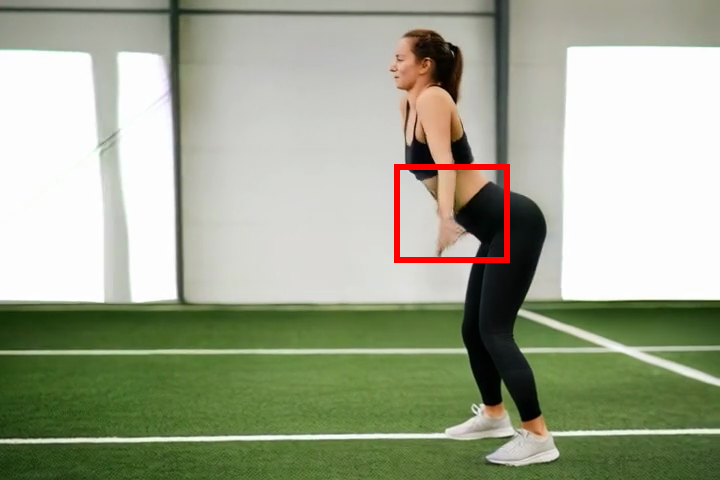} &

    \includegraphics[width=\imW]{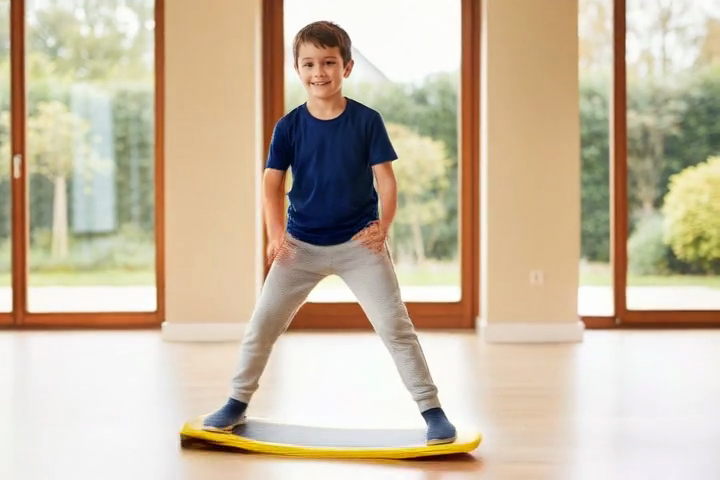} &
    \includegraphics[width=\imW]{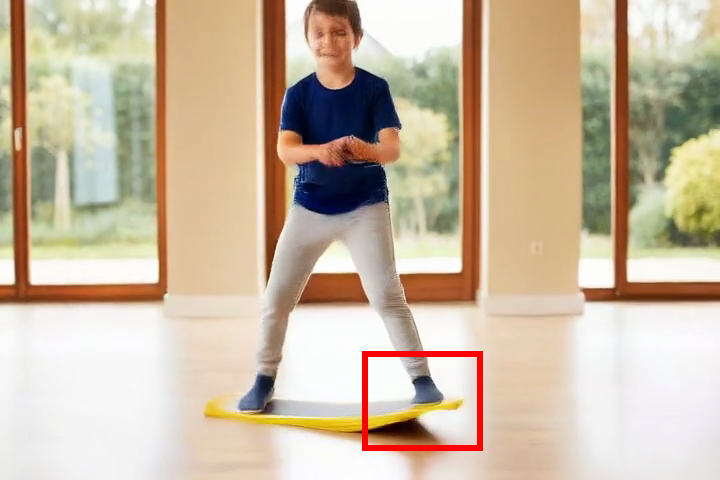} &
    \includegraphics[width=\imW]{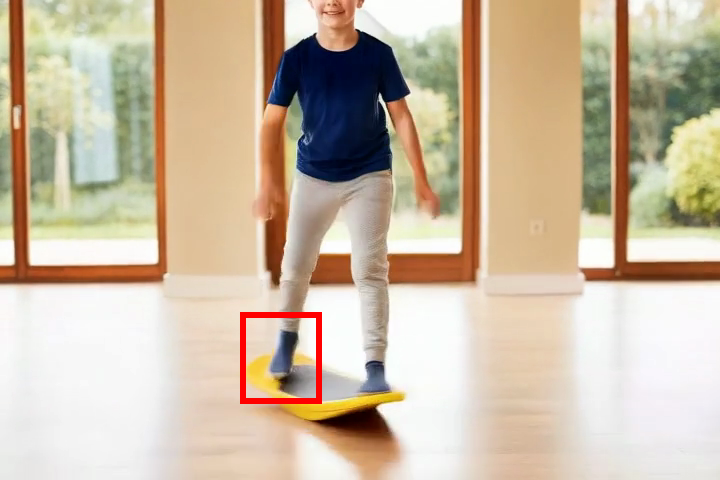} \\

    \rotatebox{90}{\small \textbf{\model{}}} &
    \includegraphics[width=\imW]{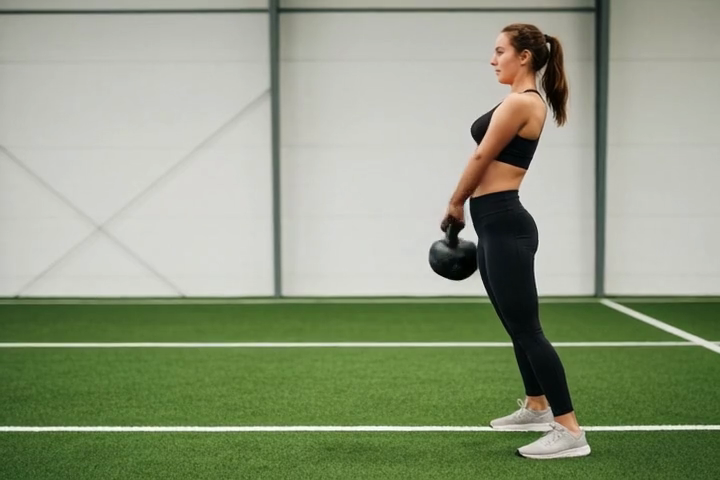} &
    \includegraphics[width=\imW]{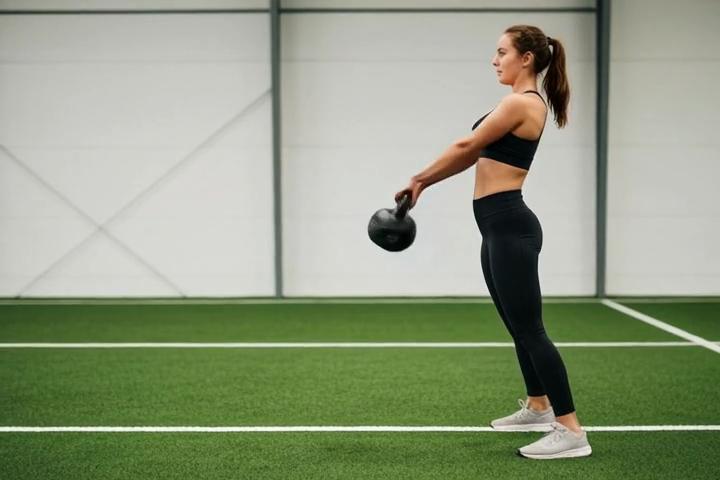} &
    \includegraphics[width=\imW]{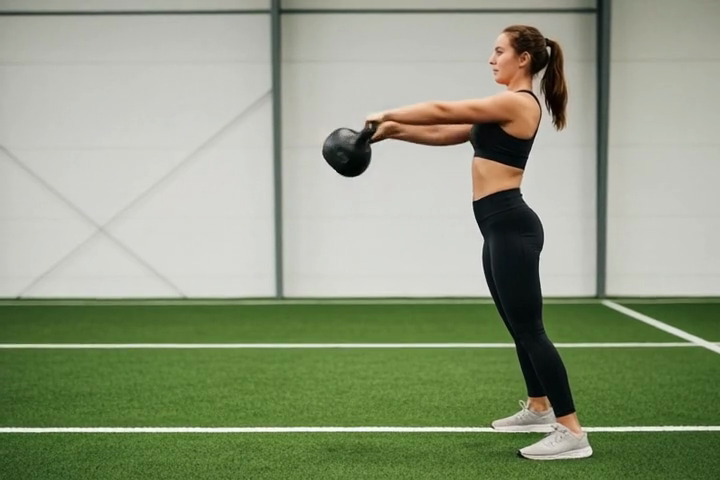} &
    
    \includegraphics[width=\imW]{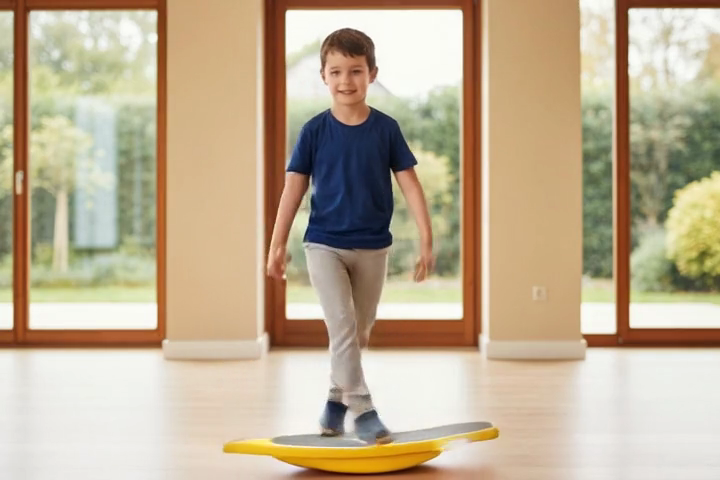} &
    \includegraphics[width=\imW]{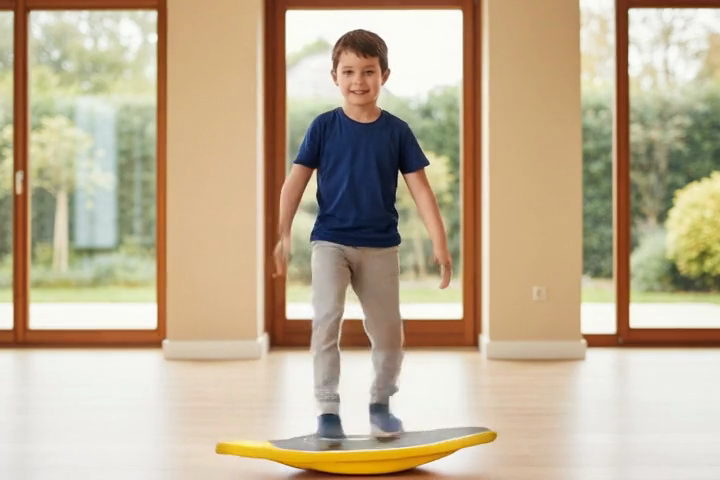} &
    \includegraphics[width=\imW]{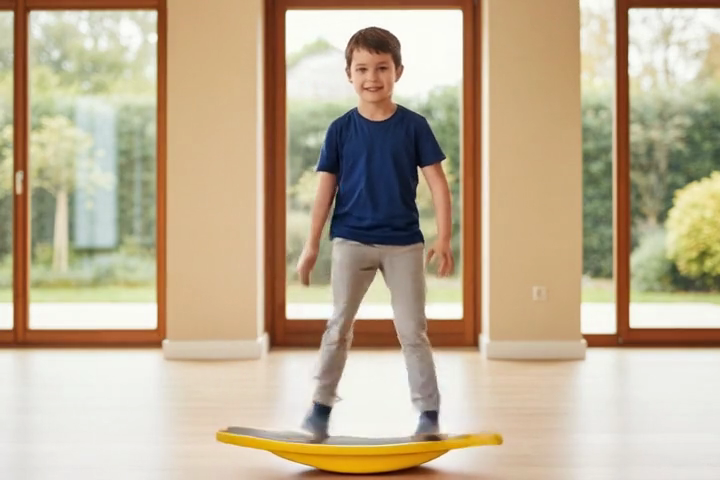} \\
    
\end{tabular}

\caption{
    We present a training algorithm to distill structure-preserving motion priors from SAM2 into a video diffusion model to improve motion accuracy and smoothness in generated videos. Compared to advanced image-to-video models CogVideoX~\cite{yang2024cogvideox} and HunyuanVid~\cite{li2024hunyuan}, our \model{} produces videos with superior or highly competitive fidelity, despite HunyuanVid having more than twice number of parameters (13B vs. our 5B).
}
\label{fig:teaser_comparison}

\end{figure*}

%% file: sec/2_relat.tex
\section{Related Work}
\label{sec:relat}

\noindent\textbf{Video diffusion models.}  
Video diffusion models have progressed rapidly, spanning both UNet-based~\citep{ronneberger2015u} architectures such as Stable Video Diffusion~\citep{blattmann2023stable}, and DiT-based~\citep{peebles2023scalable} models including Sora~\citep{sora2024}, CogVideoX~\citep{yang2024cogvideox}, HunyuanVid~\citep{li2024hunyuan}, OpenSora~\citep{opensora2}, Open-Sora-Plan~\citep{lin2024open}, Wan-Video~\citep{wan2025wan}, Cosmos~\citep{nvidia2025worldsimulationvideofoundation}. While these models can generate visually impressive videos, producing realistic and coherent structure-preserving motion remains challenging, especially for articulated and deformable entities. 
Given the difficulty of designing reliable inference-time control signals~\citep{zhang2023adding} for such objects, our goal is to enhance the base model’s intrinsic ability to generate structure-preserving motion without \emph{relying on auxiliary handcrafted motion controls during inference}.

\vspace{0.2cm}
\noindent\textbf{Motion understanding.}
Understanding motion has long been central to video analysis~\citep{horn1981determining,perazzi2016benchmark}. Point trajectories and optical flow~\citep{teed2020raft, karaev2024cotracker3} describe local, adjacent-frame changes and carry limited semantics; they often degrade under fast or long-range motion and struggle to maintain object identity through occlusions.
Mask-based tracking offers instance-level signals that are more stable in cluttered scenes.
SAM2~\citep{ravi2024sam} tracks user-prompted regions across long sequences and is known to generalize well across domains while preserving object identity through occlusions.
However, raw masks are discrete and boundary-focused, which discards much of the appearance and motion structure that video diffusion models need.
We therefore leverage SAM2’s internal features as motion priors: they are dense, continuous and temporally consistent, and they provide long-range correspondences that are useful for structure preservation.

\vspace{0.2cm}
\noindent\textbf{Representation alignment.}
Representation alignment was introduced for image generation by REPA~\citep{yu2024representation} and inspired several follow-ups.
In the video domain, two main approaches exist: aligning diffusion features to structured motion signals such as trajectories or flow~\citep{jeong2024track4gen, chefer2025videojam}; or to generic video encoders like VideoMAEv2~\citep{tong2022videomae}, as in VideoREPA~\citep{zhang2025videorepa}. Both suffer from key limitations. First, Trajectories and flow provide local supervision and are sensitive to long-range dynamics, which weakens their usefulness for structure preservation.
Second, popular video encoders like VideoMAEv2 are optimized for high-level semantic tasks rather than low-level motion understanding. In contrast, our approach aligns diffusion features to SAM2’s internal features.
This prior is unified and structure-centric, allowing transfer across humans and animals while yielding object-consistent motion representations without requiring a specific controller at inference.

%% file: sec/3_method.tex
\begin{figure*}[t]
    \centering
    \includegraphics[width=.8\linewidth]{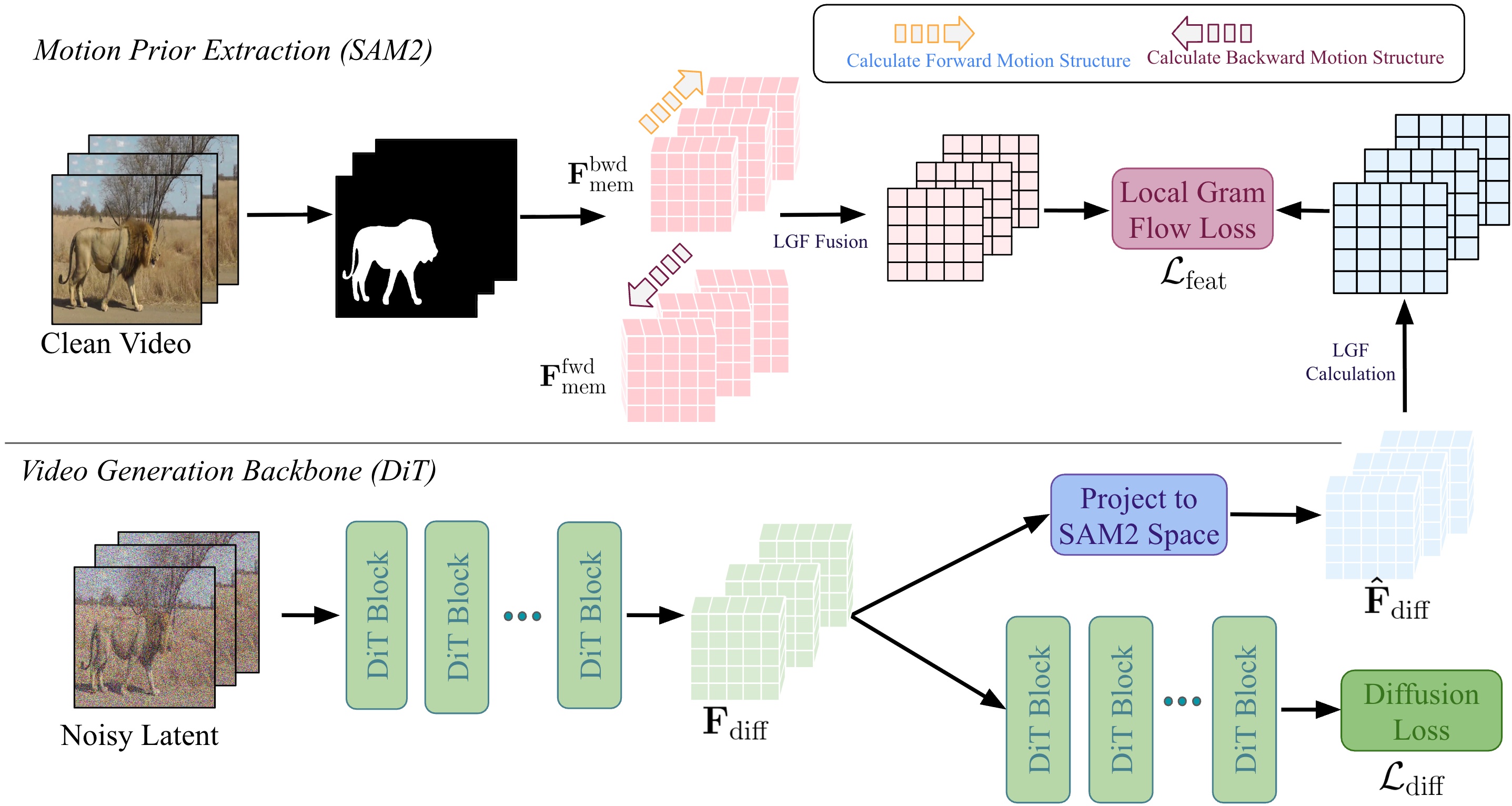}    
    \caption{\textbf{Method overview.} The framework consists of two parallel branches: \textbf{(Top)} The {\it Motion Prior Extraction} branch extracts forward and backward memory features ($\mathbf{F}^\mathrm{fwd}_\mathrm{mem}, \mathbf{F}^\mathrm{bwd}_\mathrm{mem}$) from SAM2 given a clean video, and fuses them into a bidirectional teacher representation. \textbf{(Bottom)} The {\it Video Generation Backbone} takes noisy latents as input, and the intermediate DiT features $\mathbf{F}_\mathrm{diff}$ are projected into the SAM2 space as $\mathbf{\hat F}_\mathrm{diff}$. Then the proposed \textbf{Local Gram Flow loss} ($\mathcal{L}_\mathrm{feat}$) is used to align the spatio-temporal structure of the projected student features with the teacher priors.}
    \label{fig:method_overview}
\end{figure*}

\section{Preliminaries}\label{sec:Preliminary}
In this work, we employ diffusion transformer (DiT)~\citep{peebles2023scalable} as our base video generation model, aiming to distill SAM2's motion understanding into the DiT to enhance its ability to learn structure-preserving motion from in-the-wild videos. We first introduce preliminary background on latent video diffusion models and the SAM2 architecture.

\vspace{0.2cm}
\noindent\textbf{Latent Video Diffusion Model.}
Diffusion models generate samples by inverting a forward noising process~\cite{song2019generative, song2020denoising, dhariwal2021diffusion}. In the latent setting, a pre-trained autoencoder, with encoder $\mathcal{E}(\cdot)$ and decoder $\mathcal{D}(\cdot)$, maps a video $\mathbf{x}=\{I_0,\dots, I_{N-1}\} \in \mathbb{R}^{N \times H \times W \times C}$ to a latent representation $\mathbf{z}=\mathcal{E}(\mathbf{x})$, where $z \in \mathbb{R}^{N'\ \times H' \times W' \times C'}$. At timestep $t$, noisy latent $\mathbf{z}_t$ is sampled as
\[
\mathbf{z}_t = \alpha_t \,\mathbf{z} + \sigma_t \,\boldsymbol{\epsilon}, \quad \boldsymbol{\epsilon}\sim\mathcal{N}(\mathbf{0},\mathbf{I}).
\]
A DiT $f_\theta$ is trained to predict the velocity target $\mathbf{v} = \alpha_t \boldsymbol{\epsilon} - \sigma_t \mathbf{z}$ under the standard v-prediction objective. At inference, iterative denoising maps $\mathbf{z}_T \!\to\! \mathbf{z}_0$, after which the decoder $\mathcal{D}(\mathbf{z}_0)$ produces the final video.

\vspace{0.2cm}
\noindent\textbf{Segment-Anything Model 2 (SAM2).}
The Segment-Anything Model 2 (SAM2) extends the image segmentation capability of SAM~1~\citep{kirillov2023segment} to the video domain, i.e., tracking an object in a video conditioned on given prompts such as points or boxes.

Given a video $\mathbf{x} = \{I_0, I_1, ..., I_{N-1}\}$, SAM2 processes the video recurrently and produces a segmentation mask for each frame. Specifically, an image encoder $\mathcal{I}$ extracts frame embedding $F$ from the current frame, which is then enhanced by a memory attention module $\mathcal{M}$ aggregating historical context from a memory bank $\mathcal{B}$ to produce $F_{\mathrm{mem}}$. A mask decoder $\mathcal{D}_{\text{mask}}$ subsequently takes $F_{\mathrm{mem}}$ and user prompts to generate the segmentation mask. By applying this recurrent procedure across all frames, SAM2 produces both a sequence of masks:
\[
\mathbf{M} = \{M_0, M_1, ..., M_{N-1}\}
\]
and a sequence of memory features from $\mathcal{M}$:
\[
\mathbf{F}_{\mathrm{mem}} = \{F_{\mathrm{mem}, 0}, F_{\mathrm{mem}, 1}, ..., F_{\mathrm{mem}, N-1}\}.
\]

\section{Method}
 We distill SAM2’s motion structure prior into the video DiT model by aligning their internal feature representations. This is achieved through a learnable feature alignment module (Sec.~\ref{sec:feat_align}) that projects intermediate DiT features into a latent space where they can be matched to those of SAM2. To align their relational motion structures,  we introduce a novel Local Gram Flow (LGF) feature matching operator (Sec.~\ref{sec:lgf}). Furthermore, due to the autoregressive nature of SAM2'feature, in contrast to the bidirectional features in DiT,  we propose a method that combines SAM2's forward and backward video features into a single bidirectional representation, providing a more suitable teacher signal to be distilled from (Sec.~\ref{sec:bi_sam}). Finally, a Local Gram Flow motion distillation loss (Sec.~\ref{sec:loss}) is applied to enforce this alignment in the latent space. Fig.~\ref{fig:method_overview} provides an overview of our full method.

 \subsection{Feature Alignment Network}\label{sec:feat_align}
 Formally, given a training video $\mathbf{x}$, we encode it to a latent representation $\mathbf{z} = \mathcal{E}(\mathbf{x})$ using the video VAE, then add noise at timestep $t$ to produce $\mathbf{z}_t$. The noised latent $\mathbf{z}_t$ and timestep $t$ are then fed into the denoising network $f_\theta$, from which we extract intermediate activations as video diffusion features $\mathbf{F}_{\mathrm{diff}} \in \mathbb{R}^{N' \times H' \times W' \times C'}$, where $N'$ is the number of latent frames.  Here $\mathbf{F}_{\mathrm{diff}}$ denotes the activations from a selected intermediate layer.
For the same video, we extract SAM2's internal features $\mathbf{F}_{\mathrm{SAM2}}$ as a distillation teacher. Compared with  the output segmentation masks, the internal feature representations provide richer spatio-temporal information. They capture object motion and part-level dynamics and can teach diffusion model internalize motion priors beyond simple boundary cues.

To align $\mathbf{F}_\mathrm{diff}$ and $\mathbf{F}_\mathrm{SAM2}$, We project $\mathbf{F}_{\mathrm{diff}}$ into SAM2's feature space. Specifically, we add a projection module $\mathcal{P}$ on top of $\mathbf{F}_{\mathrm{diff}}$, 
which consists of an interpolation layer with skip connections for temporal dimension matching, and then followed by a three-layer MLP, yielding:
\[
\mathbf{\hat{F}}_{\mathrm{diff}} = \mathcal{P}(\mathbf{F}_{\mathrm{diff}}).
\]  

We then compute a motion distillation loss between $\mathbf{\hat{F}}_{\mathrm{diff}}$ and $\mathbf{F}_{\mathrm{SAM2}}$. Our final objective combines the alignment loss with the standard diffusion loss:  
\[
\min_{f_\theta, \mathcal{P}} \; \mathcal{L}_{\mathrm{diff}} + \lambda\, \mathcal{L}_{\mathrm{feat}}(\mathbf{\hat{F}}_{\mathrm{diff}}, \mathbf{F}_{\mathrm{SAM2}}),
\]  
where $\mathcal{L}_{\mathrm{diff}}$ is the v-prediction loss and $\lambda=0.5$ balances the terms. 

\subsection{Local Gram Flow  Feature Matching}\label{sec:lgf}
\begin{figure}[t]
    \centering
    \includegraphics[width=1.0\linewidth]{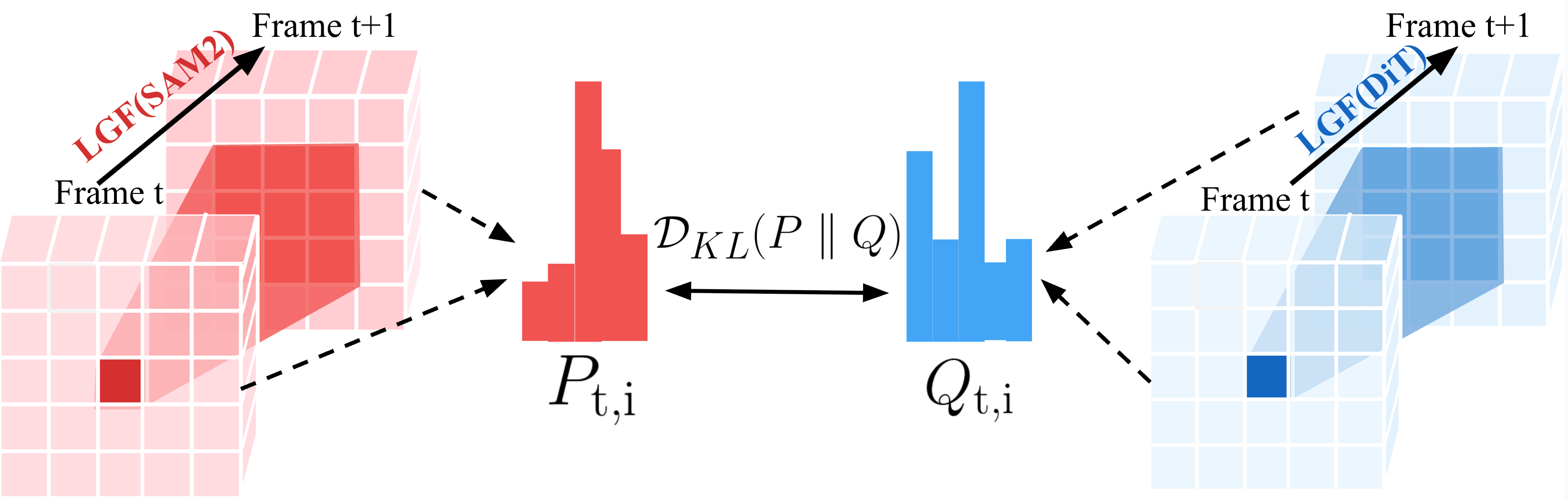}    
    
    \caption{\textbf{Illustration of the Local Gram Flow (LGF) Loss.} The LGF operator captures motion structure by computing similarities between a token at frame $t$ and its 7x7 spatial neighborhood in the subsequent frame $t+1$. Instead of matching absolute values (e.g., $\ell_2$), we convert the resulting similarity vectors ($P_\mathrm{t, i}, Q_\mathrm{t,i}$) into probability distributions and align them using the KL Divergence. This forces the model to learn relative motion patterns, not just feature values.}
    \label{fig:local_gram_flow}
\end{figure}

To capture cross-frame spatio-temporal motion structure, rather than performing direct one-to-one feature matching between $\mathbf{\hat{F}}_{\mathrm{diff}}$ and $\mathbf{F}_{\mathrm{SAM2}}$, we instead match their respective \emph{Gram matrices}, which encode pairwise dot products of token feature embeddings, \ie, pairwise token similarities. However, computing the full Gram matrices is computationally prohibitive for video diffusion models due to the large number of tokens. We therefore propose \emph{Local Gram Flow}, which computes the dot products only between each token and the tokens within its $7\times7$ spatial neighborhood in the subsequent frame (see Fig.~\ref{fig:local_gram_flow}). This yields local similarity vectors at each position that model likely motion trajectories to the next frame. We denote this operator as $\LGF(\cdot)$.

\subsection{Bidirectional Fusion of Causal SAM2 Features}\label{sec:bi_sam}
 A key challenge in aligning the feature representation of DiT and SAM2 is the architectural asymmetry between them. Video DiTs typically employ bidirectional attention that allows each token to attend to all frames, whereas SAM2’s recurrent mechanism constrains its memory feature $\mathbf{F}_{\mathrm{mem}}$ at each timestep to encode only current and past frames. To bridge this gap and create a teacher feature $\mathbf{F}_\mathrm{SAM2}$ that is aware of the full video context (similar to DiT), we construct it from both a forward and backward pass of SAM2.  The backward pass is obtained by feeding the temporally reversed video into SAM2 to extract its backward features. After obtaining the backward features, we remap them to the original temporal order (i.e., $t\mapsto N{-}1{-}t$) to align with the forward features, producing: 
\[
\mathbf{F}^{\mathrm{fwd}}_{\mathrm{mem}} = \{F^{\mathrm{fwd}}_{\mathrm{mem}, 0}, F^{\mathrm{fwd}}_{\mathrm{mem}, 1}, \ldots, F^{\mathrm{fwd}}_{\mathrm{mem}, N-1}\},
\]
\[
\mathbf{F}^{\mathrm{bwd}}_{\mathrm{mem}} = \{F^{\mathrm{bwd}}_{\mathrm{mem}, 0}, F^{\mathrm{bwd}}_{\mathrm{mem}, 1}, \ldots, F^{\mathrm{bwd}}_{\mathrm{mem}, N-1}\}.
\]

Empirically, using separate projectors to align separately to $\mathbf{F}_{\mathrm{mem}}^\mathrm{fwd}$ and $\mathbf{F}_{\mathrm{mem}}^\mathrm{bwd}$ provides marginal improvement, as gradient conflicts destabilize training. We therefore fuse them into a unified bidirectional teacher feature. 

However, fusing these two features is non-trivial. As our ablation study  demonstrate (Table~\ref{tab:ablations}), naively adding them ($k \mathbf{F}^{\mathrm{fwd}}_{\mathrm{mem}}+ (1-k)\mathbf{F}^{\mathrm{bwd}}_{\mathrm{mem}}$) leads to severe performance degradation. Because the final fused feature will be aligned with the projected DiT feature through their Local Gram Flows, we instead directly fuse their Local Gram Flows via a convex combination, which stabilizes training while preserving complementary information:
\[
\LGF(\mathbf{F}_\mathrm{SAM2}) =
\mathrm{k}\,\LGF(\mathbf{F}_{\mathrm{mem}}^{\mathrm{fwd}})
  + (1-\mathrm{k})\,\LGF(\mathbf{F}_{\mathrm{mem}}^{\mathrm{bwd}})
\\
\]

\subsection{Motion Distillation Loss}\label{sec:loss}
Finally, the motion distillation loss, $\mathcal{L}_\mathrm{feat}$, is designed to match the LGF distributions of the student $\mathbf{\hat F}_\mathrm{diff}$ and the fused teacher $\mathbf{F}_\mathrm{SAM2}$, rather than enforcing one-to-one correspondence as in a standard $\ell2$ loss.  More specifically, we align the distribution of spatio-temporal motion similarities between the video DiT feature and the SAM2 feature.  We therefore apply a softmax to each token's similarity vector (turning it into a probability distribution) and measure the distance using the KL divergence. This approach focuses on the relative ranking of similarities, which we argue better captures the underlying motion structure. Specifically, we compute probability distributions:
\[
\begin{aligned}
\mathbf{P} &= \softmax\!\Big(
  \LGF({\mathbf{F}}_{\mathrm{SAM2}})
\Big), \\
\mathbf{Q} &= \softmax\!\Big(
  \LGF({\mathbf{\hat F}}_{\mathrm{diff}})
\Big).
\end{aligned}
\]

The motion distillation loss averages the  KL divergence over all spatial tokens $\Omega$ and all $N'-1$ latent frames for which LGF is computed: 
\[
\mathcal{L}_{\mathrm{feat}}\!\left(\mathbf {\hat F}_{\mathrm{diff}},\,\mathbf F_{\mathrm{SAM2}}\right) =  \frac{1}{(N'-1)|\Omega|}\sum_{t=0}^{N'-2}\sum_{i \in \Omega} \KL\!\big( P_{\mathrm{t,i}} \,\big\|\, Q_\mathrm{t,i} \big)
\]

Our ablation studies (Table~\ref{tab:ablations}) empirically validate that this LGF-KL combination is critical, yielding significant gains over simpler alternatives (e.g., LGF with $\ell2$ loss, or direct feature matching).

%% file: sec/4_exper.tex
\section{Experiments}

\begin{figure*}[t]
\centering
\setlength{\fboxsep}{0pt}   
\setlength{\fboxrule}{1pt}  
\def\imW{0.18\linewidth}

\begin{tabular}{c@{\hskip 1pt}c@{\hskip 0pt}c@{\hskip 0pt}c@{\hskip 0pt}c@{\hskip 0pt}c}
    \rotatebox{90}{\small Base model} 
   &\fcolorbox{blue}{white}{\includegraphics[width=\imW]   {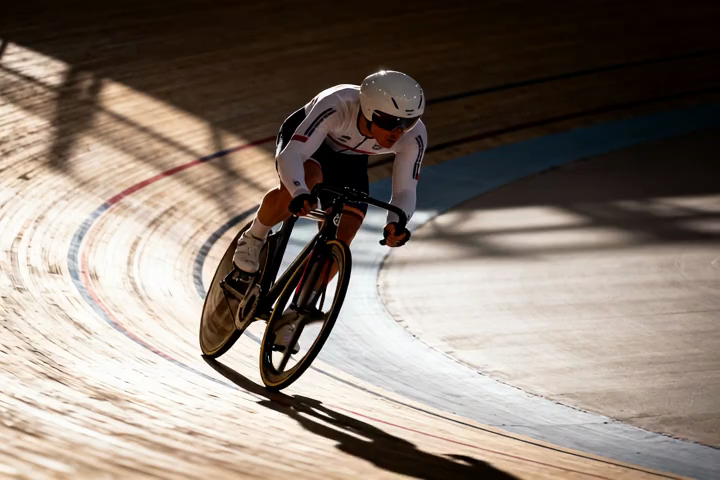}} &
    \includegraphics[width=\imW]{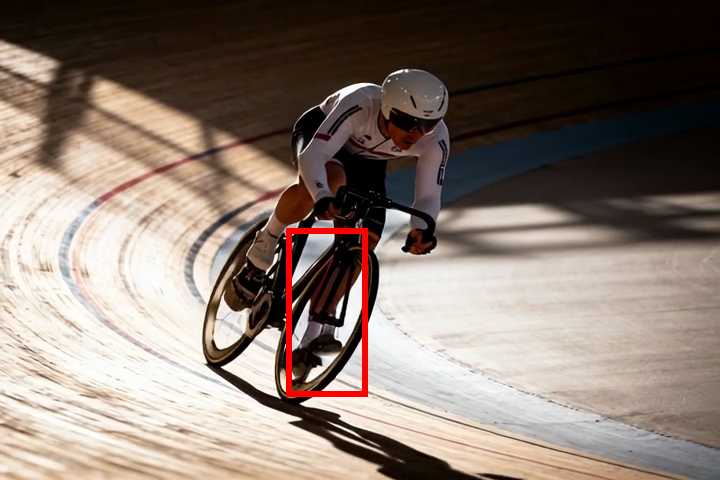} &
    \includegraphics[width=\imW]{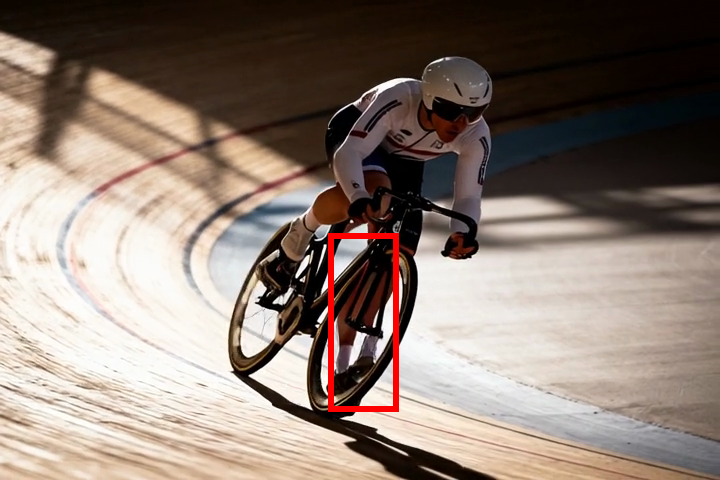} &
    \includegraphics[width=\imW]{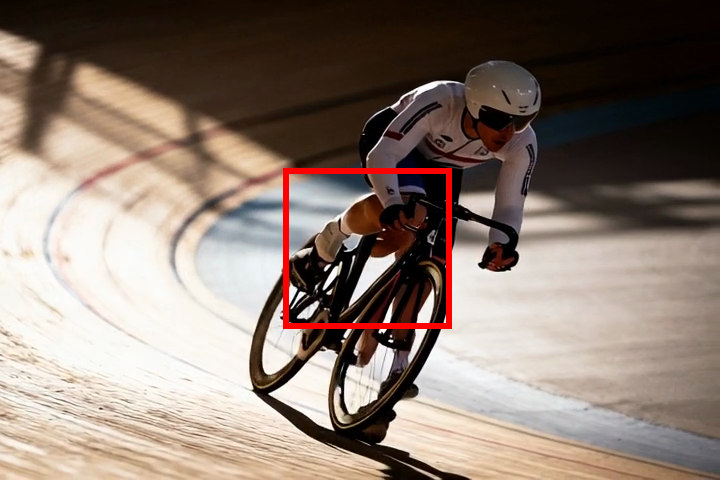} &
    \includegraphics[width=\imW]{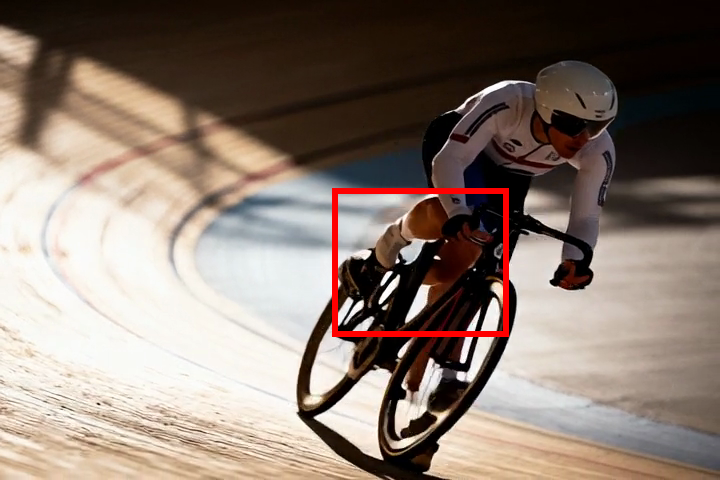} \\

    \rotatebox{90}{\small + Fine-tuning} 
   &\includegraphics[width=\imW]{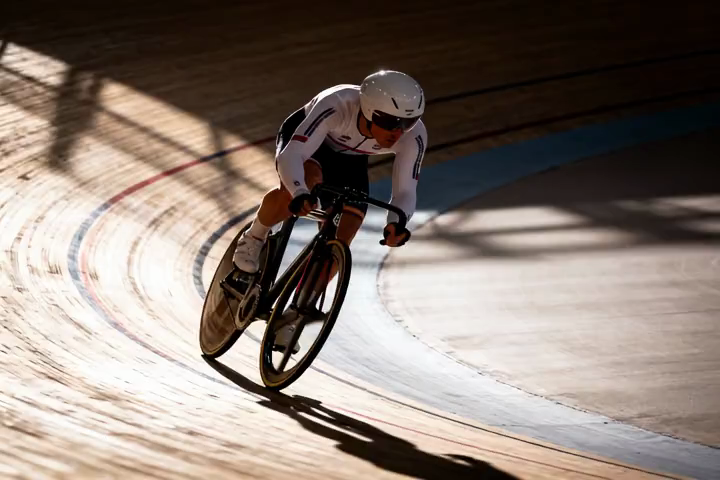} &
    \includegraphics[width=\imW]{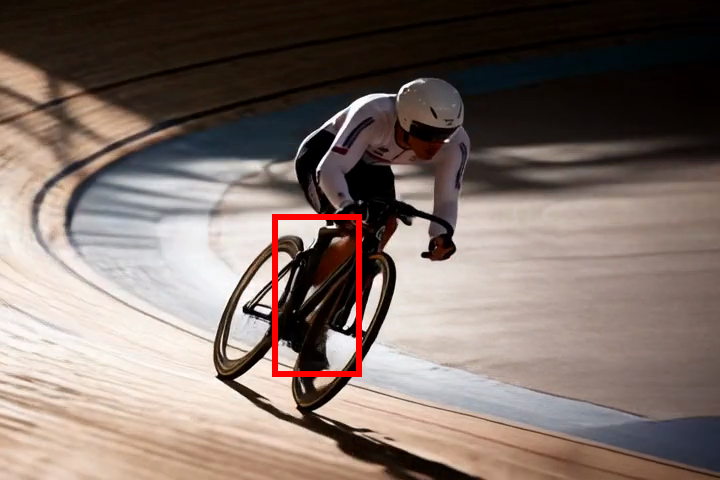} &
    \includegraphics[width=\imW]{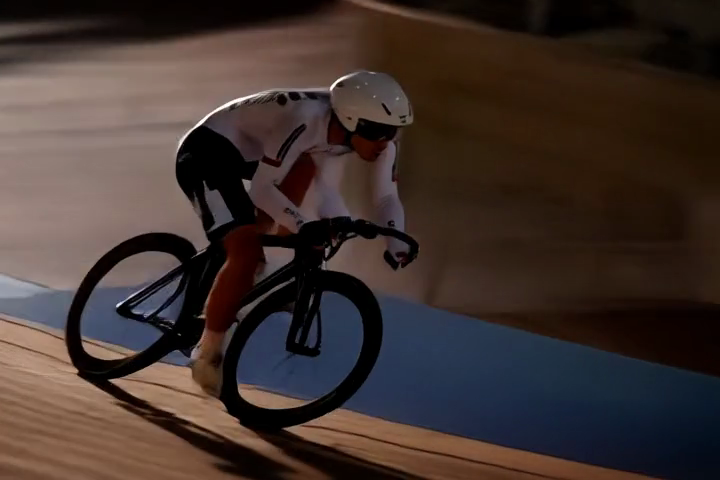} &
    \includegraphics[width=\imW]{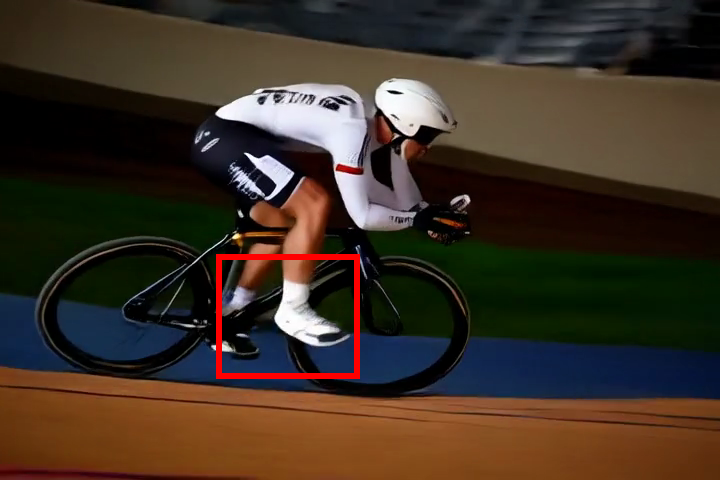} &
    \includegraphics[width=\imW]{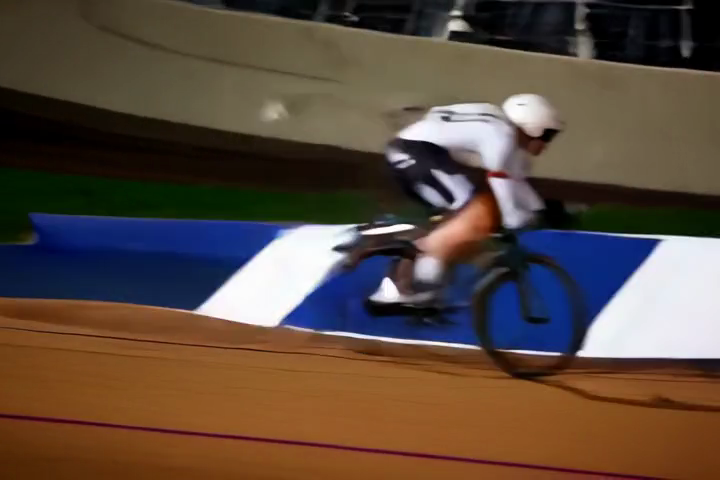} \\

    \rotatebox{90}{\small + Mask sup.} 
    &\includegraphics[width=\imW]{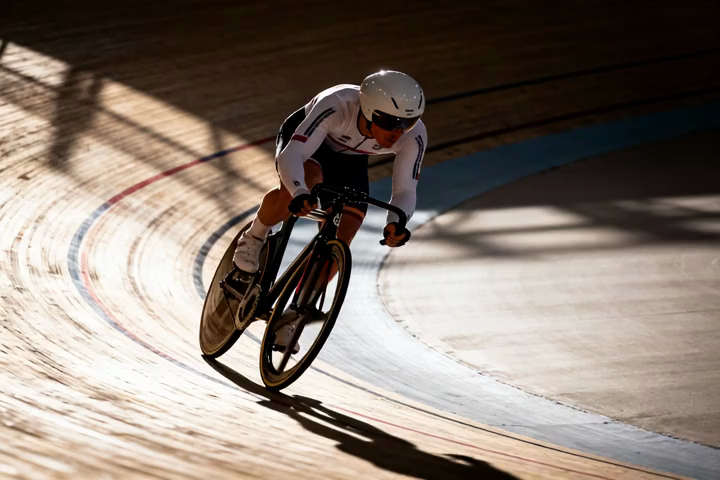} &
    \includegraphics[width=\imW]{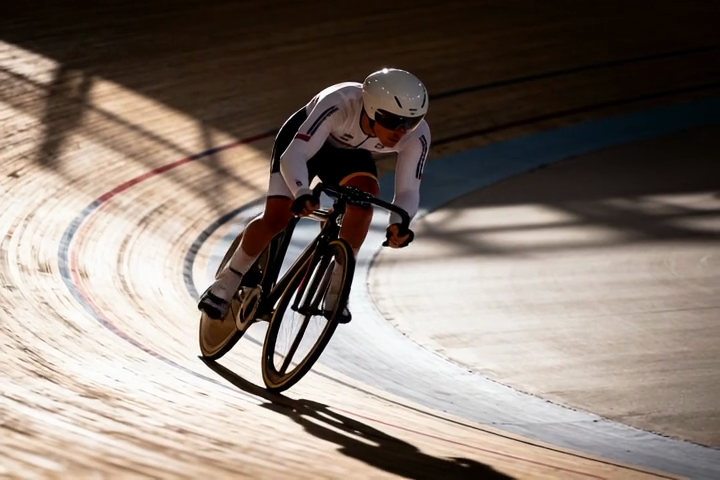} &
    \includegraphics[width=\imW]{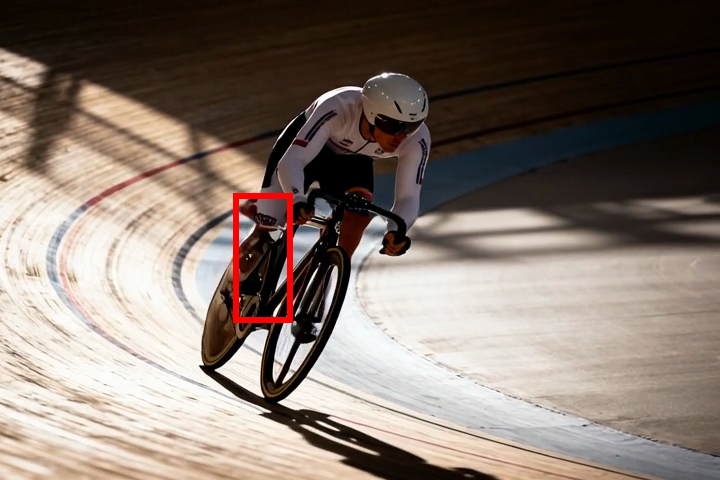} &
    \includegraphics[width=\imW]{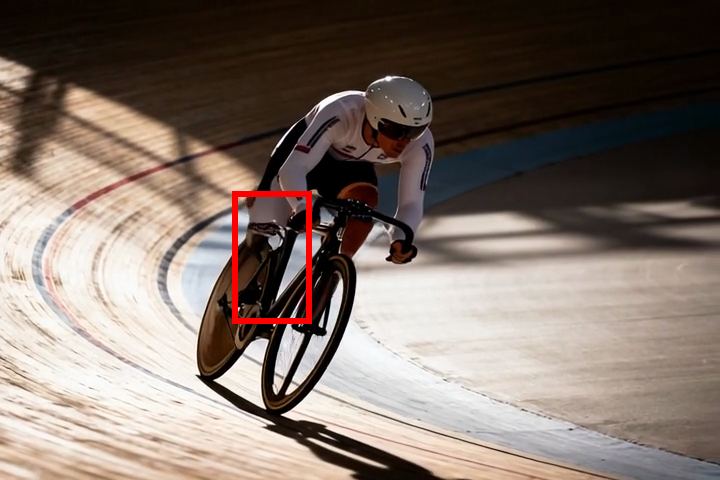} &
    \includegraphics[width=\imW]{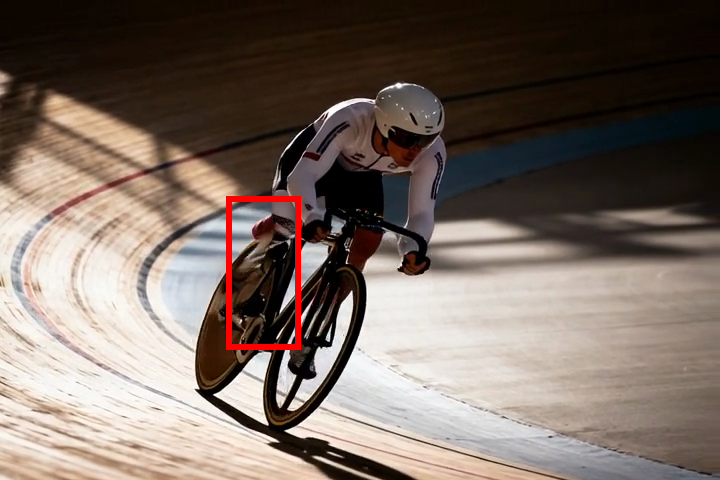} \\

    \rotatebox{90}{\small \textbf{\model{}}} 
     &\includegraphics[width=\imW]{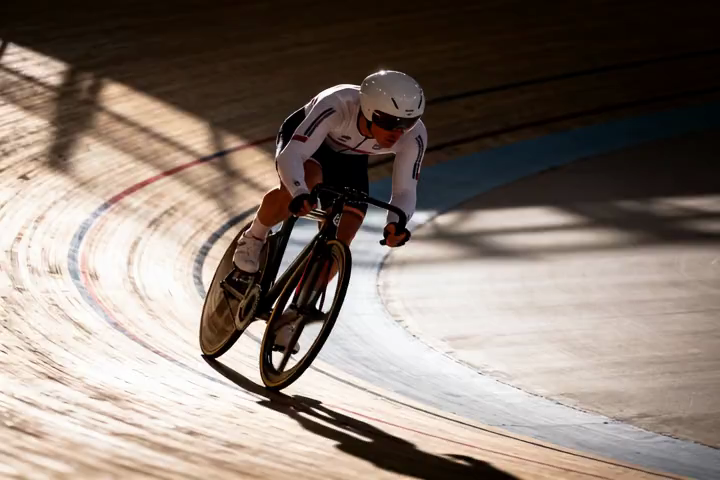} &
    \includegraphics[width=\imW]{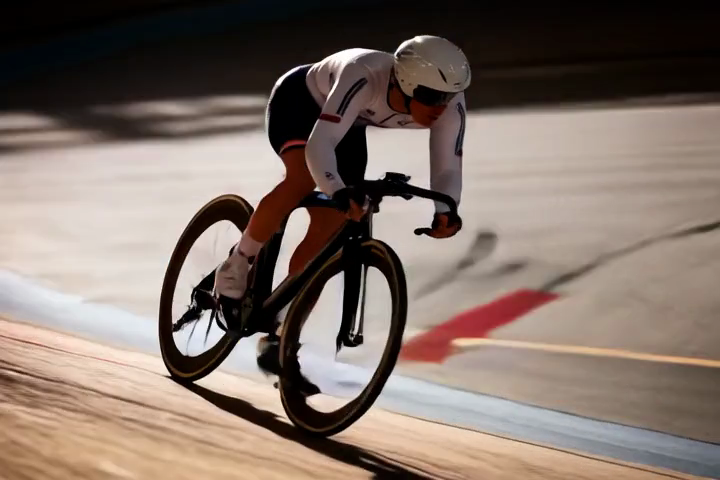} &
    \includegraphics[width=\imW]{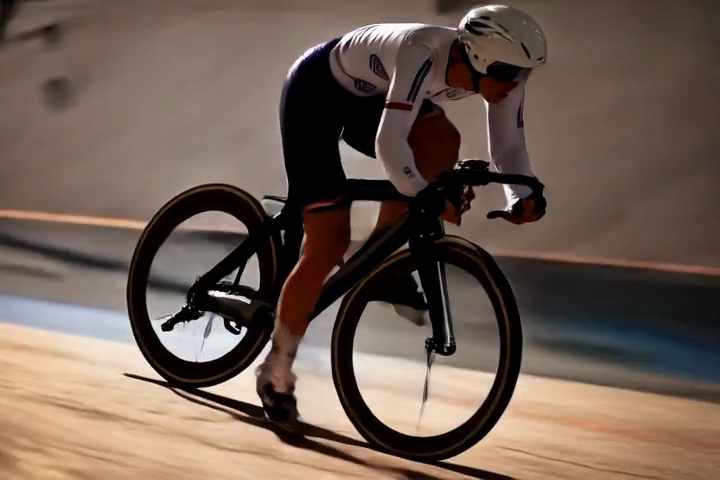} &
    \includegraphics[width=\imW]{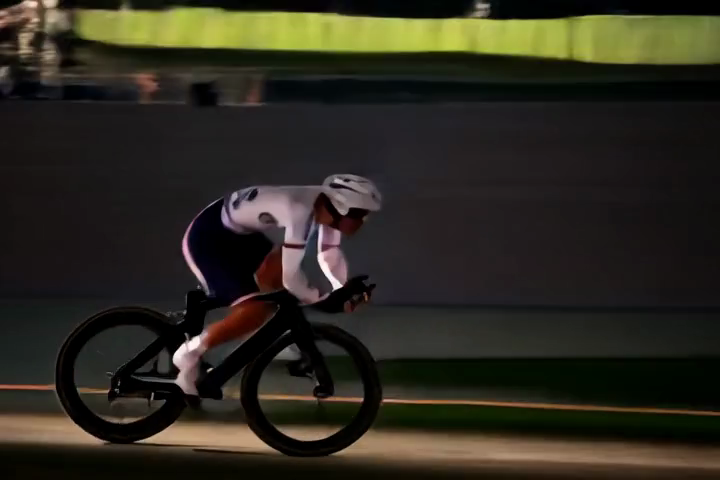} &
    \includegraphics[width=\imW]{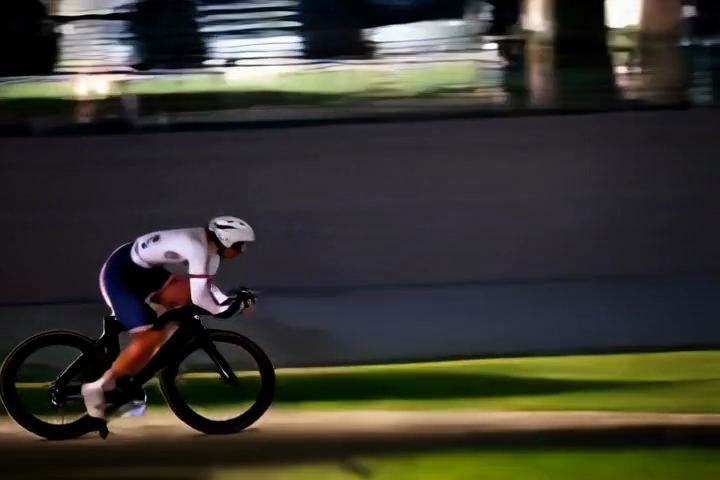} \\

    \rotatebox{90}{\small Base model}  &
     \fcolorbox{blue}{white}{\includegraphics[width=\imW]{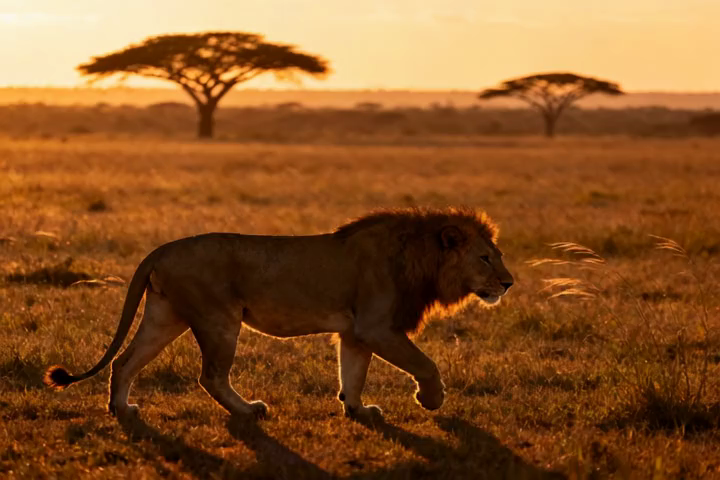}} &
    \includegraphics[width=\imW]{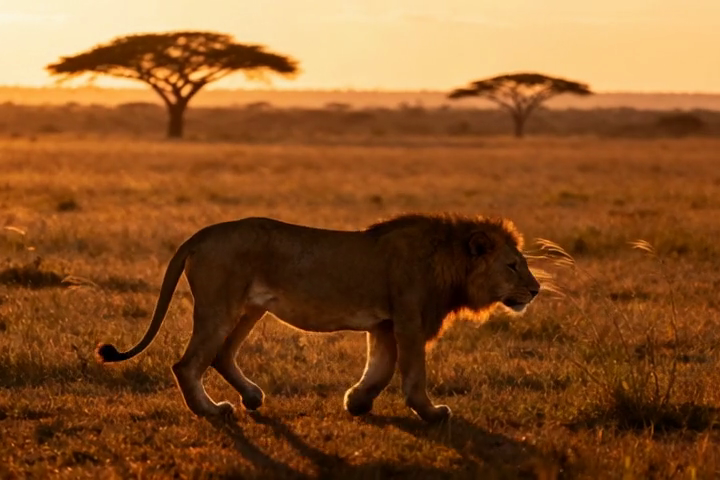} &
    \includegraphics[width=\imW]{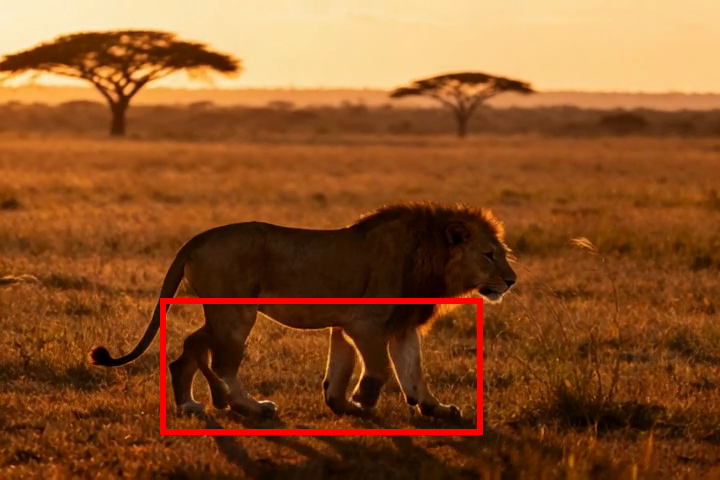} &
    \includegraphics[width=\imW]{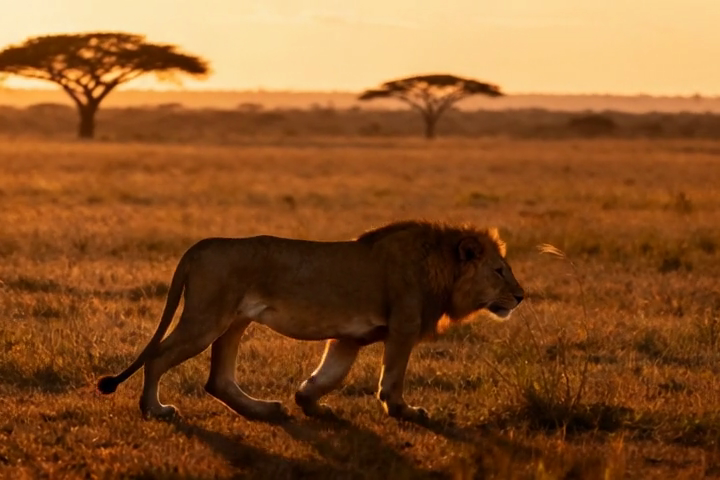} &
    \includegraphics[width=\imW]{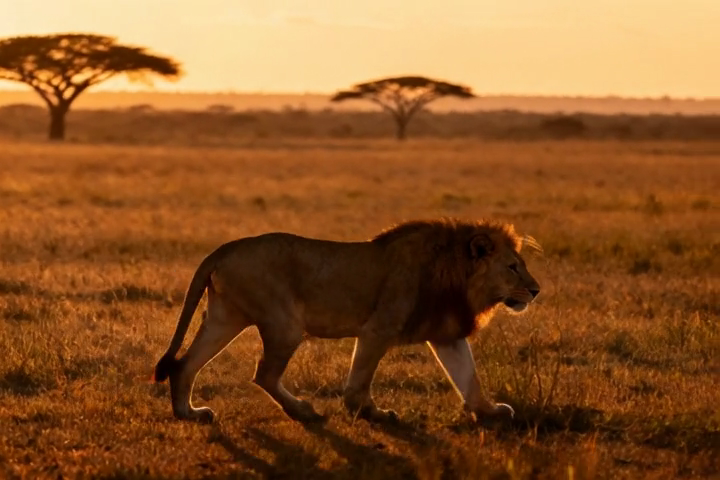} \\

      \rotatebox{90}{\small + Fine-tuning}  &
   \includegraphics[width=\imW]{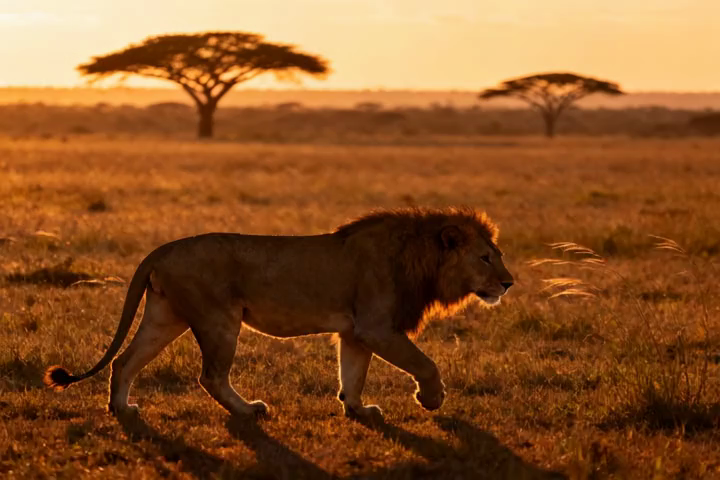} &
    \includegraphics[width=\imW]{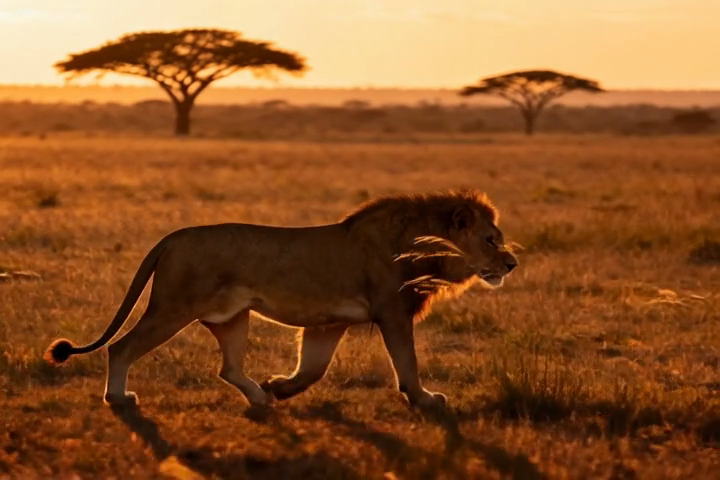} &
    \includegraphics[width=\imW]{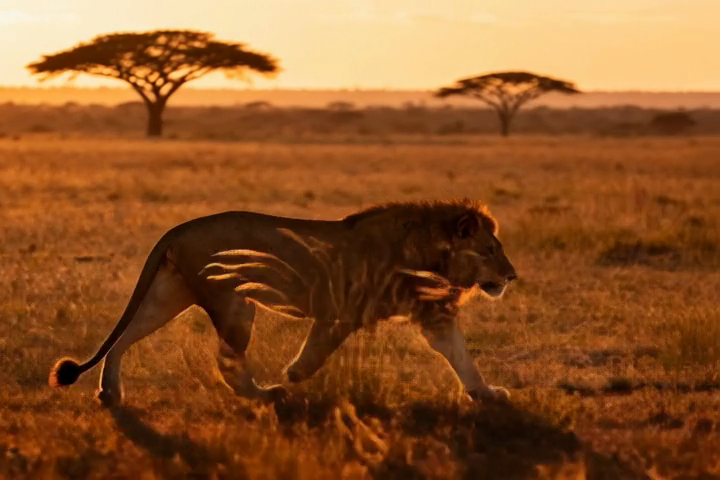} &
    \includegraphics[width=\imW]{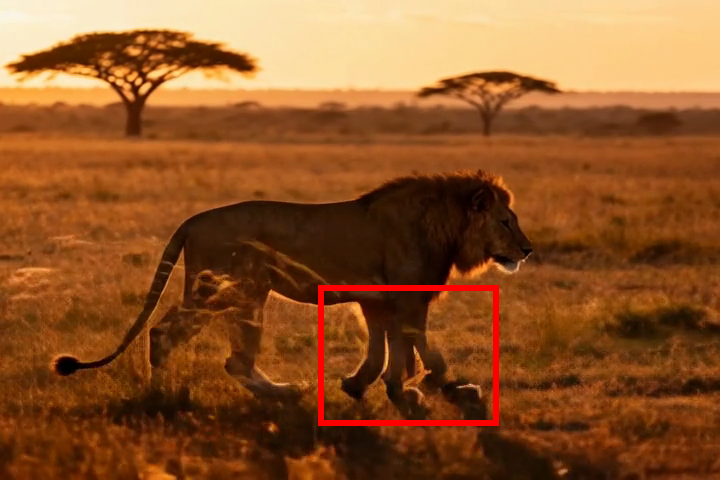} &
    \includegraphics[width=\imW]{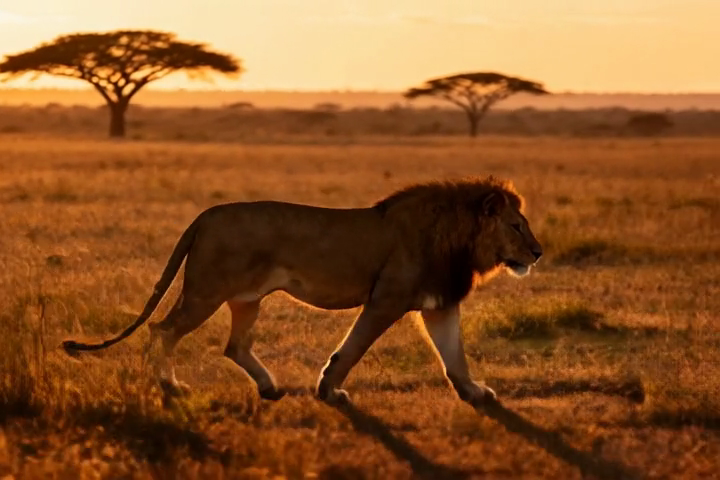} \\

     \rotatebox{90}{\small + Mask sup.}  &
    \includegraphics[width=\imW]{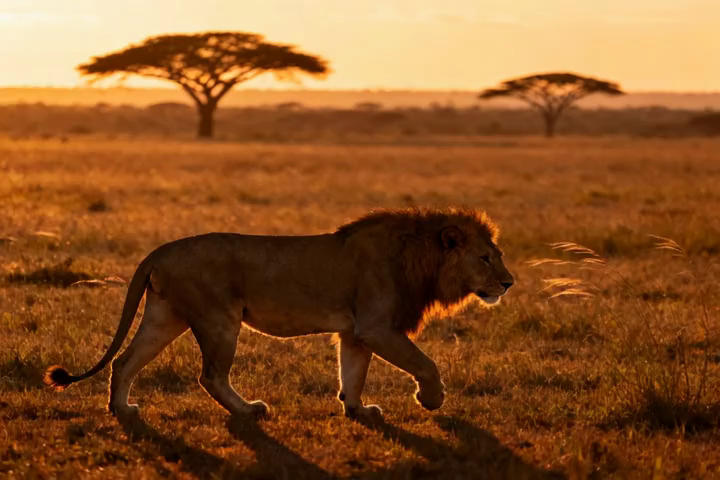} &
    \includegraphics[width=\imW]{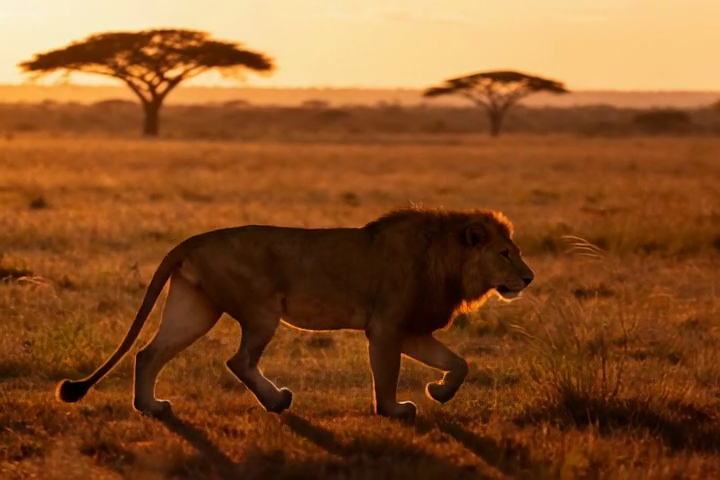} &
    \includegraphics[width=\imW]{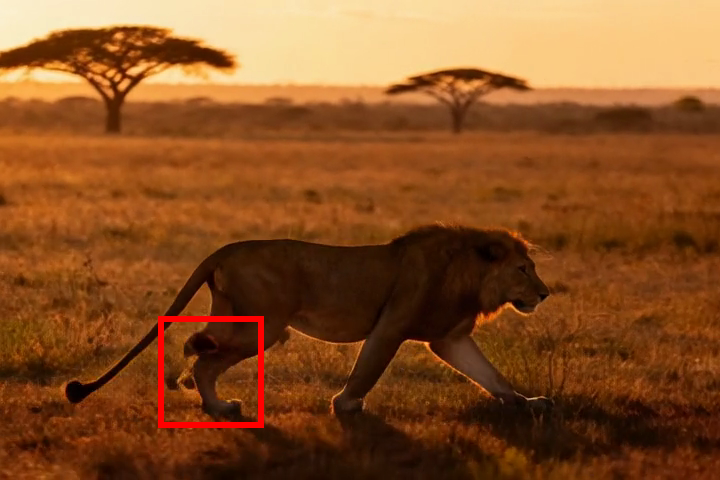} &
    \includegraphics[width=\imW]{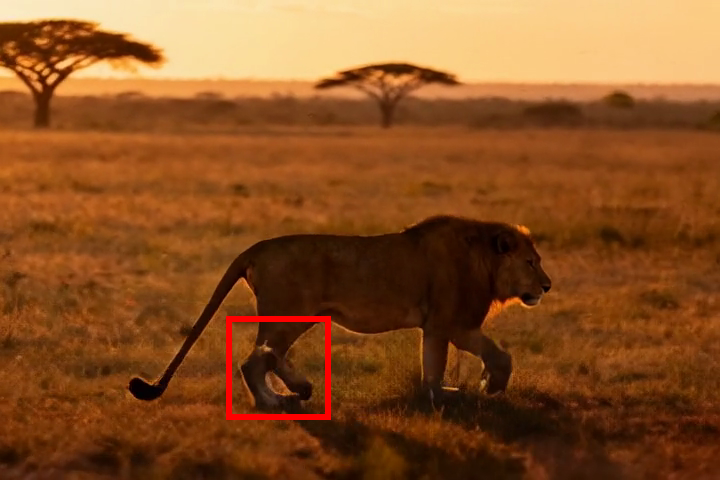} &
    \includegraphics[width=\imW]{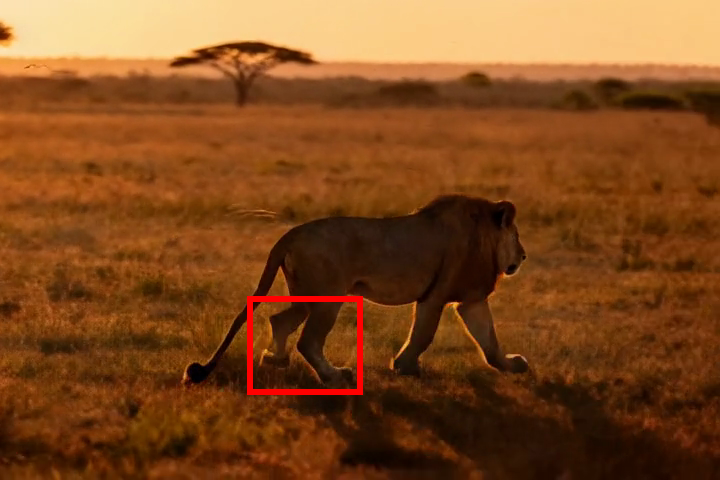} \\

    \rotatebox{90}{\small \textbf{\model{}}} &
    \includegraphics[width=\imW]{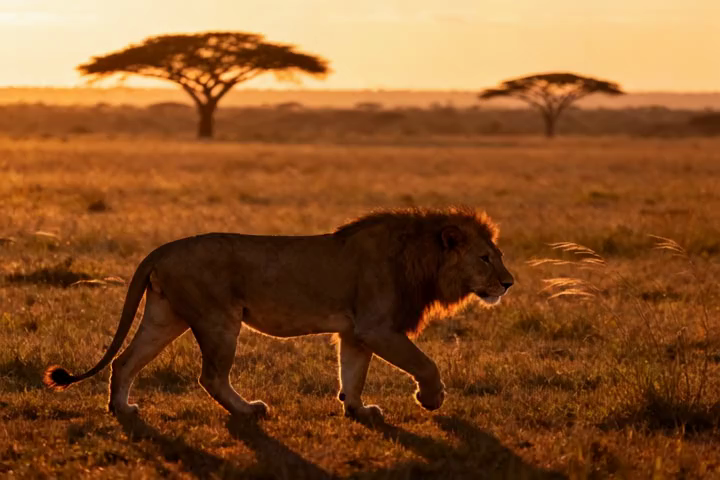} &
    \includegraphics[width=\imW]{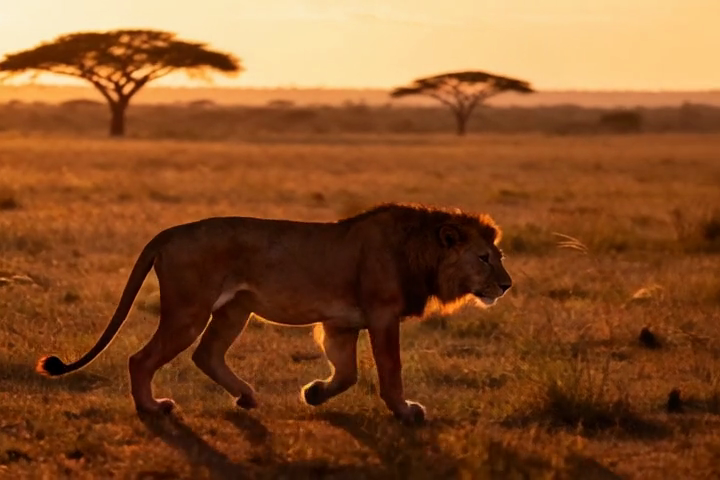} &
    \includegraphics[width=\imW]{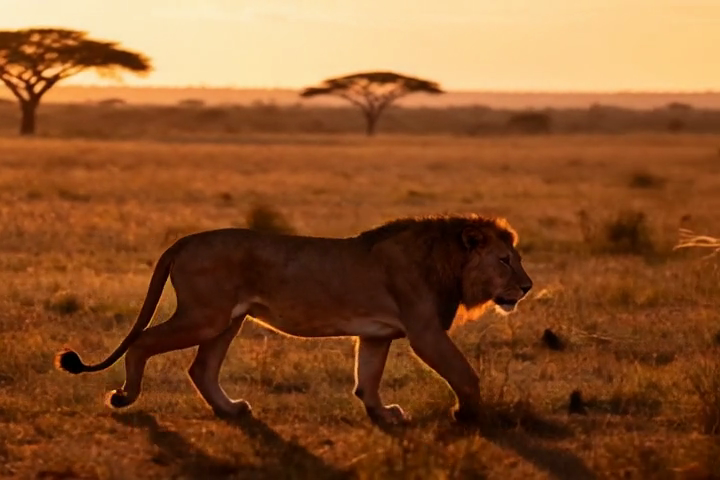} &
    \includegraphics[width=\imW]{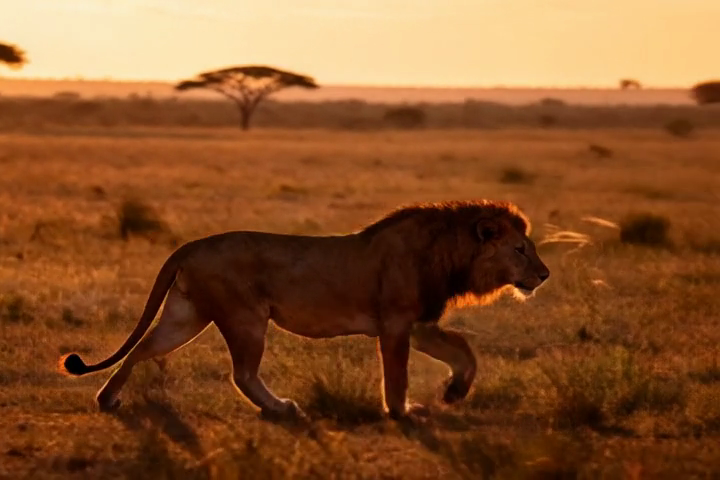} &
    \includegraphics[width=\imW]{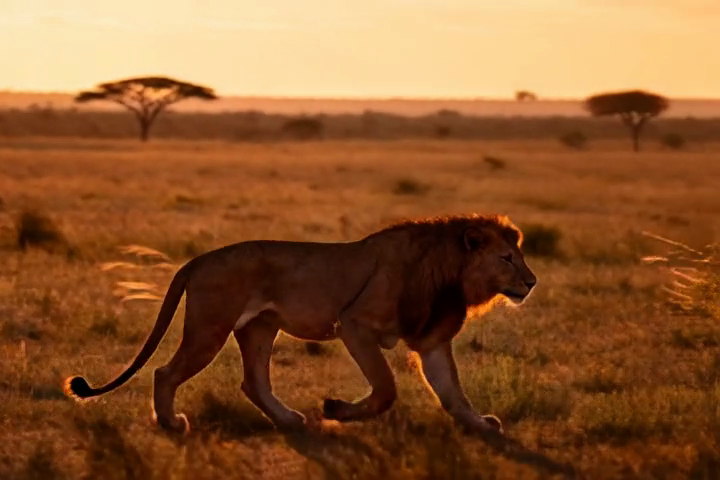} \\
\end{tabular}

\caption{
    \textbf{Qualitative comparison on articulated motion.} The blue box indicates the input image. Red boxes highlight common failure modes in baselines, including structural distortion (cyclist's legs) and physical implausibility (three-legged lion, inconsistent leg motion). In contrast, our method (\model{}) consistently maintains structural integrity. ''+ Fine-tuning" is the LoRA fine-tuning baseline; ''+ Mask sup." is the mask supervision baseline. We recommend viewing the supplementary videos.
}

\label{fig:comparison}

\end{figure*}

\begin{table*}[t]
  \centering
  \caption{\textbf{Quantitative comparison on established video generation benchmarks.} Our method outperforms all fine-tuning baselines and achieves performance comparable to the strong open-source model HunyuanVid, while significantly surpassing it in perceptual quality (FVD). \textbf{BC}: Background Consistency; \textbf{SC}: Subject Consistency; \textbf{MS}: Motion Smoothness. Extended Motion Score incorporates I2V Subject and Background Consistency (weight 0.5). Higher Motion/Extended Motion Scores indicate better structure preservation; lower FVD indicates superior perceptual quality. Baselines with Dynamic Degree lower than the base model are excluded from VBench comparisons to ensure fairness.}
  
  \label{tab:main_results}
    \begin{tabular}{lcccccccc}
      \toprule
      Method & BC & SC & MS & Motion Score$\uparrow$ & Ext Motion Score$\uparrow$ & FVD$\downarrow$ \\
      \midrule
      CogVideoX (base model) & 97.30 & 94.43 & 98.17 & 94.80 & 95.50 & 660.29  \\
      \quad + LoRA Fine-tuning & 97.44 & 93.47 & 97.76 & 94.02 & 94.74 & 465.00     \\
      \quad + Mask Supervision & - & - & - & - & - & 397.73       \\
      \quad + REPA & 97.41 & 91.99 & 97.31 & 92.91 & 93.77 & 457.59               \\
      \quad + \textbf{\model{} (Ours)} & \textbf{97.88} & \textbf{94.76} & \textbf{98.45} & \textbf{95.51} & \textbf{96.03} & \textbf{360.57}   \\
      \midrule 
      HunyuanVid & 96.85 & 95.32 & 98.76 & 95.62 & 96.24 & 583.99 \\
      \bottomrule
    \end{tabular}%
\end{table*}

\begin{figure}[t]
\centering
\includegraphics[width=0.48\textwidth]{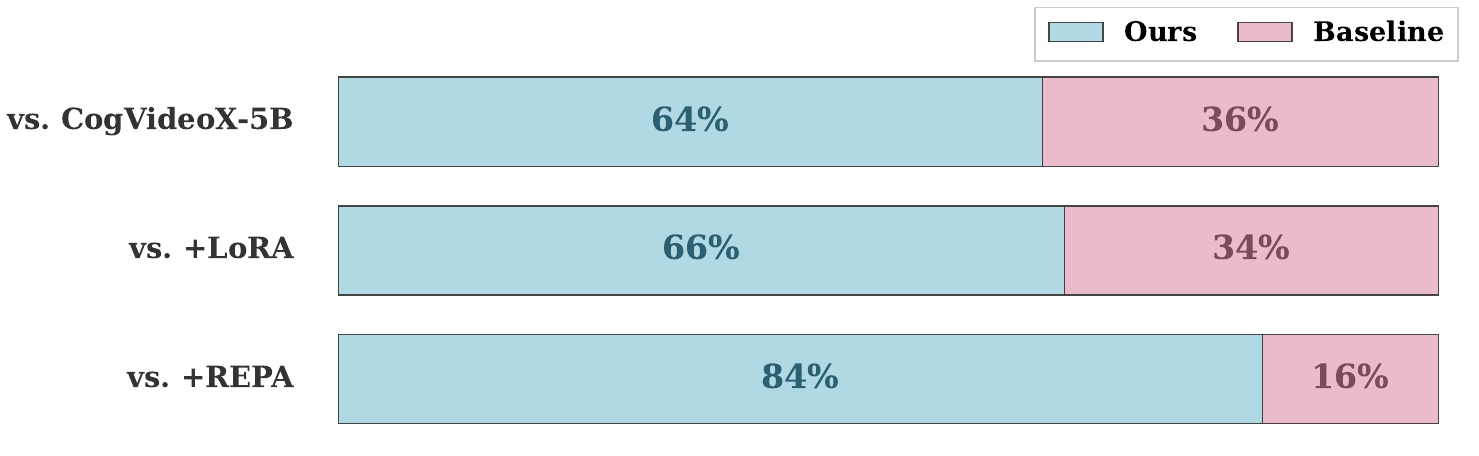}
\caption{\textbf{Human preference win rates.} Participants were shown pairwise comparisons and asked to select the video with 'superior limb consistency and fewer motion artifacts'. Our method was strongly preferred ($>64\%$) over all baselines, confirming its superior ability to generate plausible, artifact-free motion.}
\label{fig:user-study}
\end{figure}

\paragraph{Dataset \& training configuration.}
We curate a motion-focused dataset of 9,837 single-subject video clips from open-source video generation datasets(Panda70M~\cite{chen2024panda70m}, MMTrailer~\cite{chi2024mmtrailmultimodaltrailervideo}, MotionVid~\cite{wang2025humandreamer}), capturing diverse motion patterns across animals and humans. All videos are at 8 FPS, capped at 100 frames. Our approach builds upon CogVideoX-5B-I2V~\cite{yang2024cogvideox}, extracting intermediate DiT features from the 25th block output as $F_{\mathrm{diff}}$. We first obtain object bounding boxes using GroundingDINO~\cite{liu2024grounding} to prompt SAM2 mask generation. To align with DiT features, we propagate subject masks from both temporal directions: using the first-frame mask to compute forward features $\mathbf{F}_\mathrm{mem}^\mathrm{fwd}$ and the last-frame mask for backward features $\mathbf{F}_\mathrm{mem}^\mathrm{bwd}$. To avoid SAM2 inference overhead during training, we precompute features for clips starting at every 20 frames. 

We implement LoRA fine-tuning with rank 256 and scaling factor $\alpha=128$. Training uses AdamW optimization with learning rate $1\times 10^{-4}$ and momentum parameter $(\beta_1, \beta_2)=(0.9,0.95)$. We train for 3,000 steps on $8 \times \text{H200}$ GPUs, with global batch size 32 through gradient accumulation over four steps per GPU.

\vspace{0.2cm}

\noindent \textbf{Baselines.}
We compare our complete approach, which uses bidirectional feature fusion and Local Gram Flow loss, with four baselines: (1) CogVideoX-5B-I2V as the base model; (2) "+ LoRA fine-tuning", 
which simply fine-tunes the base model with LoRA on our curated motion dataset; (3) "+ Mask supervision", which adds a linear projection layer on top of our projection head $\mathcal{P}$ to directly predict the subject mask, trained with mask loss supervision instead of feature alignment; (4) "+REPA" , which utilizes REPA Loss~\cite{yu2024representation} to align DiT features with external DINOv3~\cite{simeoni2025dinov3} features.

\vspace{0.2cm}

\noindent \textbf{Evaluation protocol.}
We evaluate across three complementary axes: objective motion metrics, perceptual quality, and human preference. For objective evaluation, we filter 85 images and compute four metrics from the VBench-I2V suite~\cite{huang2023vbench}: Motion Smoothness, Subject Consistency, and Background Consistency, and Dynamic Degree. We define consistency metrics as measures of temporal structure stability, while Dynamic Degree quantifies the magnitude of motion. Since consistency scores often correlate negatively with motion magnitude(i.e., static videos trivially achieve perfect consistency), to ensure a fair comparison of motion quality, we exclude baselines with a lower Dynamic Degree than the base model. The overall Motion Score is obtained by averaging the smoothness and consistency metrics (min-max normalized). For the Extended Motion Score, we incorporate I2V-Subject and I2V-Background Consistency with a 0.5 weight, following the official VBench protocol. For perceptual quality, we compute Fréchet Video Distance (FVD)~\cite{unterthiner2018towards} on a separate set of 200 videos randomly sampled from the training dataset. 
For human preference, we conduct a double-blind user study using 40 randomly sampled prompts. Participants are presented with two side-by-side videos (Ours vs. Baseline) in random order and asked to select the preferred one based on motion smoothness and subject consistency. All methods generate 49-frame videos at 8 fps and 720×480 resolution, using 50 denoising steps with guidance scale 6.0.

\subsection{Results}
\noindent\textbf{\model{} achieves superior structure preservation and perceptual quality compared to all baselines.} As shown in Table~\ref{tab:main_results}, our method outperforms the base CogVideoX model and other fine-tuning strategies across all reported metrics. Notably, we achieve an FVD of 360.57, a substantial reduction compared to the strongest baseline (LoRA Fine-tuning at 465.00) and the REPA baseline (457.59). This indicates that our model generates videos with significantly higher perceptual fidelity. Furthermore, our method attains the highest scores in Motion Score (95.51) and Extended Motion Score (96.03), confirming that distilling internal priors from SAM2 effectively suppresses the temporal flickering and identity degradation often observed in standard video diffusion models.

\vspace{0.2cm}
\noindent\textbf{Feature distillation outperforms coarse mask supervision.} We compare our feature-level alignment against a baseline trained with explicit mask supervision (``+ Mask Supervision"). We build this baseline by adding a linear projection layer on top of $\mathcal{P}$ to predict the mask. While mask supervision delineates subject boundaries, it lacks the fine-grained internal correspondence information necessary for articulating complex motion. Consequently, the mask supervision baseline suffers from severe structural artifacts (e.g., cyclist and lion legs in Fig.~\ref{fig:comparison}) and achieves a poorer FVD score of 397.73 compared to our 360.57. Note that we exclude the mask supervision baseline from VBench consistency metrics in Table~\ref{tab:main_results} because it collapses towards static generation (Dynamic Degree 44.59 vs. Base model 45.95), which would yield artificially inflated consistency scores.

\vspace{0.2cm}
\noindent \textbf{Video-aware priors are essential for temporal consistency.} To validate the necessity of using a video foundation model (SAM2) as the teacher, we compare against REPA~\cite{yu2024representation}, which aligns DiT features with the image-based DINOv3 encoder. As DINO is trained on static images, it lacks inherent knowledge of temporal continuity. This limitation is reflected in Table~\ref{tab:main_results}, where REPA yields a significantly lower Motion Score (92.91) compared to our method (95.51). This result confirms that aligning with SAM2's memory-based features transfers crucial temporal coherence signals that image-only encoders cannot provide.

\vspace{0.2cm}

\noindent \textbf{Human evaluators consistently prefer our method.} As illustrated in Figure~\ref{fig:user-study}, our method achieves the highest win rates in a double-blind user study, outperforming the base model and fine-tuning baselines by a wide margin. Qualitative results in Figure~\ref{fig:comparison} corroborate this preference: while baselines often struggle with limb consistency (e.g., the disappearing/reappearing legs of the cyclist), our method maintains subject identity and structural integrity throughout the sequence. This demonstrates that our bidirectional alignment strategy successfully translates the robust segmentation priors of SAM2 into high-fidelity video generation.

\subsection{Ablations}

\begin{table}[t]   
  \centering
  \resizebox{\columnwidth}{!}{
      \begin{tabular}{lcc}
          \toprule
          Configurations & Motion Score$\uparrow$ & Ext Motion Score$\uparrow$ \\
          \midrule
          + LoRA only & 94.02 & 94.74 \\
          \midrule
          w/o LGF & 94.58 & 95.24 \\
          w/o KL & 94.51 & 95.16  \\
          \midrule
          w/ Forward-Only Teacher & 95.07 & 95.58 \\
          w/ Separate Projectors & 94.68 & 95.24 \\
          Feature-Space Fusion & 94.16 & 94.83 \\
          \midrule 
          \textbf{\model{} (LGF Fusion)} & \textbf{95.51} & \textbf{96.03} \\
          \bottomrule       
      \end{tabular}
    }
    \caption{\textbf{Ablation study on core components (VBench-I2V, $\uparrow$).} Using a simple $\ell_2$ loss (''w/o LGF") and an $\ell_2$ loss within LGF space (''w/o KL") both yield marginal improvements only, confirming our full LGF-KL's superiority in capturing motion structure. Removing bidirectional processing (''w/ Forward-Only Teacher") or fusion mechanisms (''w/ Separate Projectors") degrades performance, while feature-level adding (''Feature-Space Fusion") underperforms ours (''LGF Fusion"), validating each design choice.} 
  \label{tab:ablations}
\end{table}

\begin{figure}[h]
\centering
\setlength{\fboxsep}{0pt}   
\setlength{\fboxrule}{1pt}  
\def\imW{0.3\linewidth}

\begin{tabular}{c@{\hskip 0pt}c@{\hskip 0pt}c@{\hskip 0pt}c@{\hskip 0pt}c@{\hskip 0pt}c}
    \rotatebox{90}{\small w/o LGF loss} 
    \includegraphics[width=\imW]{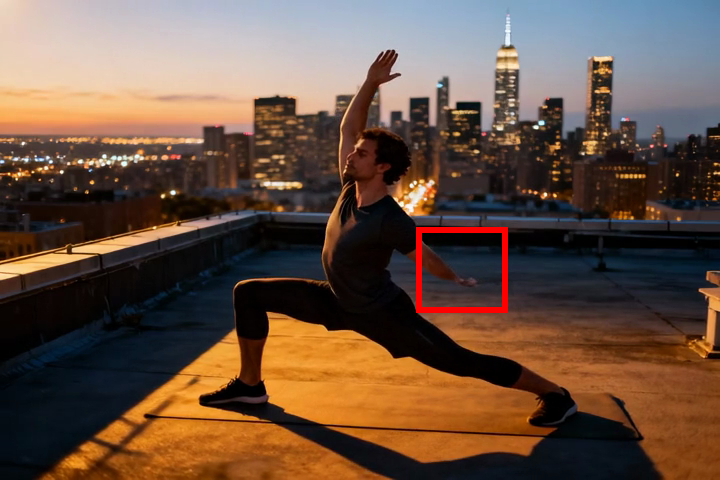} &
    \includegraphics[width=\imW]{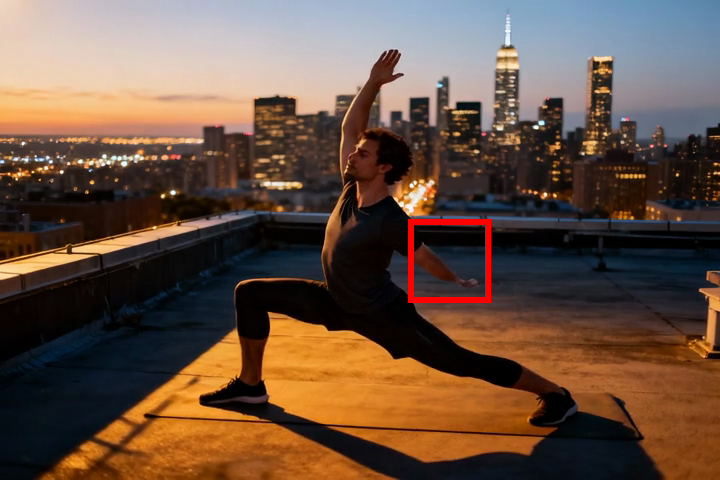} &
    \includegraphics[width=\imW]{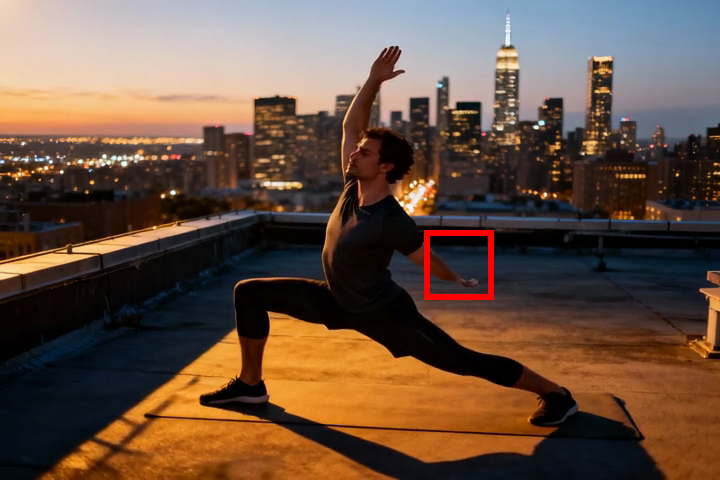} \\

    \rotatebox{90}{\small \textbf{w/ LGF Loss}} 
    \includegraphics[width=\imW]{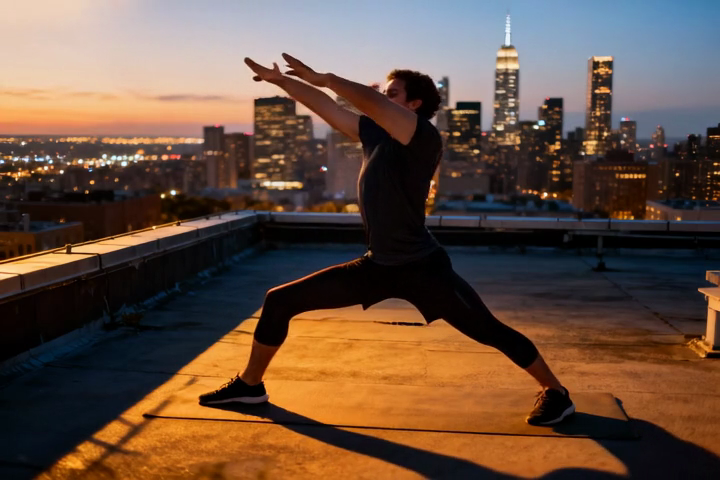} &
    \includegraphics[width=\imW]{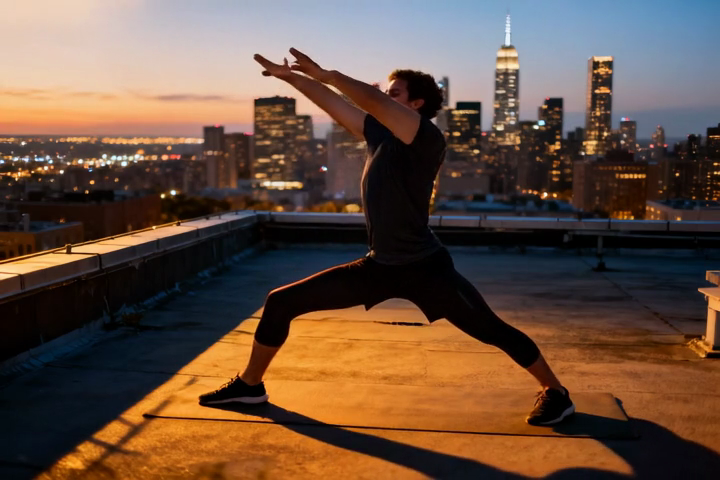} &
    \includegraphics[width=\imW]{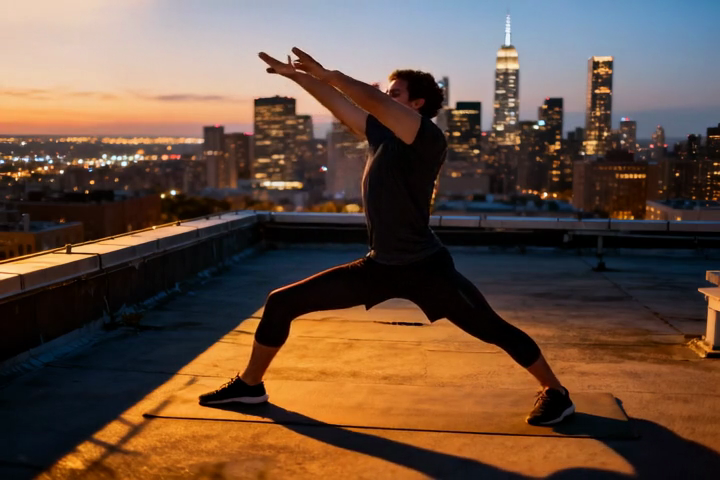} \\  
\end{tabular}

\caption{
    \textbf{Qualitative ablation of our LGF-KL Loss.}  (Top) Using a standard $\ell_2$ loss ('w/o LGF Loss') on raw features results in visible temporal jitter (note the flickering in the arm). (Bottom) Our full method ('w/ LGF Loss') produces a visibly smoother and more stable motion, validating the design of aligning relational distributions (LGF-KL) over absolute values ($\ell_2$).
}
\label{fig:ablation_gram}
\end{figure}

\begin{figure}[h]
\centering
\setlength{\fboxsep}{0pt}   
\setlength{\fboxrule}{1pt}  
\def\imW{0.3\linewidth}

\begin{tabular}{c@{\hskip 0pt}c@{\hskip 0pt}c@{\hskip 0pt}c@{\hskip 0pt}c@{\hskip 0pt}c}
    \rotatebox{90}{\small w/ Fwd-Only} 
    \includegraphics[width=\imW]
    {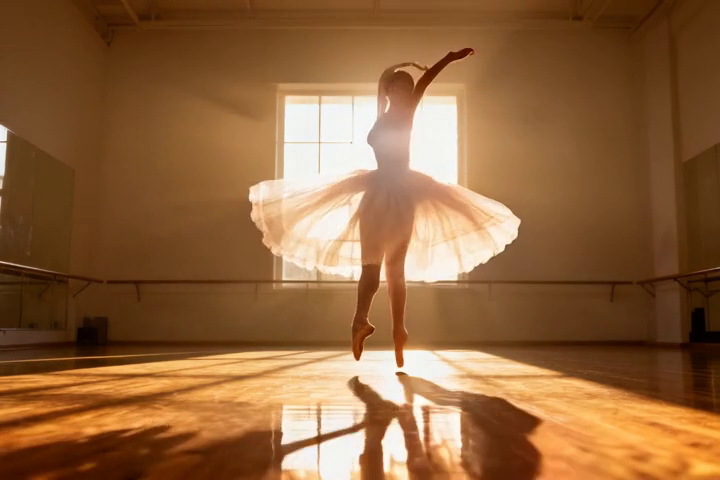} &
    \includegraphics[width=\imW]{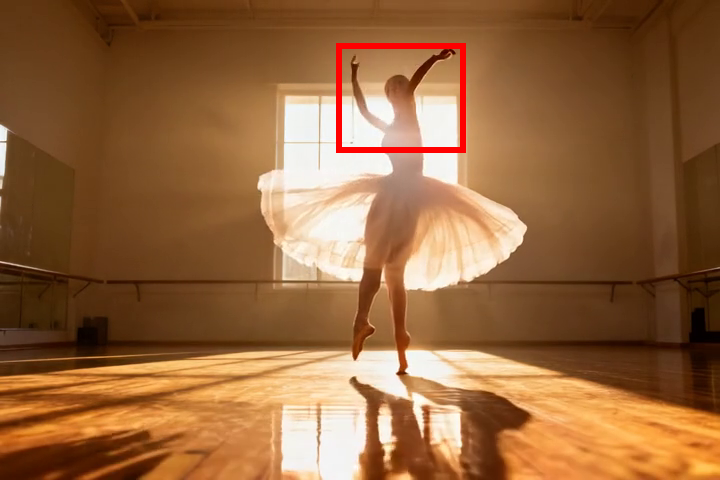} &
    \includegraphics[width=\imW]{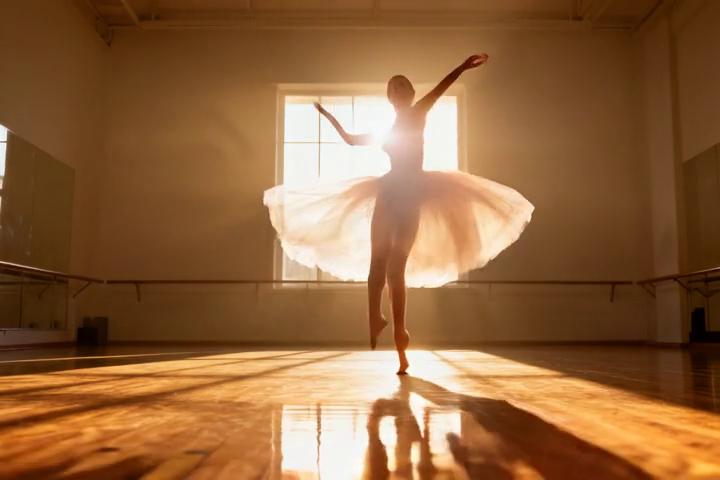} \\
    
    \rotatebox{90}{\small Feat-Space} 
     \includegraphics[width=\imW]{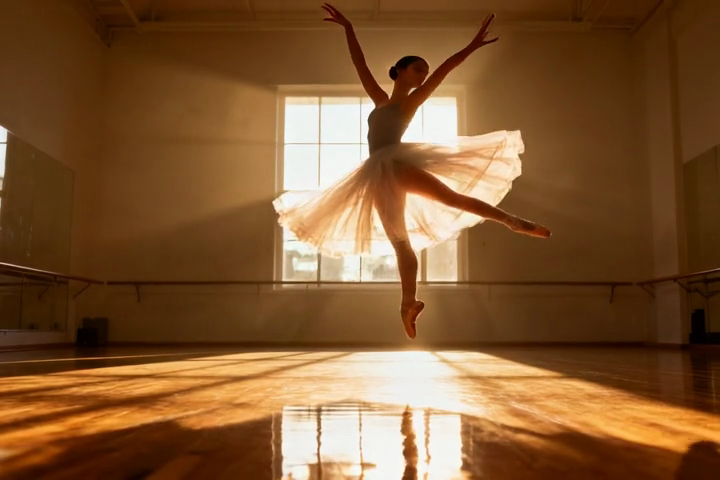} &
    \includegraphics[width=\imW]{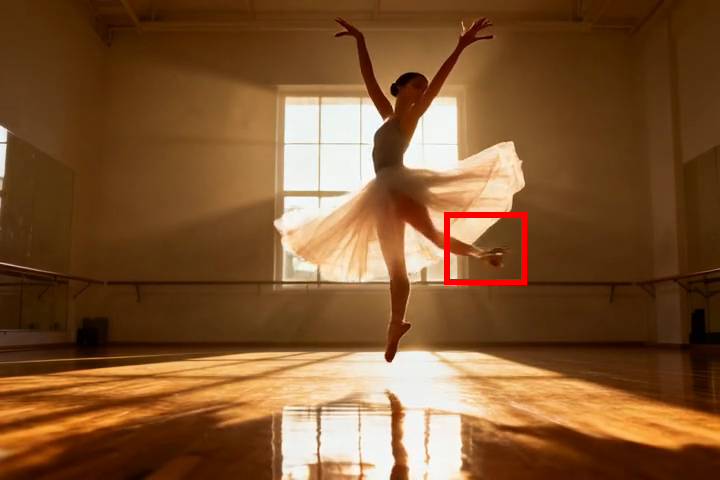} &
    \includegraphics[width=\imW]{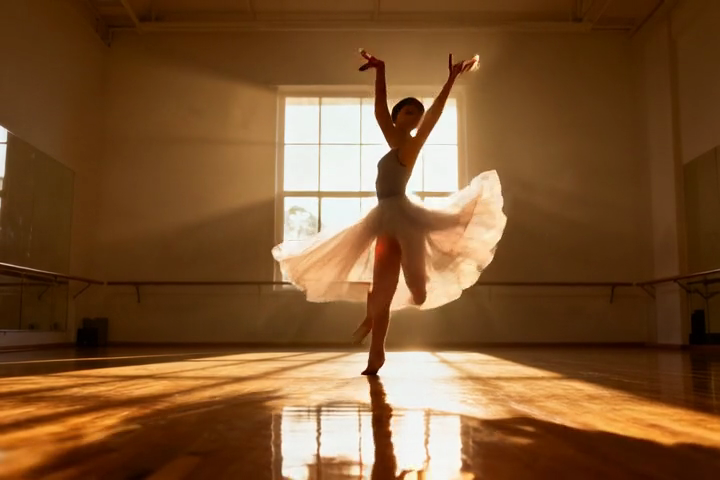} \\
    \rotatebox{90}{\small \textbf{LGF Fusion}} 
    \includegraphics[width=\imW]{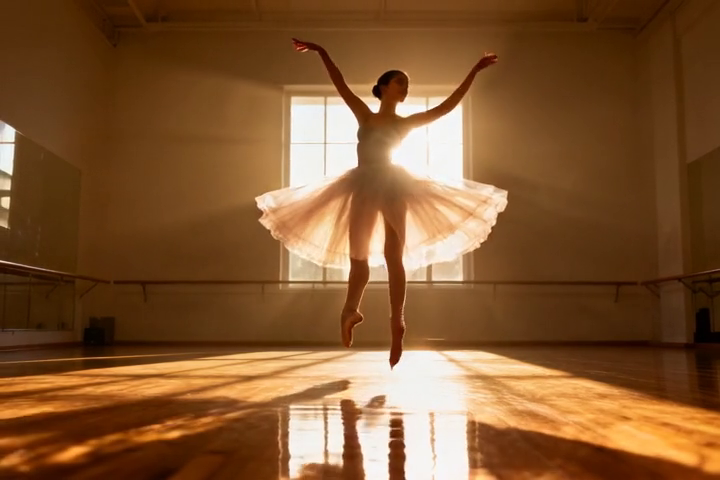} &
    \includegraphics[width=\imW]{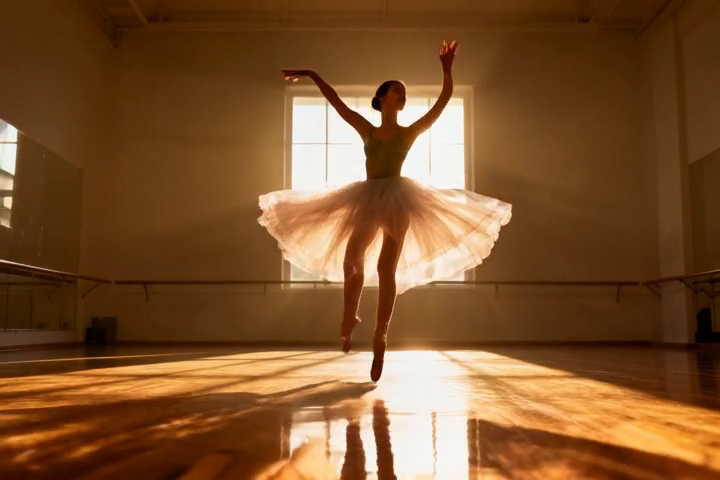} &
    \includegraphics[width=\imW]{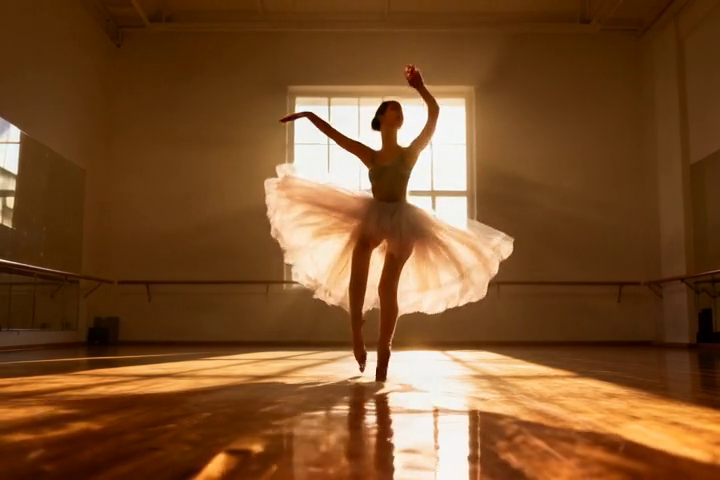} \\  
\end{tabular}
\caption{
   \textbf{Qualitative ablation of our bidirectional LGF fusion.}(Top) The forward-only teacher (''w/ Fwd-Only") produces artifacts, like the ballerina's arm 'folding' incorrectly (red box). (Middle) ''Feat-Space Fusion" causes structural tearing. (Bottom) Our ''LGF Fusion" correctly preserves the limb's topological structure, highlighting the necessity of both bidirectional information and a robust fusion strategy.
}
\label{fig:ablation_bidireictional}
\end{figure}

We conduct systematic ablation studies on the VBench-I2V 85-image subset to isolate the contribution of our two core components: the LGF fused bidirectional SAM2 teacher  and the LGF-KL distillation loss.

\vspace{0.2cm}
\noindent \textbf{LGF fusion is essential to resolve bidirectional conflicts.}
We validate our teacher design in Table~\ref{tab:ablations}. While using only the forward stream ($\mathbf{F}_\mathrm{mem}^\mathrm{fwd}$) yields a strong baseline (Motion Score 95.07), naively incorporating the backward stream via separate projectors causes gradient conflicts, destabilizing training and degrading performance. More critically, fusing forward and backward streams directly in the \textit{feature space} leads to catastrophic collapse (94.16), barely outperforming the LoRA-only baseline. This suggests that raw features from opposite temporal directions interfere destructively. By contrast, our proposed \textbf{LGF Fusion} acts as a harmonic integration, resolving these conflicts to achieve the highest score (95.51). Qualitative results in Figure~\ref{fig:ablation_bidireictional} confirm that this relational fusion is key to leveraging bidirectional priors without introducing artifacts.

\vspace{0.2cm}
\noindent \textbf{KL divergence outperforms $\ell_2$ for structural alignment.}
We further ablate the loss function design given our LGF teacher. As shown in Table~\ref{tab:ablations}, applying a standard $\ell_2$ loss directly on raw features performs poorly (94.58), proving that strict element-wise alignment is too rigid for transferring high-level motion priors. More revealingly, applying an $\ell_2$ loss within the LGF space performs even worse (94.51). This demonstrates that the LGF operator alone is insufficient; a naive value-based alignment of its relational features fails to capture the correct motion priors. It is the combination of LGF (to capture relational structure) and the KL divergence (to align probabilistic distributions) that is essential. Our full LGF-KL loss achieves the best performance (95.51), confirming that aligning the relative spatio-temporal distributions via KL divergence is superior to forcing exact value matches. Figure~\ref{fig:ablation_gram} visually demonstrates the smoother temporal transitions achieved by this design.

%% file: sec/5_conclusion.tex
\section{Conclusion}
In this paper, we propose a novel framework that effectively distills the rich, structure-preserving motion priors from SAM2 into video diffusion models. Departing from methods relying on external control signals or limited datasets, we demonstrate that aligning generative features with dense correspondence representations offers a more intrinsic solution to articulated motion generation. Our core contributions—bidirectional feature fusion and the Local Gram Flow loss—enable the seamless transfer of fine-grained motion knowledge without requiring architectural modifications. Extensive experiments validate that our approach not only achieves superior performance on standard benchmarks but also paves the way for leveraging discriminative vision foundation models to enhance generative video dynamics.

%% file: sec/99_supp.tex
\newpage
\appendix
\onecolumn

\begin{table}
\centering
    \begin{tabular}{lccccccc}
          \toprule
          Method & BC$\uparrow$ & SC$\uparrow$ & MS$\uparrow$ & Motion Score$\uparrow$ & Ext Motion Score$\uparrow$ \\
          \midrule
          $\text{Track4Gen}^*$ & 97.32 & 94.67 & 98.35 & 95.11 & 95.69 \\
          \textbf{\model{} (Ours)} & \textbf{97.88} & \textbf{94.76} & \textbf{98.45} & \textbf{95.51} & \textbf{96.03} \\
          \bottomrule
    \end{tabular}%
\caption{\textbf{Quantitative Analysis of Additional Baseline.} Comparison between point-based control and our feature-based approach. $\text{Track4Gen}^*$ denotes the modified version adapted to the CogVideoX architecture. Abbreviations: BC (Background Consistency), SC (Subject Consistency), MS (Motion Smoothness).}
\label{tab:supp_baseline}
\end{table}

\section{Implementation Details}

\noindent \textbf{Architecture.} Within the projection head of the Video Feature Alignment module, we employ SiLU activation and Group Normalization across all convolutional and MLP layers. For the interpolation layer, we utilize a temporal interpolation factor of 4, utilized alongside a skip connection with a kernel size of $(3, 1, 1)$ and 768 channels. Subsequently, the widths of the following MLP layers adhere to the sequence $768 \rightarrow 512 \rightarrow 256 \rightarrow 256$.

\vspace{0.2cm}

\noindent \textbf{Training Details.}
We apply LoRA~\cite{hu2022lora} exclusively to the attention modules, keeping all other backbone parameters frozen. For the LoRA parameters, we employ a linear warmup of 200 steps, gradually increasing the learning rate to a peak of $10^{-4}$. For the projector, we utilize a cosine decay scheduler with a 150-step warmup; the learning rate is initialized at $5 \times 10^{-4}$ and decays to a minimum of $1 \times 10^{-5}$. Additionally, we implement a gradient clipping threshold of 1.0.

\subsection{Evaluation Protocol and Metrics}

\noindent \textbf{Evaluation Subset Selection.} 
To ensure our evaluation aligns with the distribution of our training dataset (which focuses on articulated motion), we curated a subset of 85 prompts from the VBench-I2V~\cite{huang2023vbench} benchmark. The selected prompts predominantly feature human or animal subjects. We excluded generic scenery or abstract textures lacking distinct structural subjects, as these samples do not adequately challenge the model's structure-preserving capabilities.

\vspace{0.2cm}

\noindent \textbf{Metric Selection.} 
Since our primary objective is structure-preserving video generation rather than artistic creation, we exclude general visual quality metrics such as \textit{Aesthetic Quality} and \textit{Imaging Quality}, as they are less relevant to evaluating structural fidelity. To provide a unified quantitative evaluation, we formulate two composite scores: Motion Score  and Extended Motion Score.
Following the standard VBench-I2V protocol, we first apply min-max normalization to all individual sub-metrics to map them into a unified range. Let $\hat{s}_{\text{bg}}$, $\hat{s}_{\text{smooth}}$, and $\hat{s}_{\text{subj}}$ denote the normalized scores for Background Consistency, Motion Smoothness, and Subject Consistency, respectively. The Motion Score is defined as the arithmetic mean of these three core structural metrics:

\begin{equation}
    S_{\text{motion}} = \frac{\hat{s}_{\text{bg}} + \hat{s}_{\text{smooth}} + \hat{s}_{\text{subj}}}{3}
    \label{eq:motion_score}
\end{equation}

To further account for fidelity to the conditioning image, the Extended Motion Score incorporates I2V consistency metrics. We assign a weight of $0.5$ to both I2V Subject Consistency ($\hat{s}_{\text{i2v-s}}$) and I2V Background Consistency ($\hat{s}_{\text{i2v-b}}$), reflecting a balance between temporal coherence and input fidelity. The formulation is given by:

\begin{equation}
    S_{\text{ext}} = \frac{\hat{s}_{\text{bg}} + \hat{s}_{\text{smooth}} + \hat{s}_{\text{subj}} + 0.5 \cdot \hat{s}_{\text{i2v-s}} + 0.5 \cdot \hat{s}_{\text{i2v-b}}}{4}
    \label{eq:ext_motion_score}
\end{equation}

\section {Additional Experiments}
\noindent \textbf{Dense Features Outperform Sparse Trajectories.} To ensure a fair comparison within the DiT-based CogVideoX~\cite{yang2024cogvideox} architecture, we align the experimental implementation by adapting the Track4Gen~\cite{jeong2024track4gen} projector to mirror our design: specifically, employing an interpolation layer followed by three MLP layers. Furthermore, we exclude the refiner module to strictly isolate the efficacy of the distillation objective. As shown in Table~\ref{tab:supp_baseline}, our method outperforms $\text{Track4Gen}^*$. This demonstrates the superiority of dense SAM2~\cite{ravi2024sam} features over sparse point trajectories, as the latter are prone to performance degradation caused by error accumulation in optical flow estimation (e.g., RAFT~\cite{teed2020raft}).

\vspace{0.2cm}

\noindent \textbf{Explicit Backward Constraints are Redundant.} We investigate the impact of extending the Local Gram Feature (LGF) loss to incorporate a backward temporal constraint (i.e., computing similarity between frame $t$ and $t-1$). As shown in Table~\ref{tab:supp_ablations}, this formulation yields inferior results compared to our forward-only design. We attribute this to the inherent bidirectionality of the distilled SAM 2 features; explicitly enforcing backward consistency creates computational redundancy and introduces over-constraints that hamper the generation dynamics. Therefore, we retain the forward-only formulation.

\begin{table}   
  \centering
      \begin{tabular}{lcc}
          \toprule
          Configurations & Motion Score$\uparrow$ & Ext. Motion Score$\uparrow$ \\
          \midrule
          Bidirectional LGF & 94.87 & 95.52 \\
          \textbf{LGF loss (Ours)} & \textbf{95.51} & \textbf{96.03} \\
          \midrule
          23rd Transformer Block & 94.93 & 95.54 \\
          24th Transformer Block & 94.44 & 95.14 \\
          \textbf{25th Transformer Block (Ours)} & \textbf{95.51} & \textbf{96.03} \\
           26th Transformer Block & 94.44 & 95.12 \\
           27th Transformer Block & 94.68 & 95.36 \\
          \bottomrule       
      \end{tabular}
    \caption{Ablations on Temporal Directionality and Injection Depth. We analyze two key design choices: (a) the directionality of the Local Gram Feature loss, and (b) the optimal depth for feature injection within the DiT architecture. The highlighted row indicates our default configuration.}
  \label{tab:supp_ablations}
\end{table}

\section{Additional Qualitative Results}
We provide additional visual comparisons to further substantiate the effectiveness of \model{}. Specifically, we contrast our method with competing baselines and state-of-the-art models, highlighting superior structural fidelity and temporal coherence across diverse motion scenarios.

\begin{figure*}[t]
\centering
\setlength{\fboxsep}{0pt}   
\setlength{\fboxrule}{1pt}  
\def\imW{0.18\linewidth}

\begin{tabular}{c@{\hskip 1pt}c@{\hskip 0pt}c@{\hskip 0pt}c@{\hskip 0pt}c@{\hskip 0pt}c}
    \rotatebox{90}{\small Base model} 
   &\fcolorbox{blue}{white}{\includegraphics[width=\imW]   {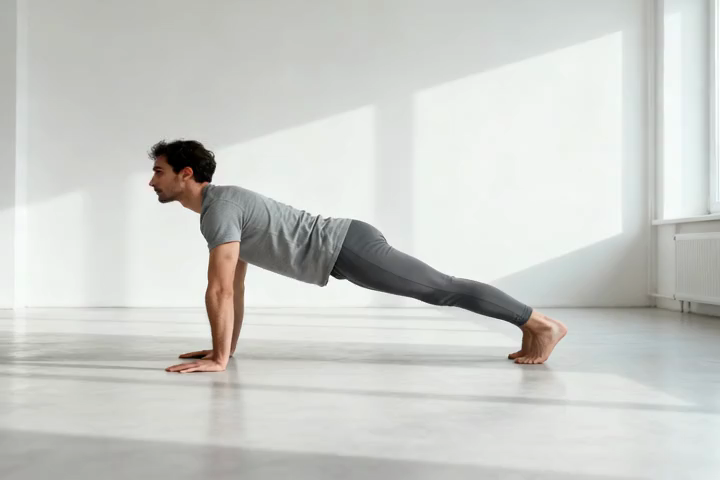}} &
    \includegraphics[width=\imW]{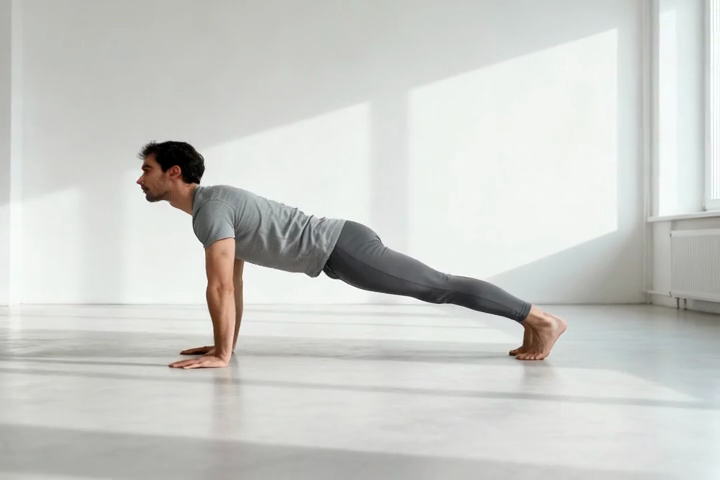} &
    \includegraphics[width=\imW]{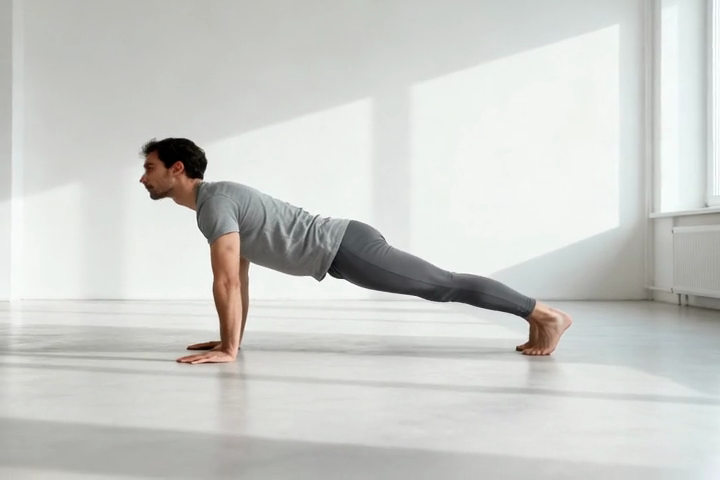} &
    \includegraphics[width=\imW]{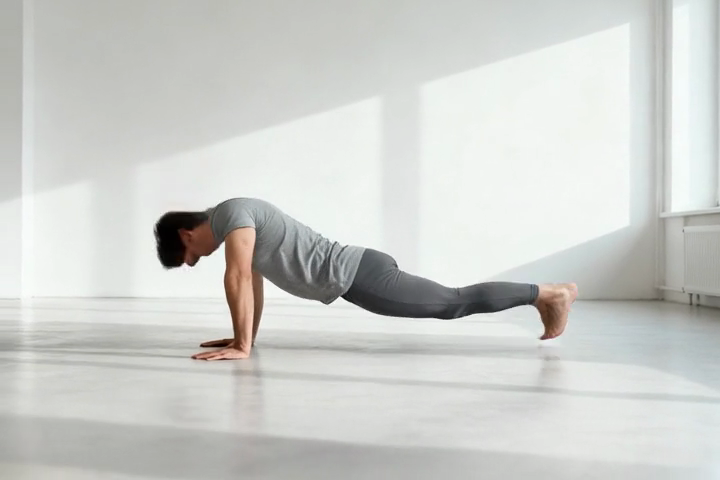} &
    \includegraphics[width=\imW]{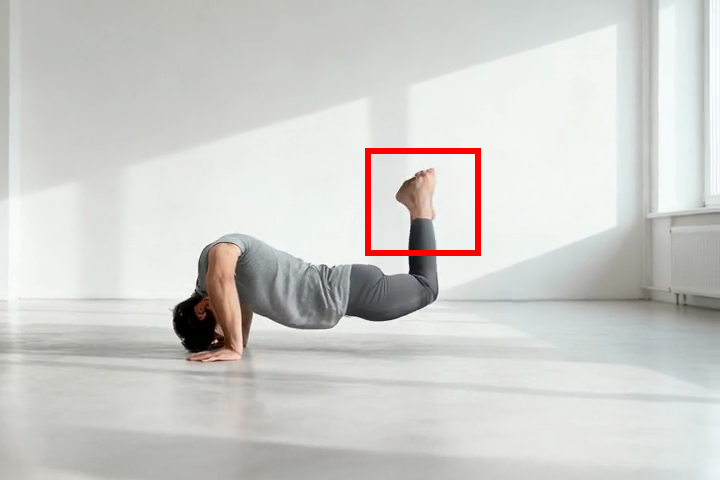} \\

    \rotatebox{90}{\small + Fine-tuning} 
   &\includegraphics[width=\imW]{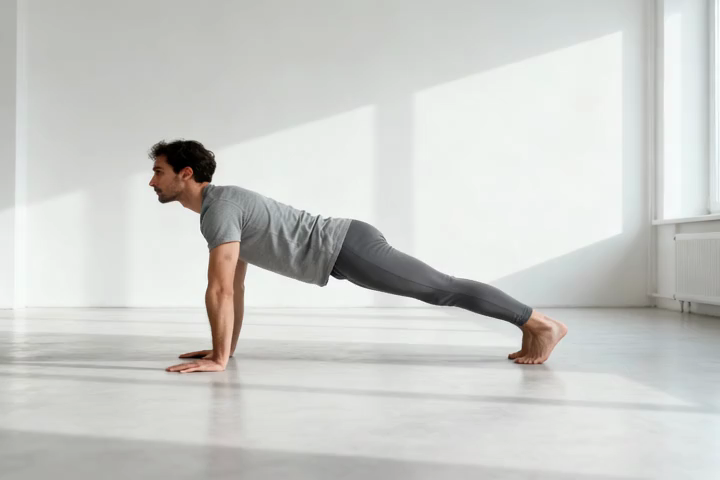} &
    \includegraphics[width=\imW]{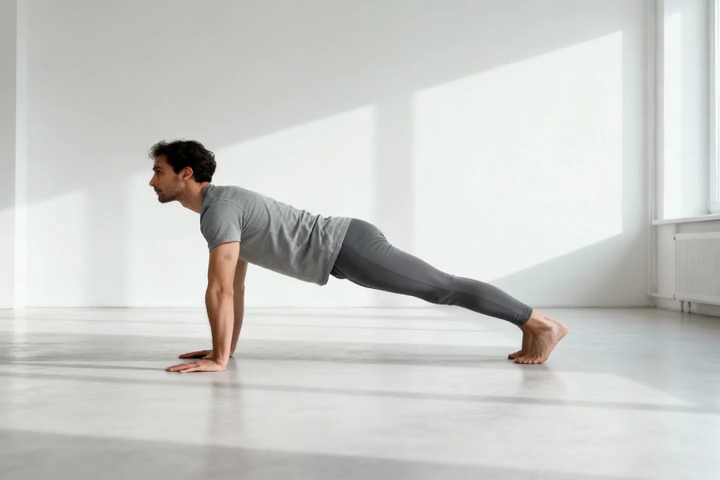} &
    \includegraphics[width=\imW]{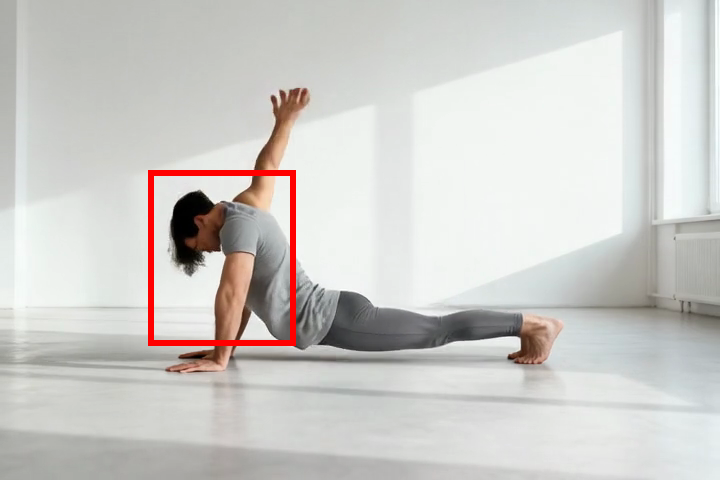} &
    \includegraphics[width=\imW]{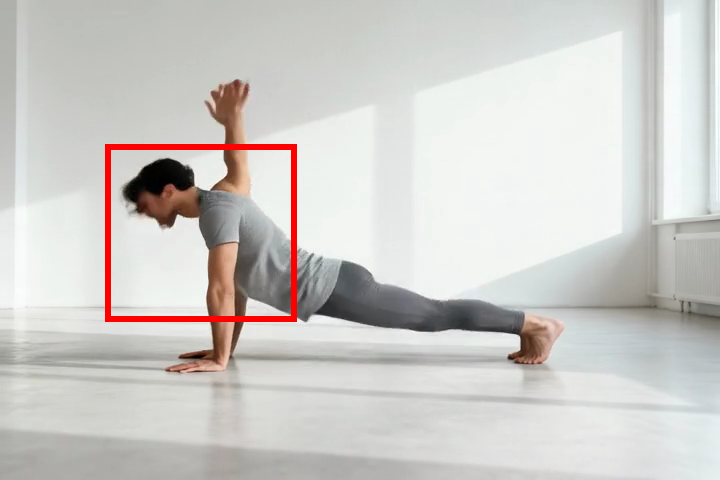} &
    \includegraphics[width=\imW]{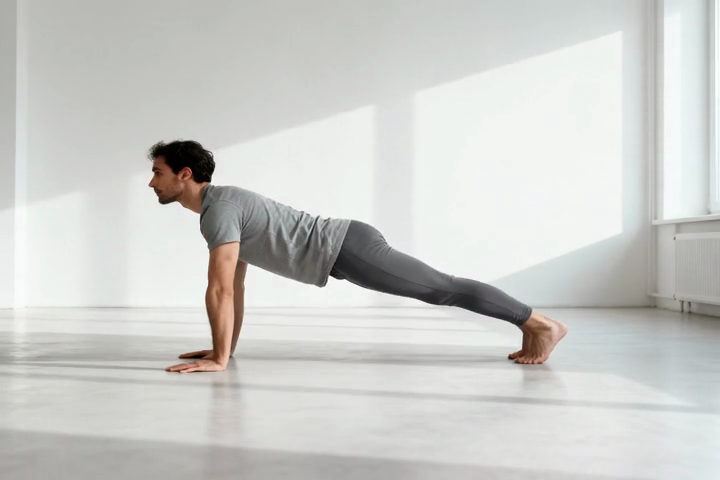} \\

    \rotatebox{90}{\small + Mask sup.} 
    &\includegraphics[width=\imW]{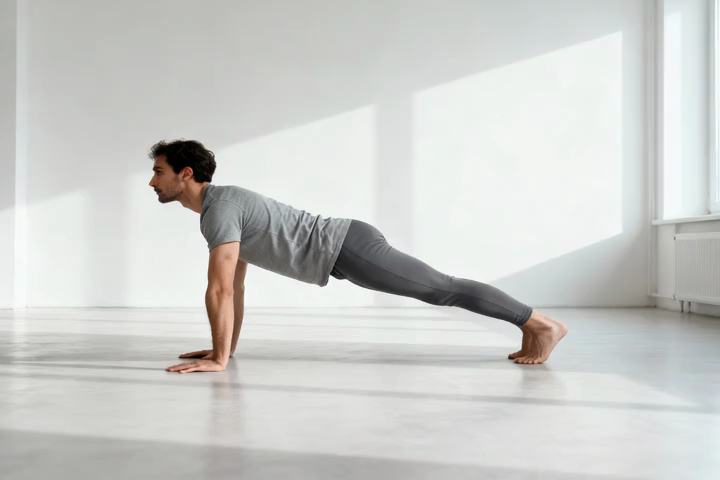} &
    \includegraphics[width=\imW]{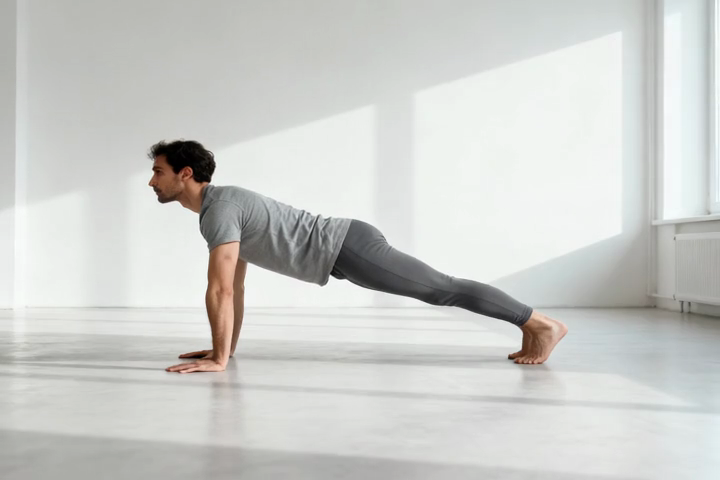} &
    \includegraphics[width=\imW]{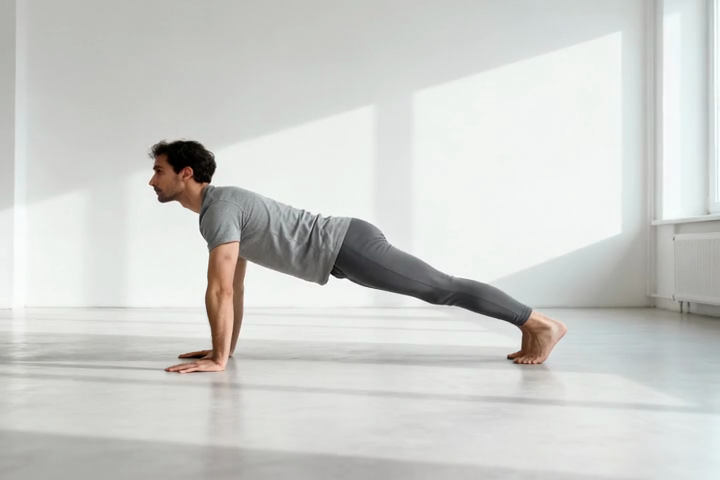} &
    \includegraphics[width=\imW]{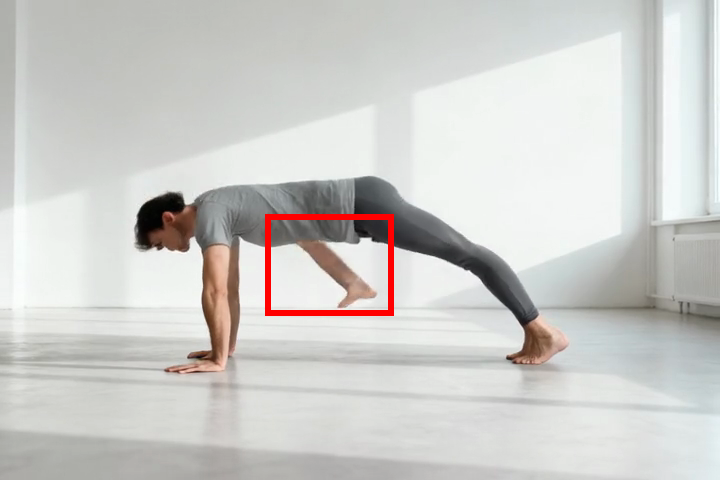} &
    \includegraphics[width=\imW]{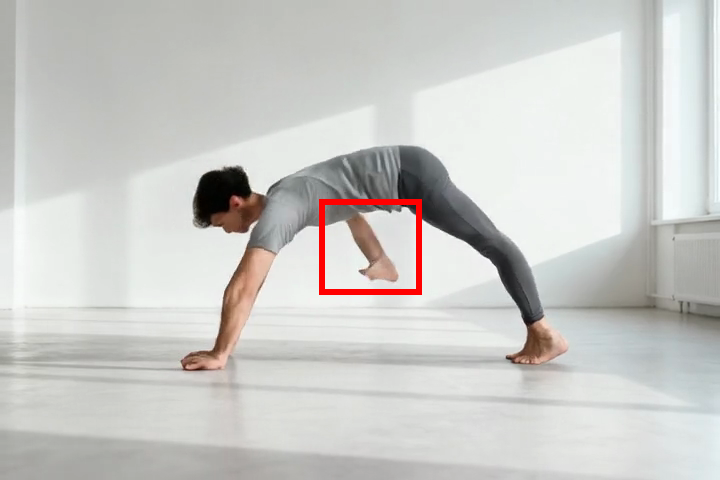} \\

    \rotatebox{90}{\small \textbf{\model{}}} 
     &\includegraphics[width=\imW]{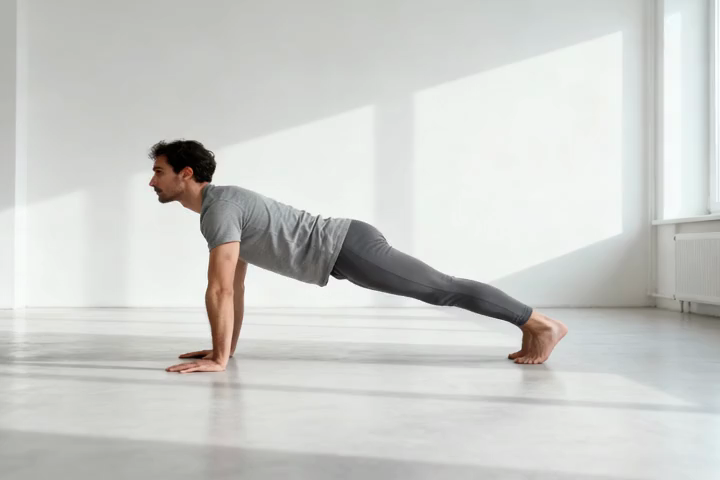} &
    \includegraphics[width=\imW]{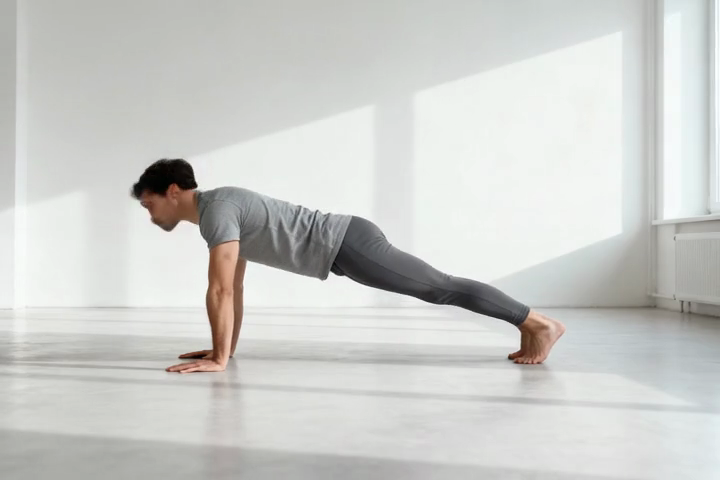} &
    \includegraphics[width=\imW]{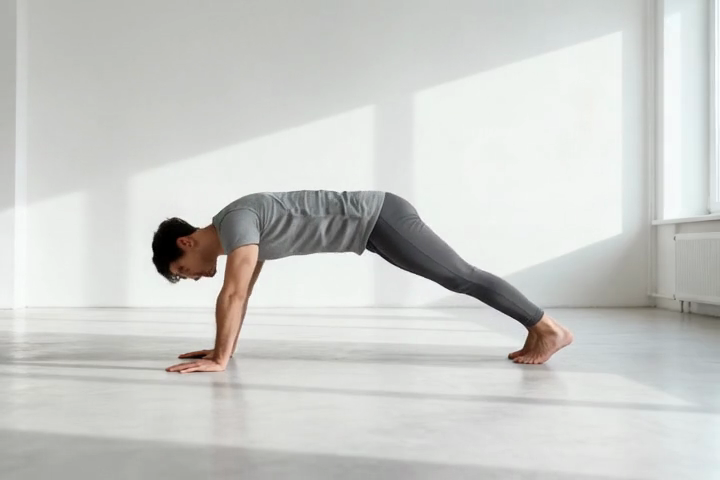} &
    \includegraphics[width=\imW]{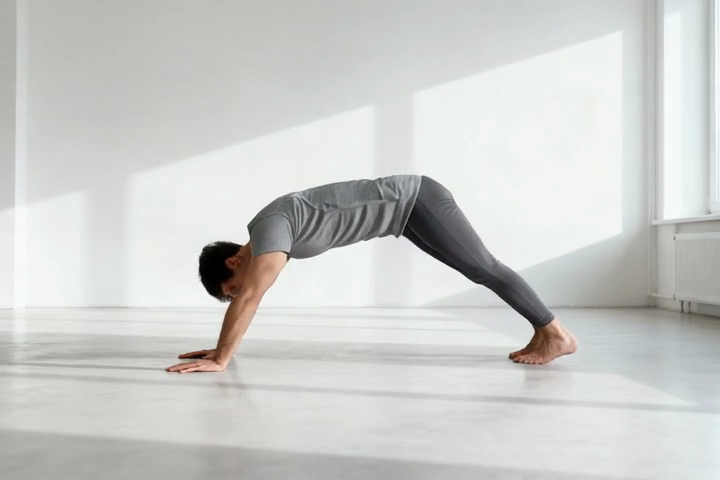} &
    \includegraphics[width=\imW]{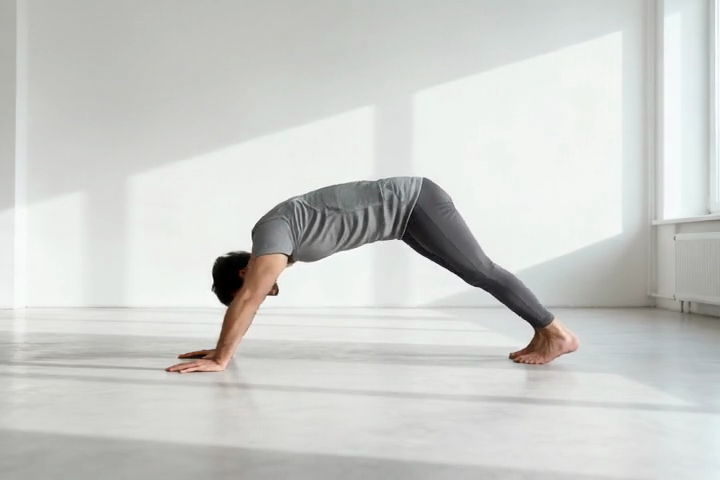} \\

    \rotatebox{90}{\small Base model}  &
     \fcolorbox{blue}{white}{\includegraphics[width=\imW]{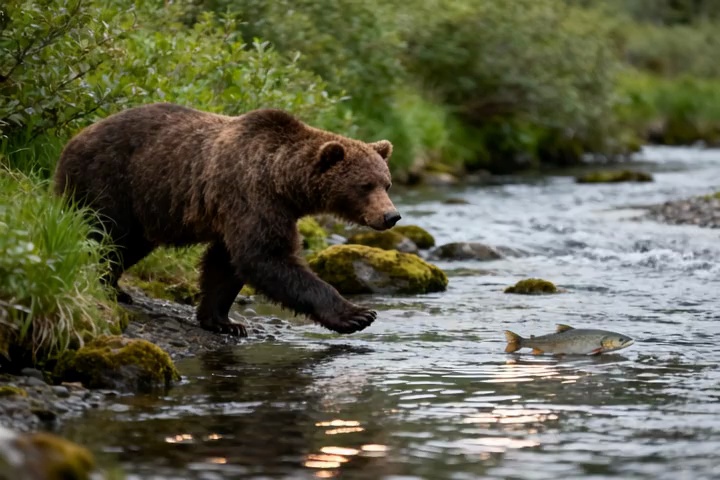}} &
    \includegraphics[width=\imW]{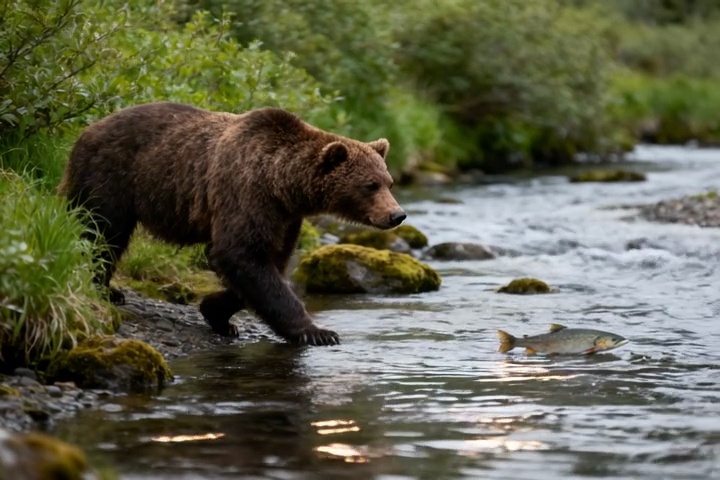} &
    \includegraphics[width=\imW]{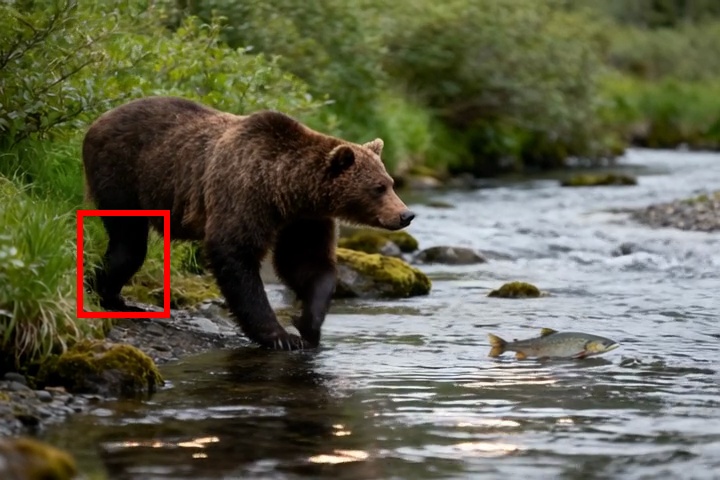} &
    \includegraphics[width=\imW]{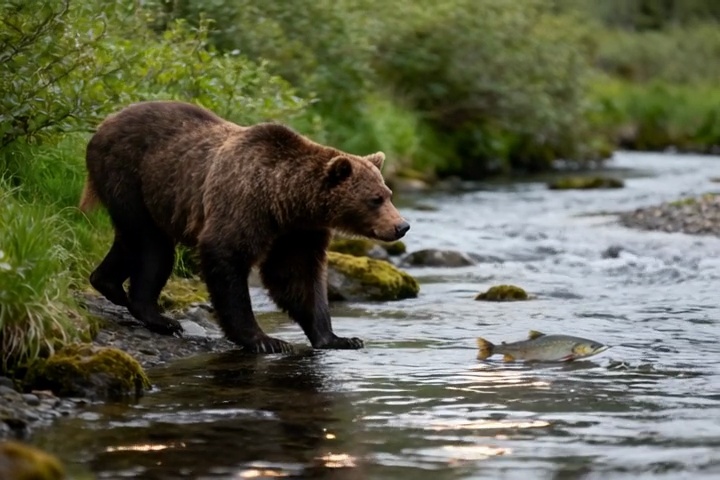} &
    \includegraphics[width=\imW]{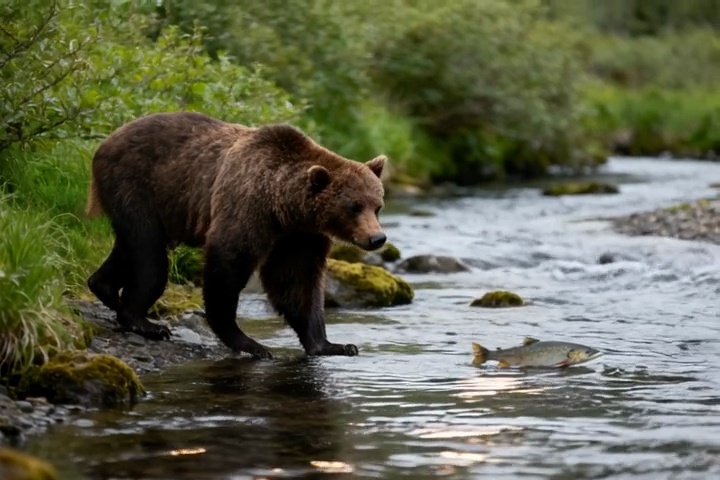} \\

      \rotatebox{90}{\small + Fine-tuning}  &
   \includegraphics[width=\imW]{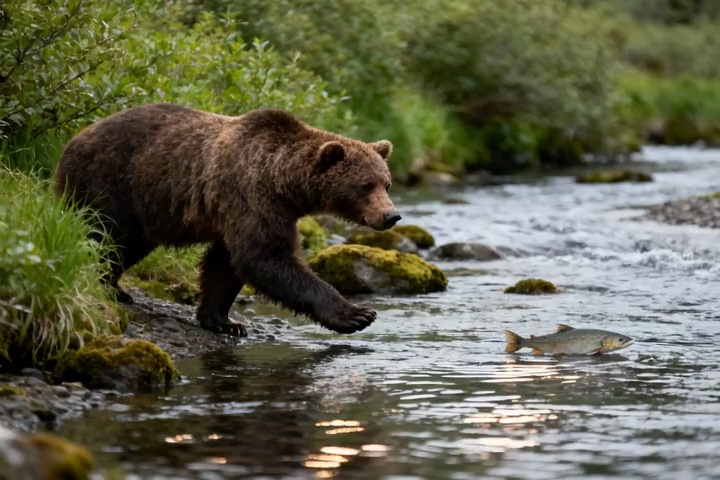} &
    \includegraphics[width=\imW]{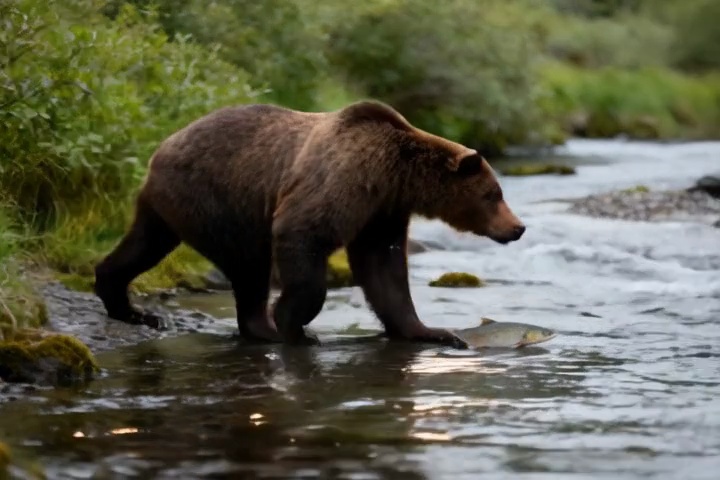} &
    \includegraphics[width=\imW]{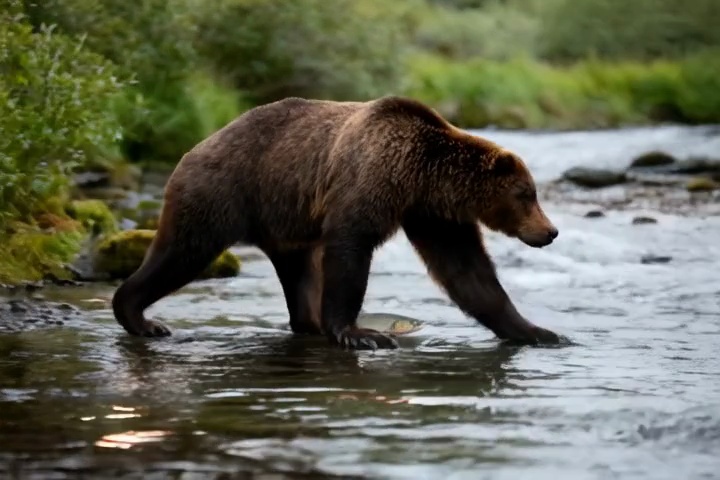} &
    \includegraphics[width=\imW]{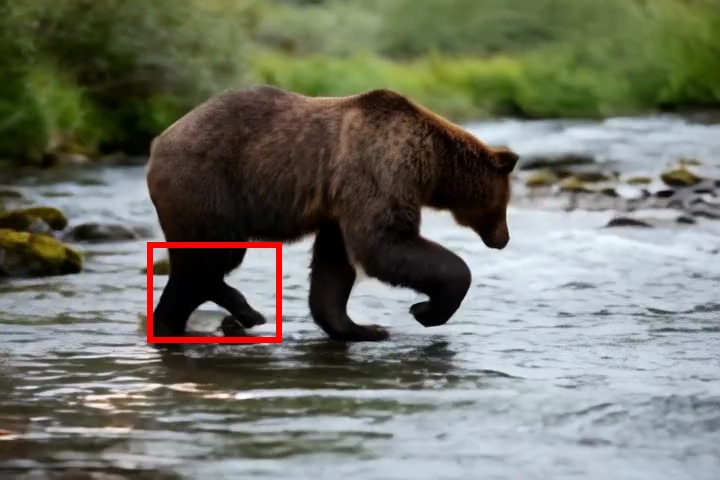} &
    \includegraphics[width=\imW]{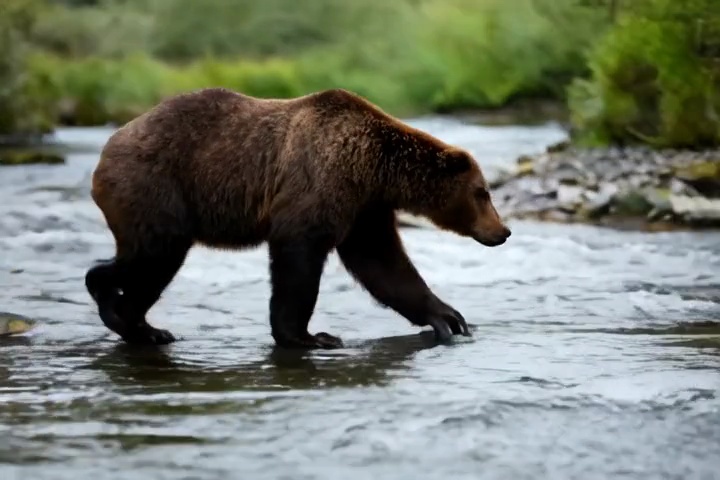} \\

     \rotatebox{90}{\small + Mask sup.}  &
    \includegraphics[width=\imW]{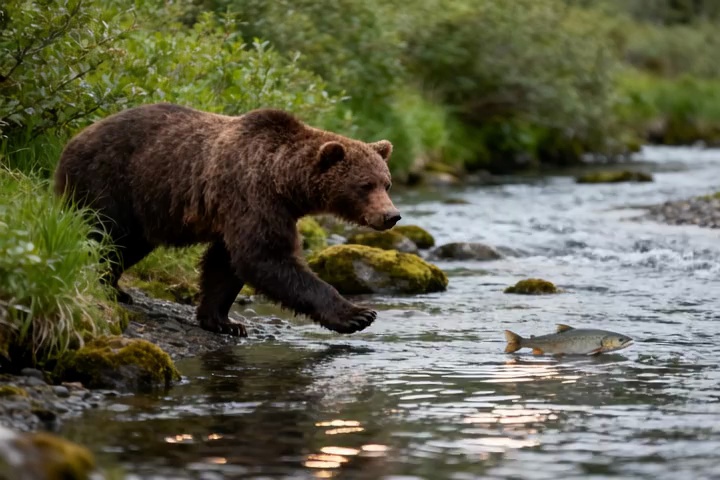} &
    \includegraphics[width=\imW]{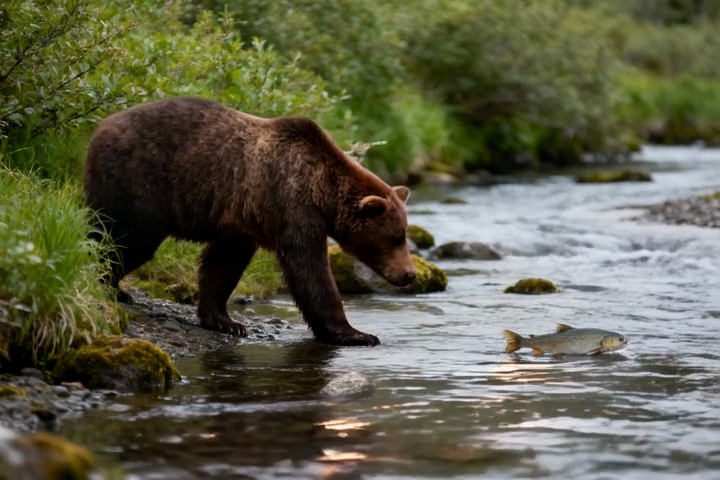} &
    \includegraphics[width=\imW]{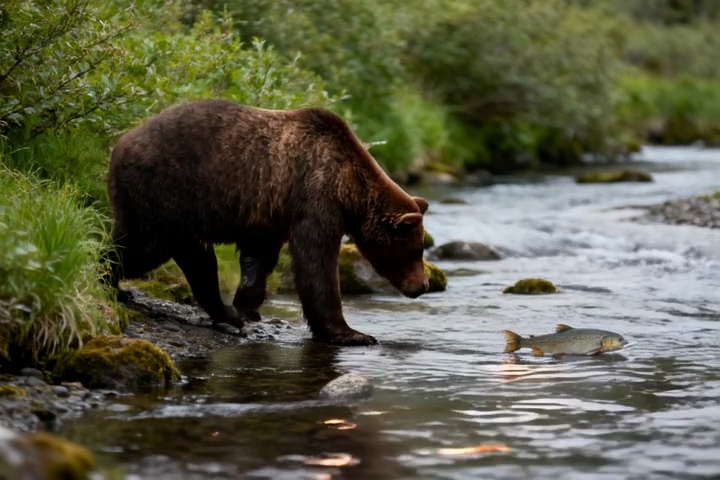} &
    \includegraphics[width=\imW]{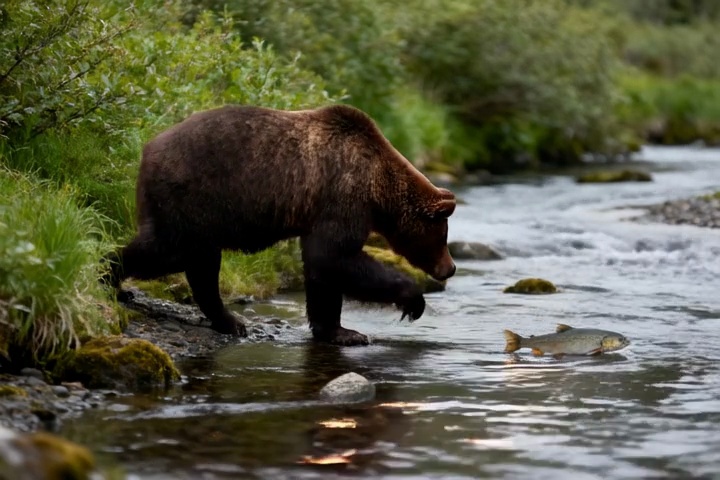} &
    \includegraphics[width=\imW]{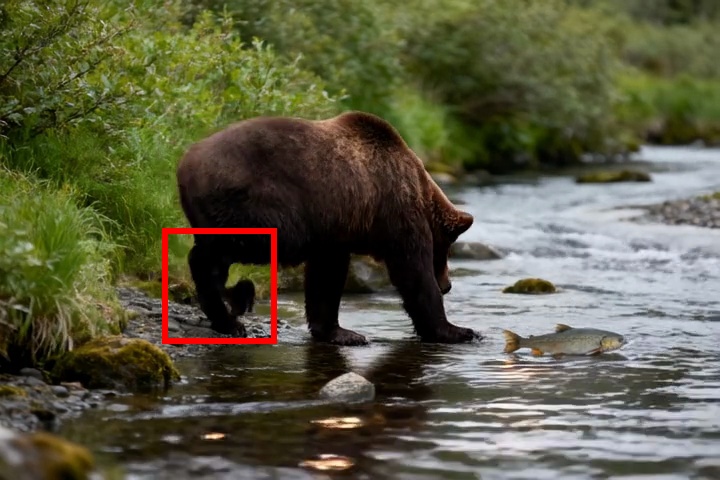} \\

    \rotatebox{90}{\small \textbf{\model{}}} &
    \includegraphics[width=\imW]{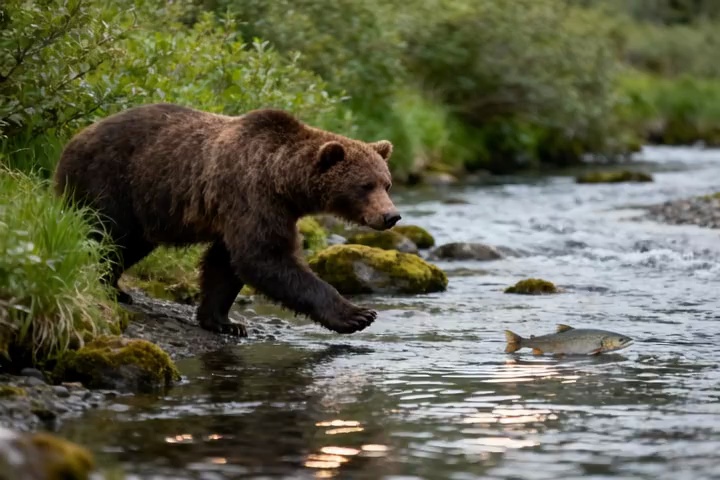} &
    \includegraphics[width=\imW]{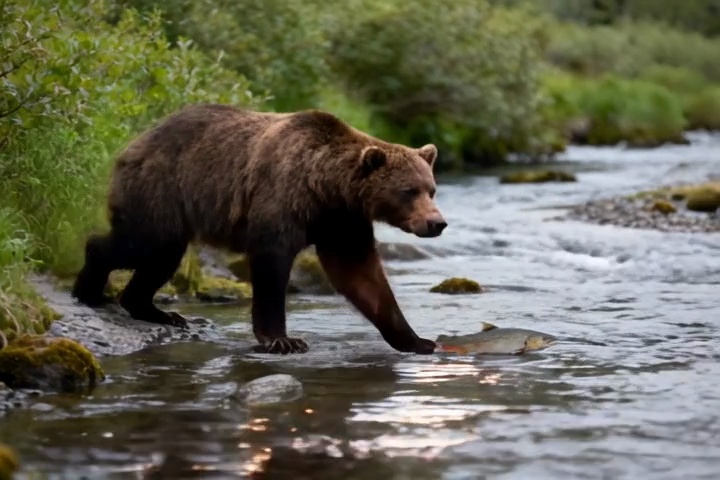} &
    \includegraphics[width=\imW]{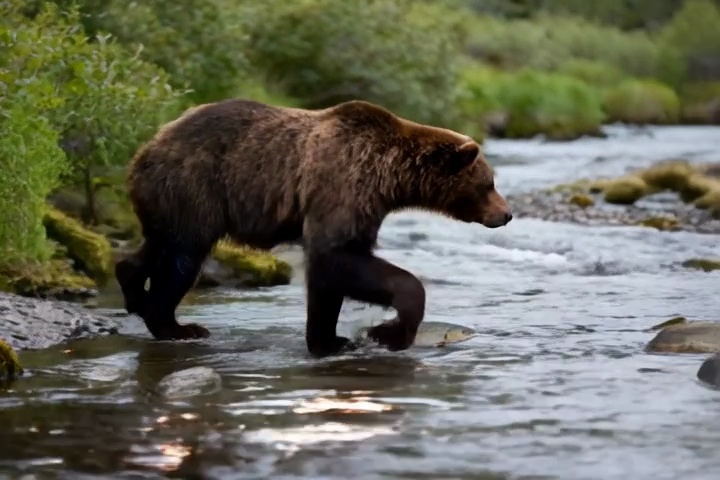} &
    \includegraphics[width=\imW]{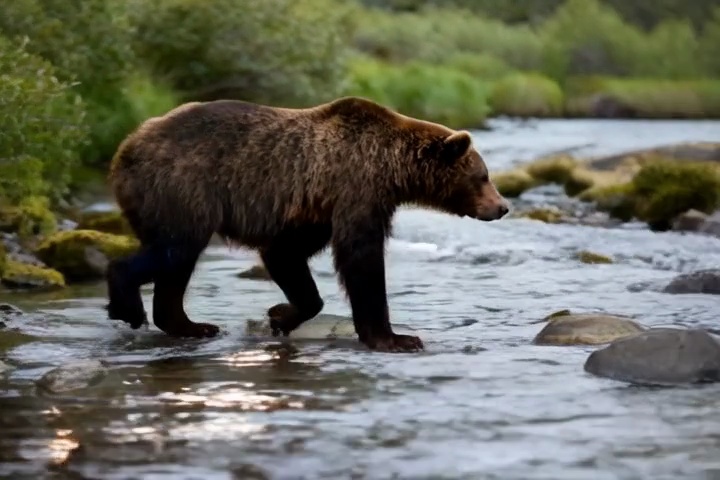} &
    \includegraphics[width=\imW]{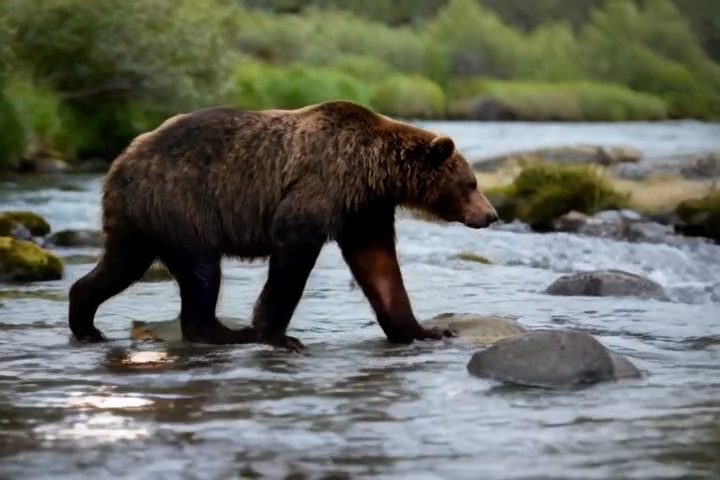} \\
\end{tabular}
\caption{\textbf{Extended Visual Comparisons with Baselines. } The blue box indicates the conditioning input image, while red boxes highlight structural failure modes observed in baseline methods.}
\end{figure*}

\begin{figure*}[t]
\centering
\setlength{\fboxsep}{0pt}   
\setlength{\fboxrule}{1pt}  
\def\imW{0.18\linewidth}

\begin{tabular}{c@{\hskip 1pt}c@{\hskip 0pt}c@{\hskip 0pt}c@{\hskip 0pt}c@{\hskip 0pt}c}
    \rotatebox{90}{\small CogVideoX} 
   &\fcolorbox{blue}{white}{\includegraphics[width=\imW]   {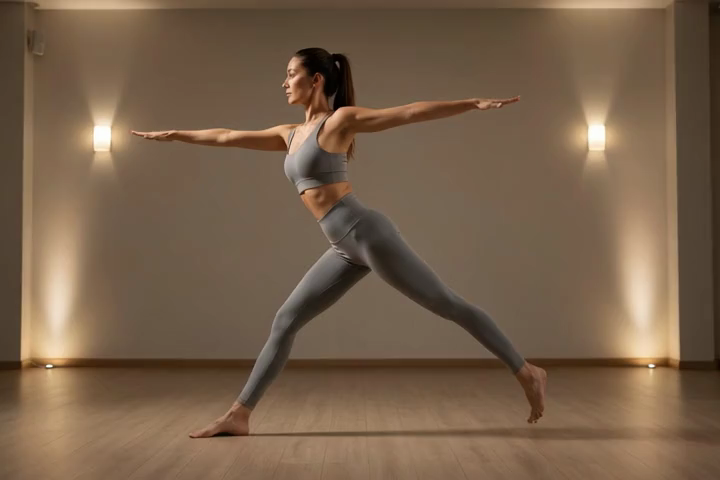}} &
    \includegraphics[width=\imW]{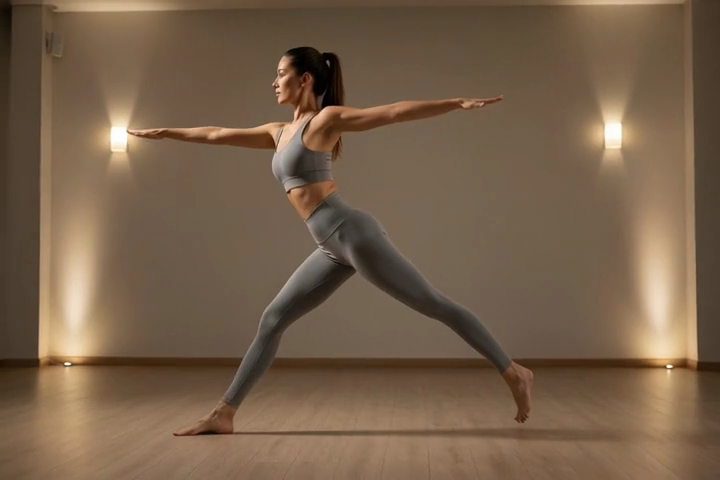} &
    \includegraphics[width=\imW]{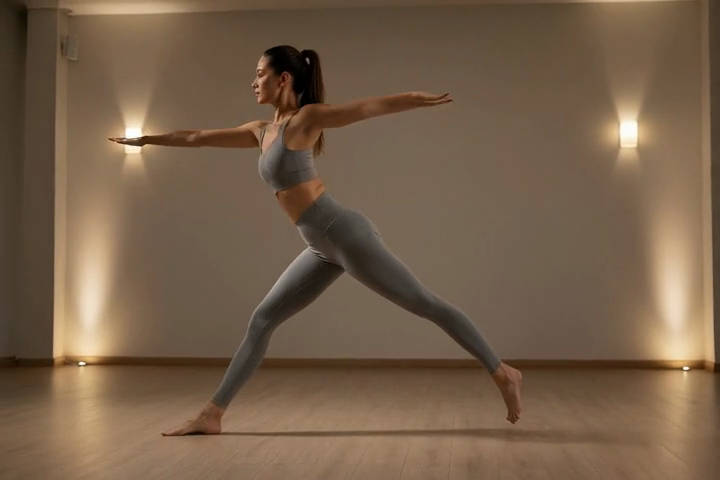} &
    \includegraphics[width=\imW]{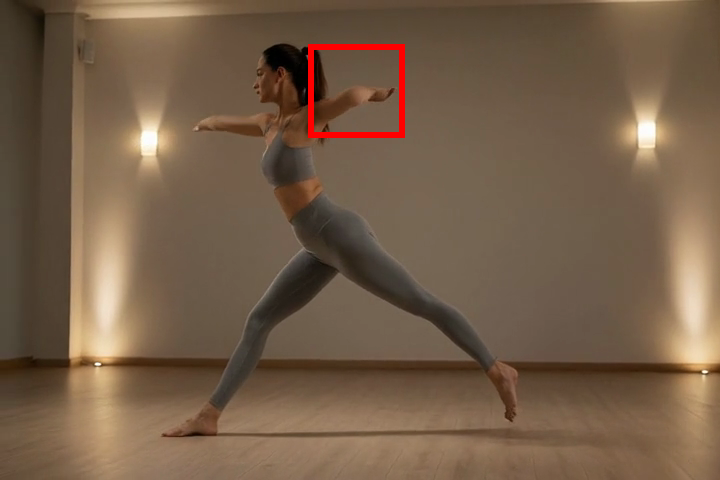} &
    \includegraphics[width=\imW]{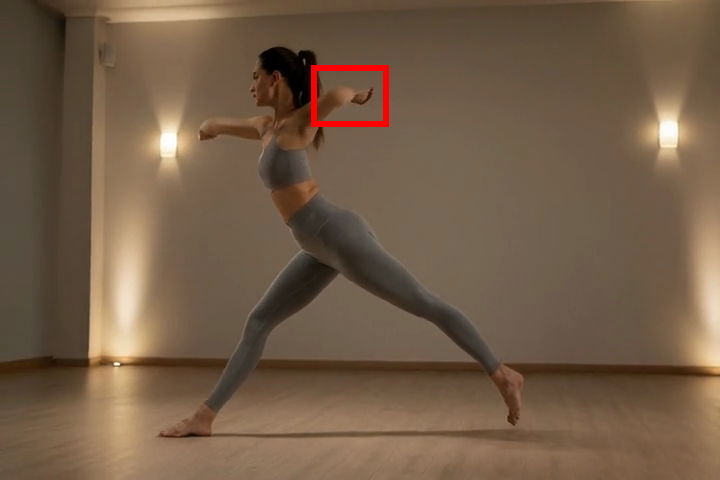} \\

    \rotatebox{90}{\small HunyuanVid} 
   &\includegraphics[width=\imW]{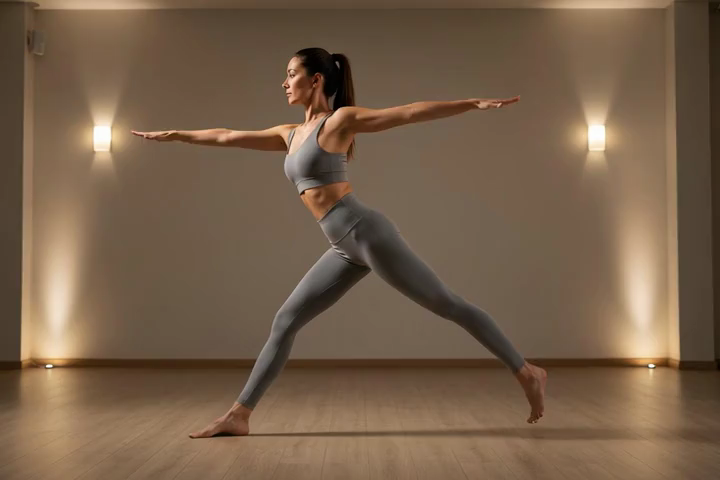} &
    \includegraphics[width=\imW]{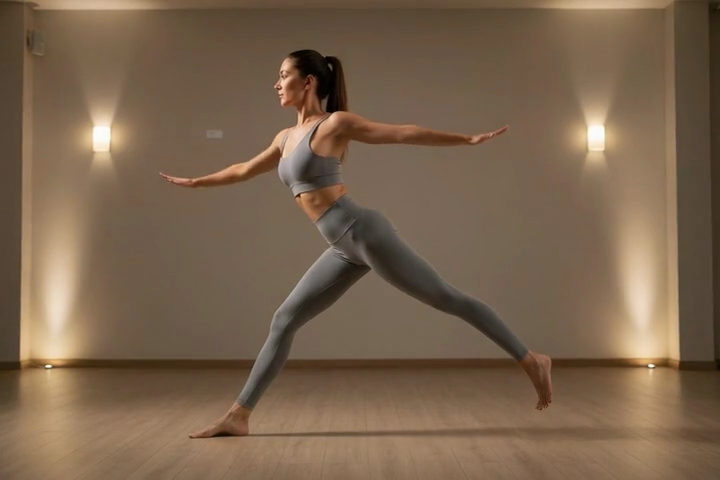} &
    \includegraphics[width=\imW]{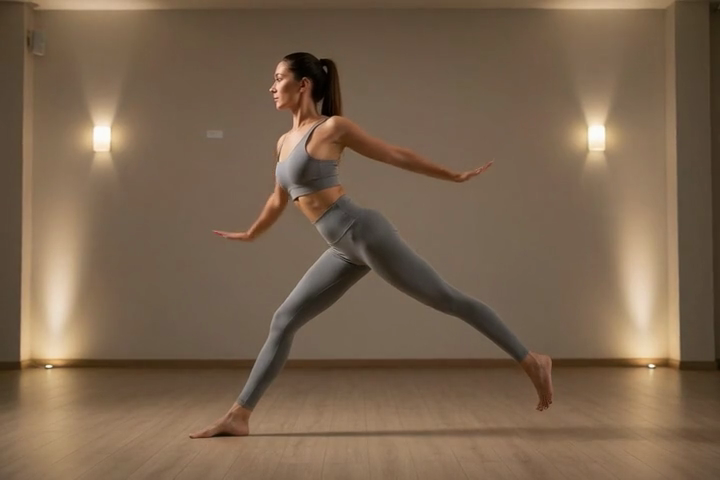} &
    \includegraphics[width=\imW]{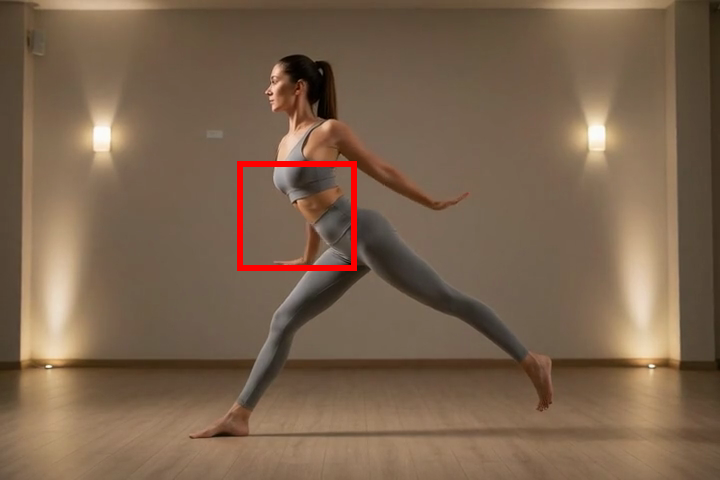} &
    \includegraphics[width=\imW]{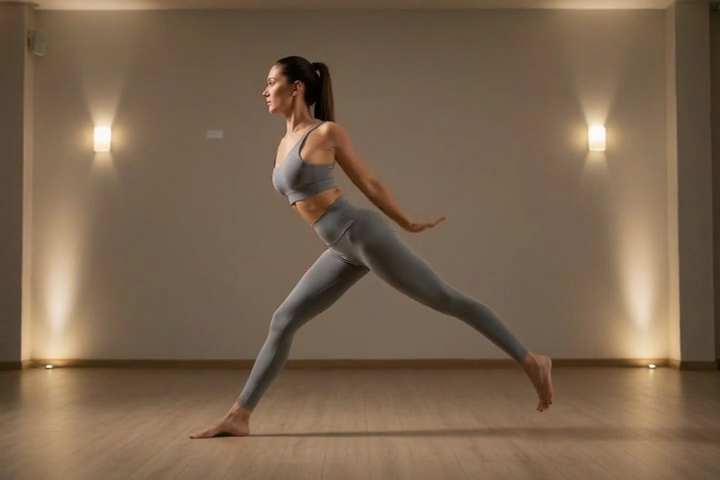} \\

    \rotatebox{90}{\small \textbf{\model{}}} 
    &\includegraphics[width=\imW]{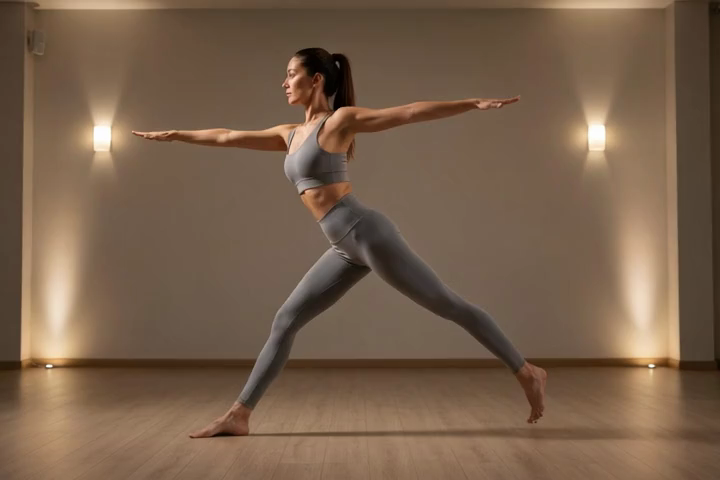} &
    \includegraphics[width=\imW]{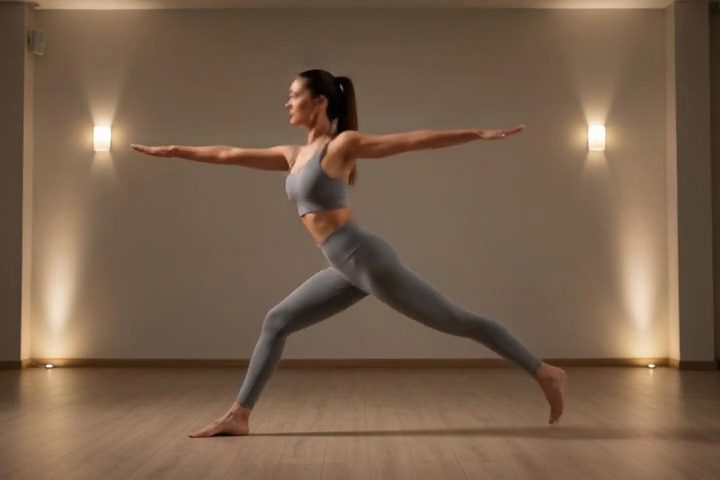} &
    \includegraphics[width=\imW]{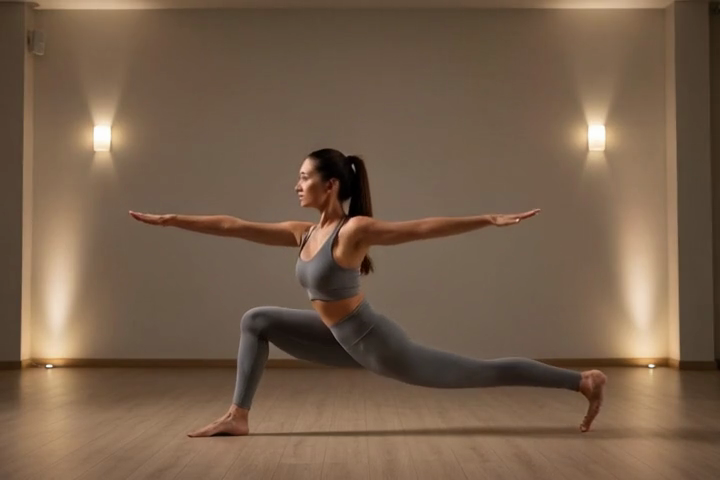} &
    \includegraphics[width=\imW]{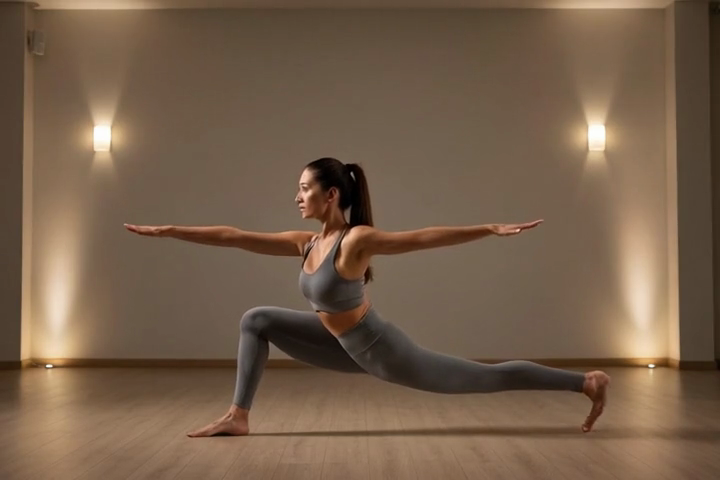} &
    \includegraphics[width=\imW]{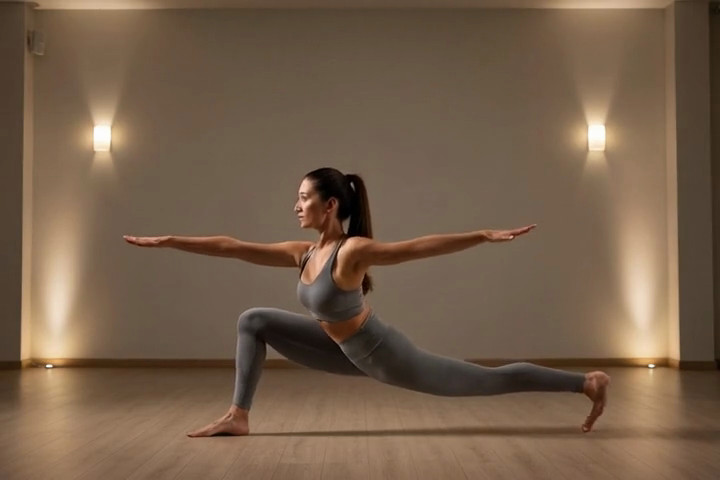} \\

    \rotatebox{90}{\small CogVideoX}  &
     \fcolorbox{blue}{white}{\includegraphics[width=\imW]{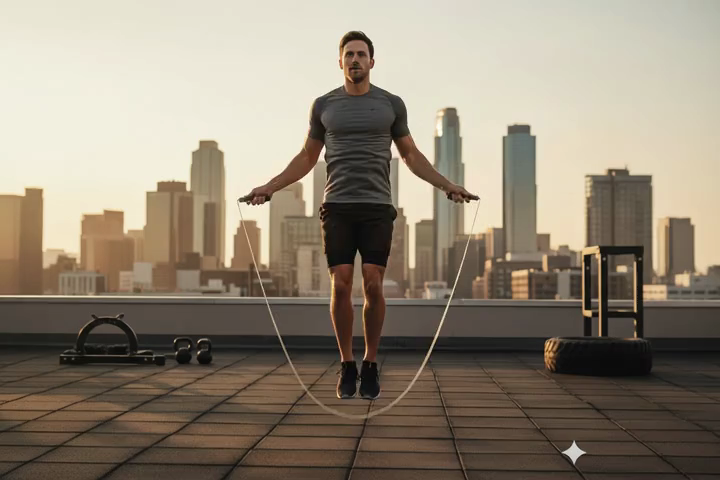}} &
    \includegraphics[width=\imW]{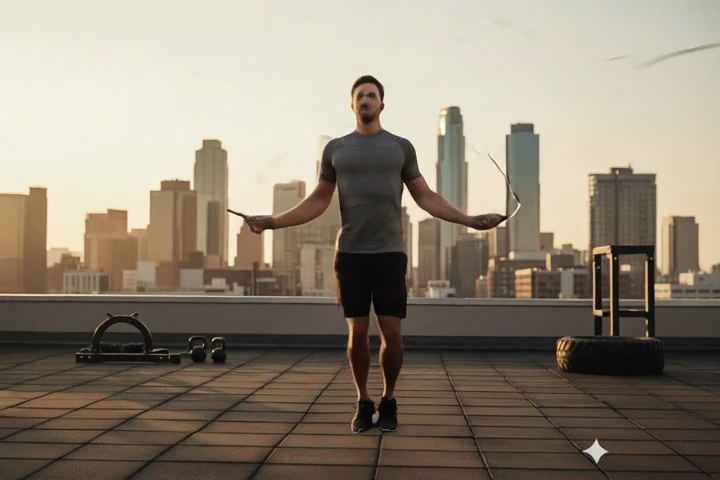} &
    \includegraphics[width=\imW]{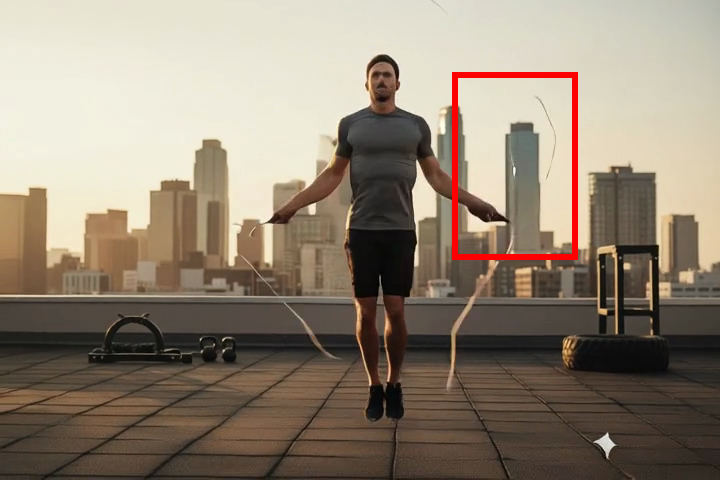} &
    \includegraphics[width=\imW]{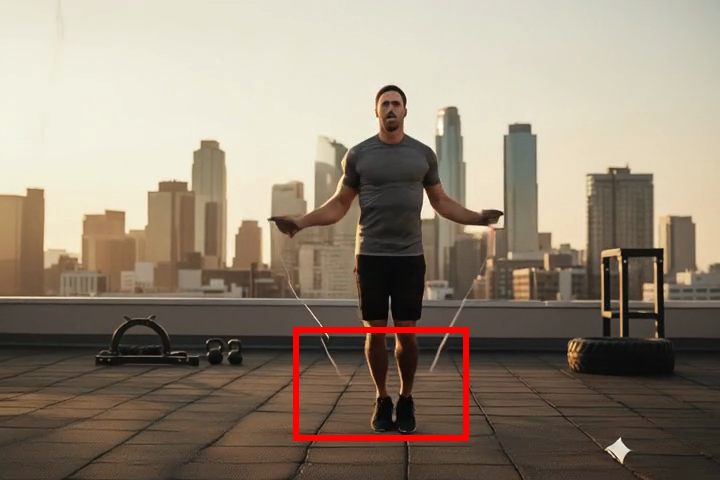} &
    \includegraphics[width=\imW]{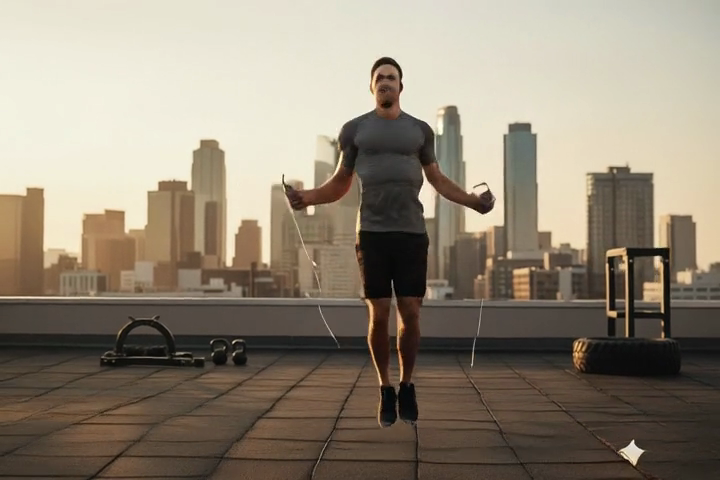} \\

      \rotatebox{90}{\small HunyuanVid}  &
   \includegraphics[width=\imW]{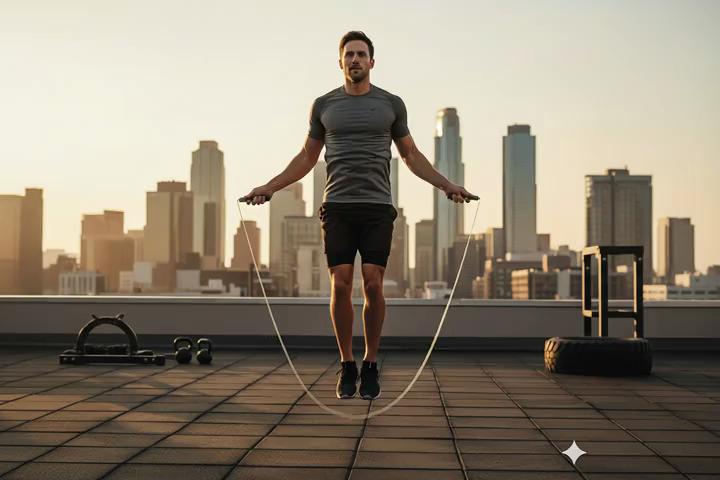} &
    \includegraphics[width=\imW]{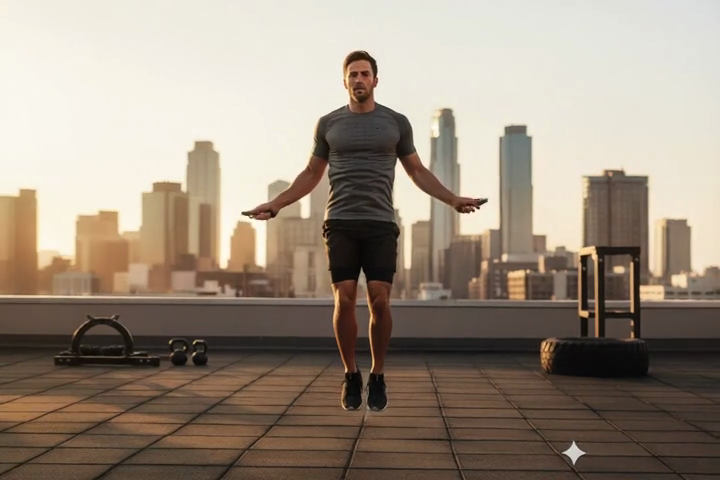} &
    \includegraphics[width=\imW]{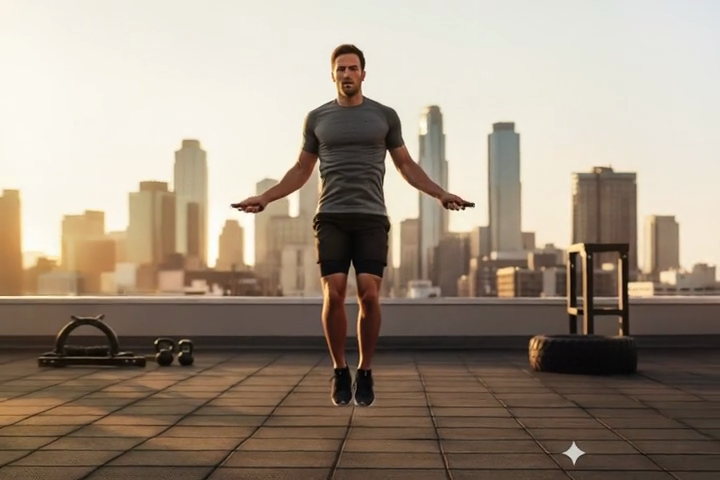} &
    \includegraphics[width=\imW]{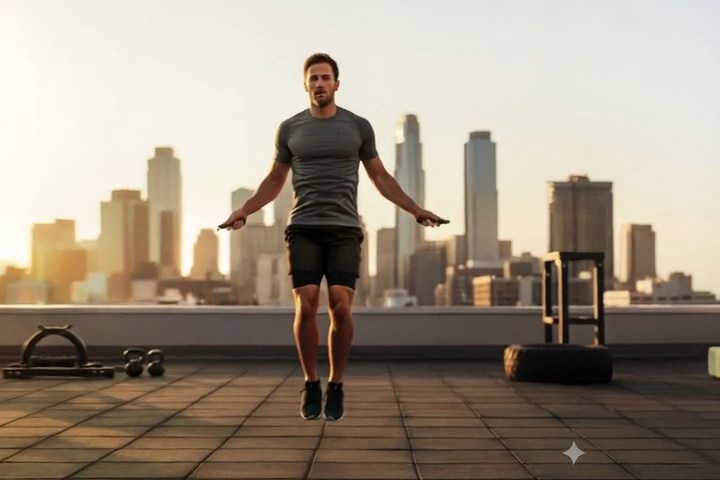} &
    \includegraphics[width=\imW]{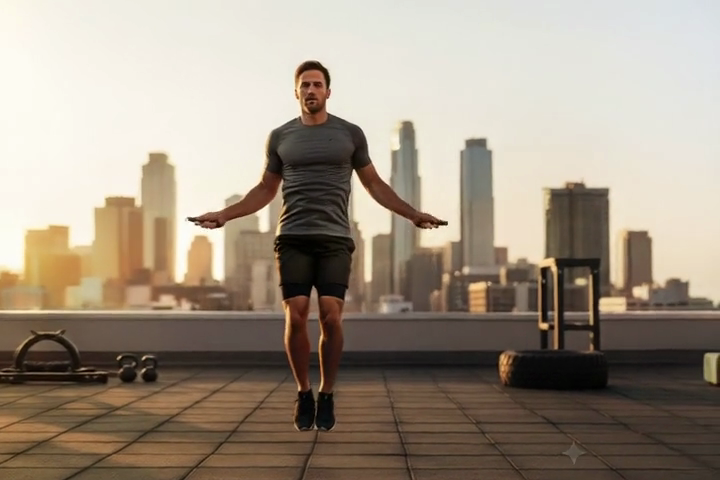} \\

     \rotatebox{90}{\small \textbf{\model{}}}  &
    \includegraphics[width=\imW]{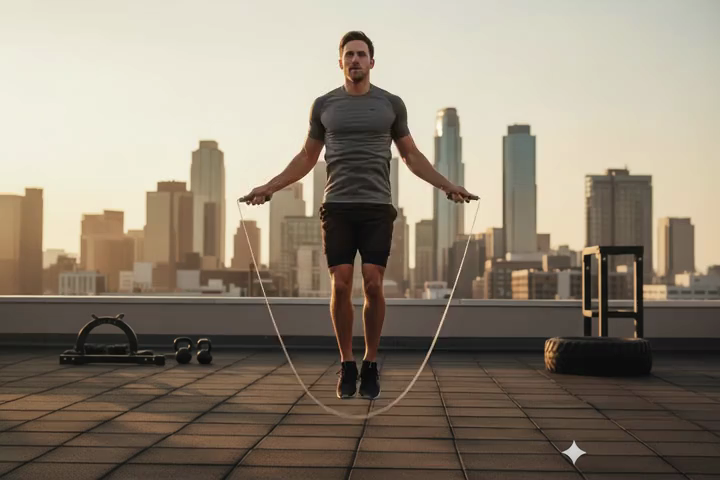} &
    \includegraphics[width=\imW]{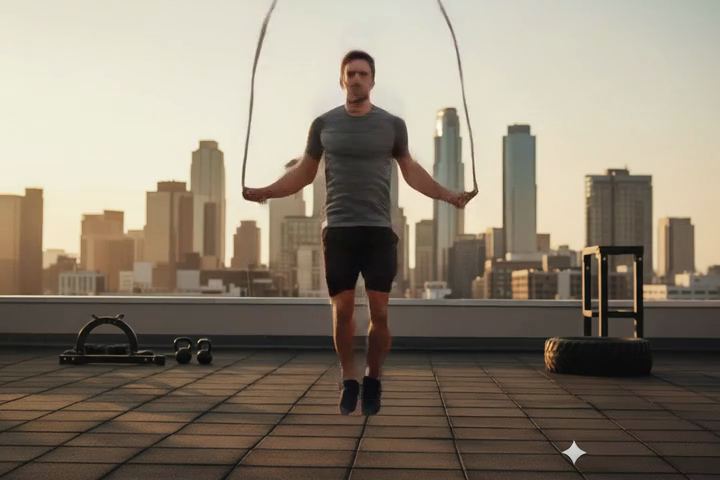} &
    \includegraphics[width=\imW]{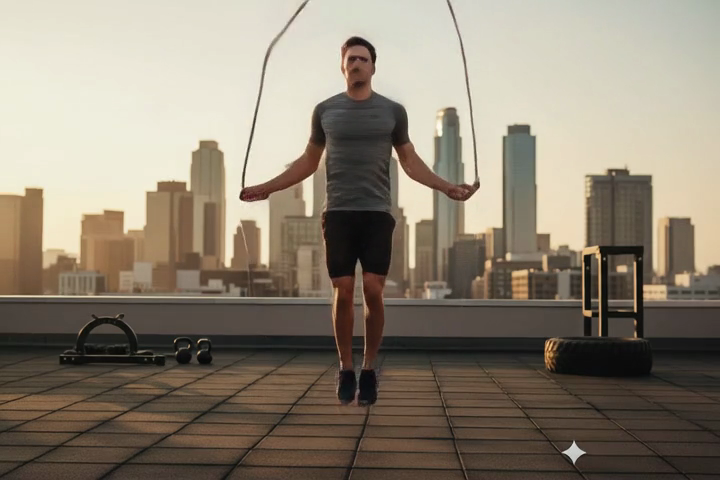} &
    \includegraphics[width=\imW]{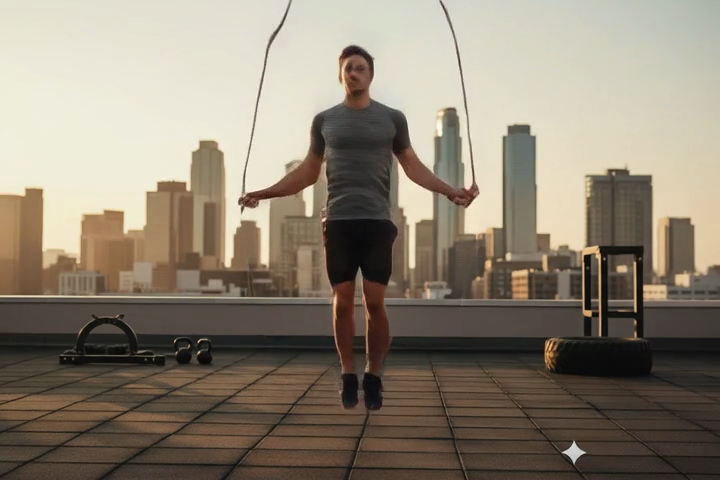} &
    \includegraphics[width=\imW]{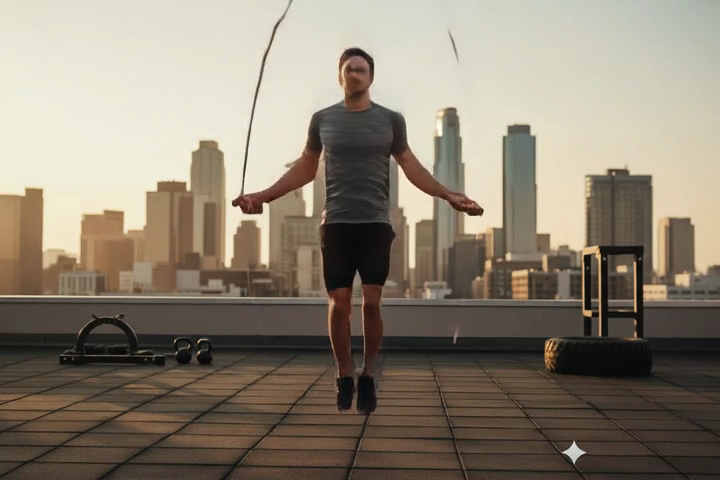} \\

\end{tabular}
\caption{\textbf{Visual Comparison with State-of-the-Art Open-Source Models.} Notably, despite HunyuanVid~\cite{li2024hunyuan} possessing 13B parameters, our \model{} (5B) achieves comparable generation quality.} 
\end{figure*}

\begin{figure*}[t]
\centering
\setlength{\fboxsep}{0pt}   
\setlength{\fboxrule}{1pt}  
\def\imW{0.18\linewidth}

\begin{tabular}{c@{\hskip 1pt}c@{\hskip 0pt}c@{\hskip 0pt}c@{\hskip 0pt}c@{\hskip 0pt}c}
    \includegraphics[width=\imW]{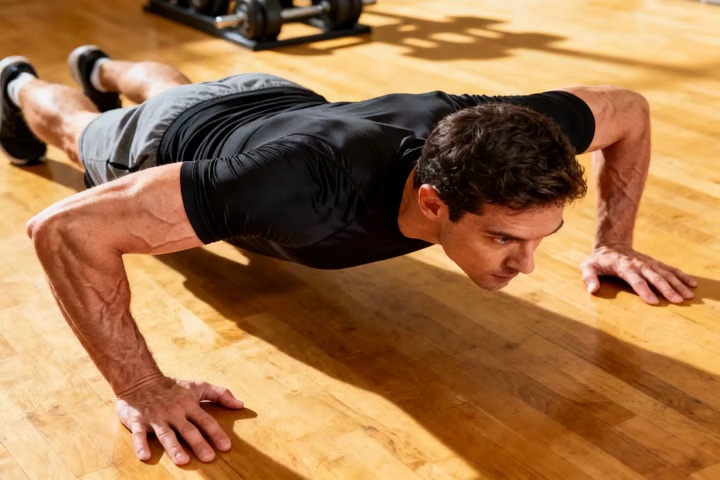} &
    \includegraphics[width=\imW]{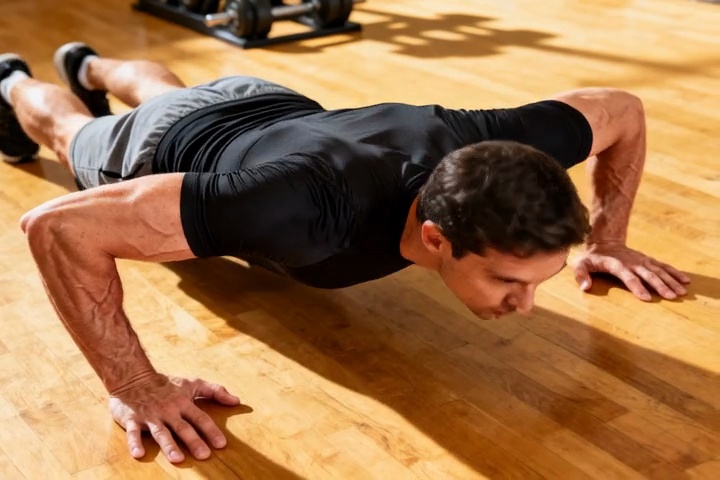} &
    \includegraphics[width=\imW]{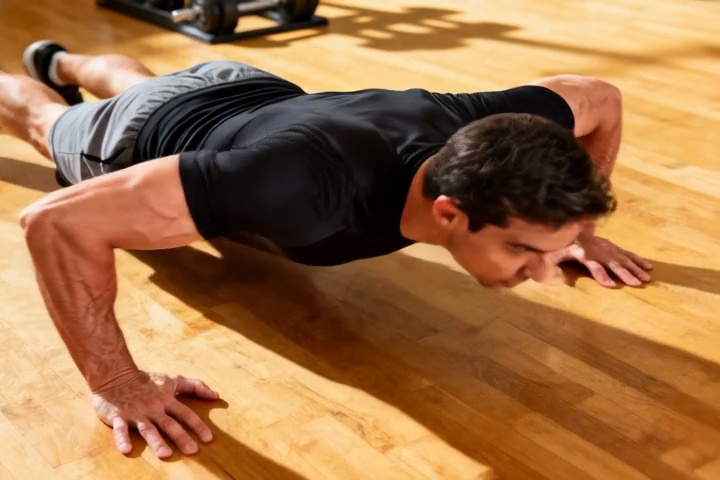} &
    \includegraphics[width=\imW]{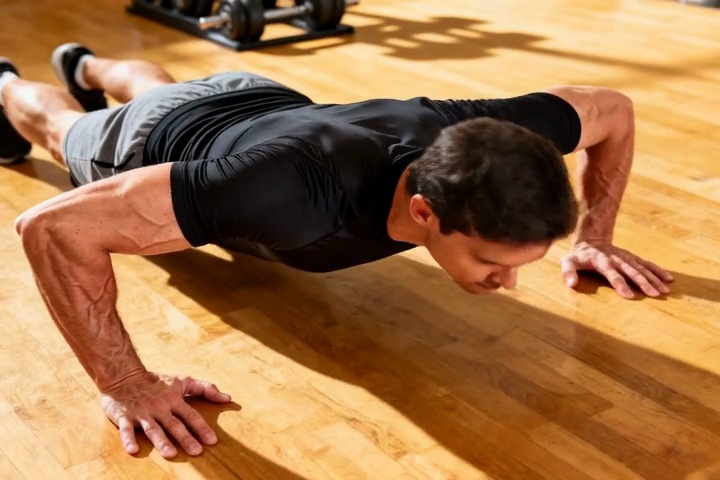} &
    \includegraphics[width=\imW]{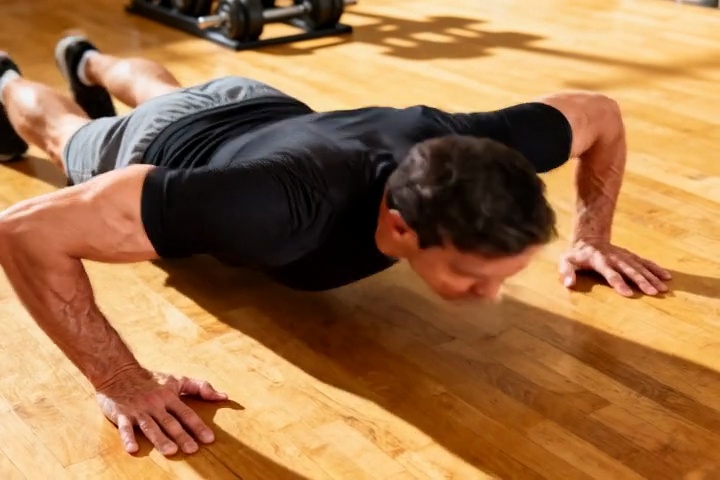} \\[5pt]

    \includegraphics[width=\imW]{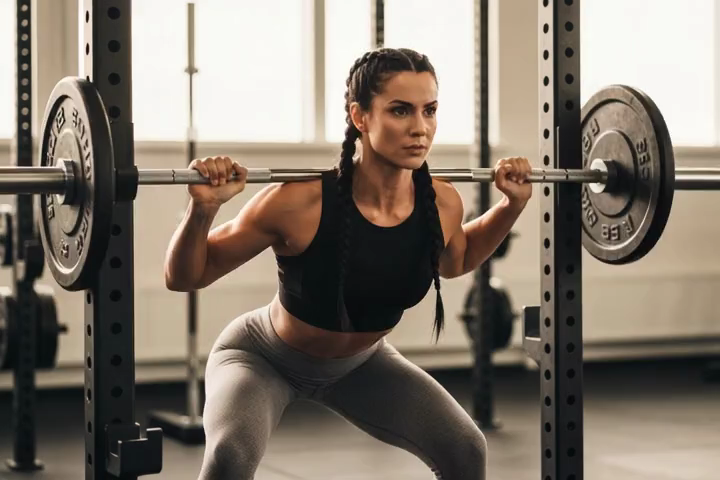} &
    \includegraphics[width=\imW]{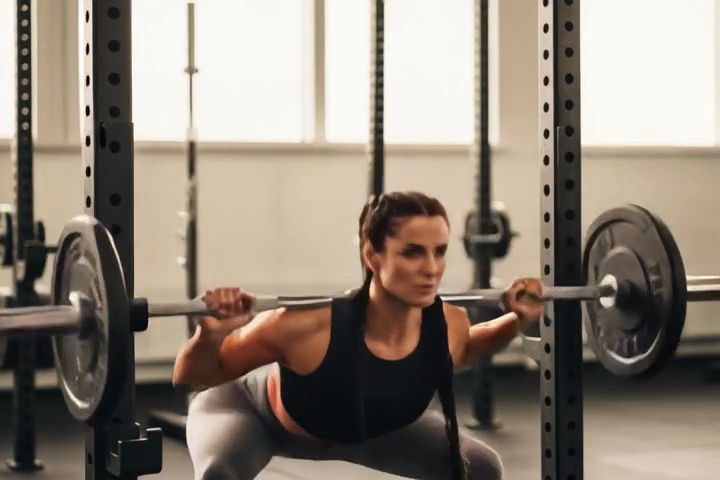} &
    \includegraphics[width=\imW]{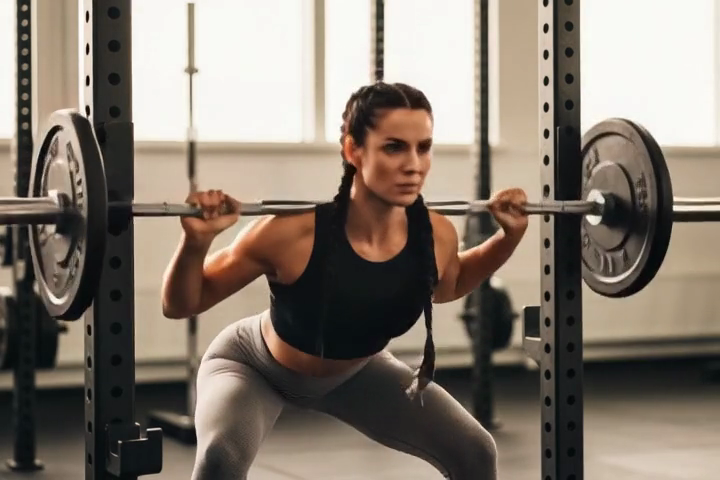} &
    \includegraphics[width=\imW]{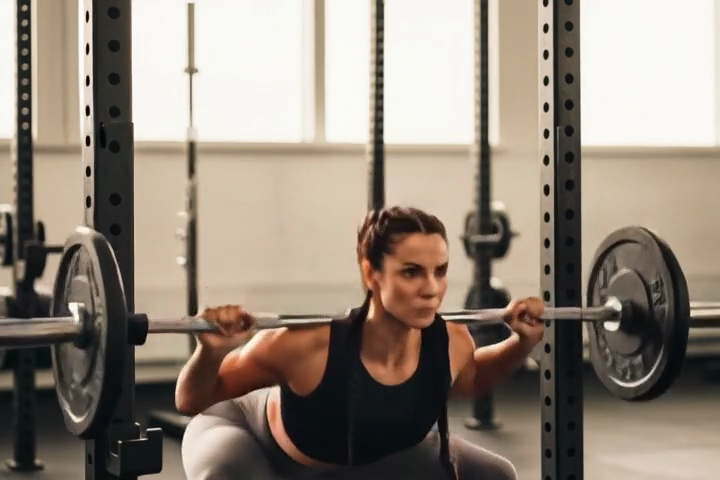} &
    \includegraphics[width=\imW]{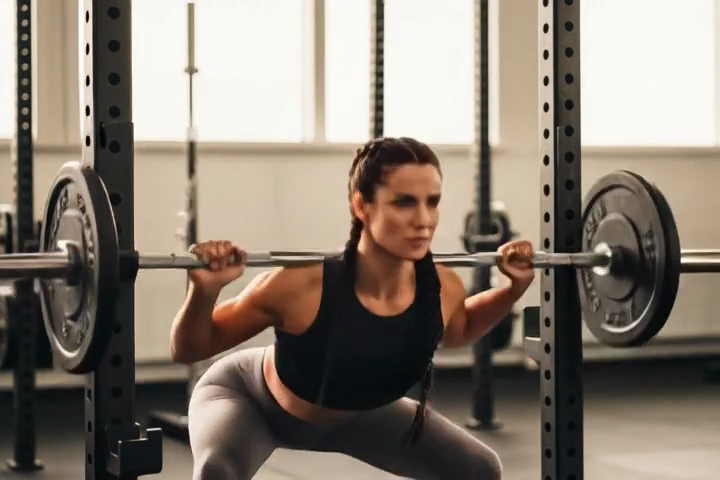} \\[5pt]

    \includegraphics[width=\imW]{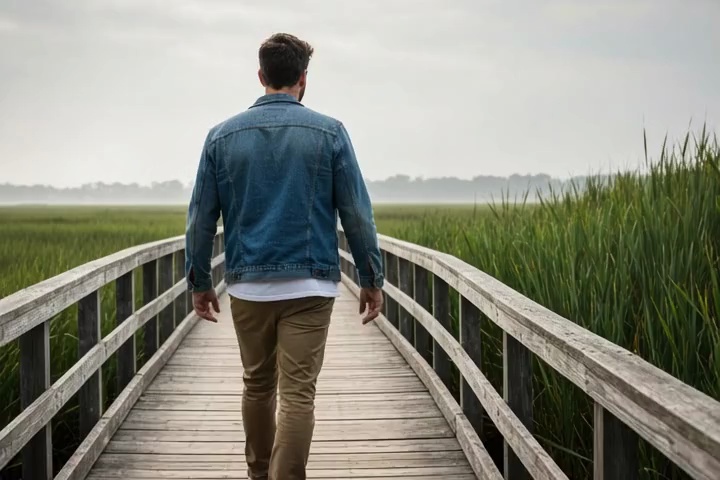} &
    \includegraphics[width=\imW]{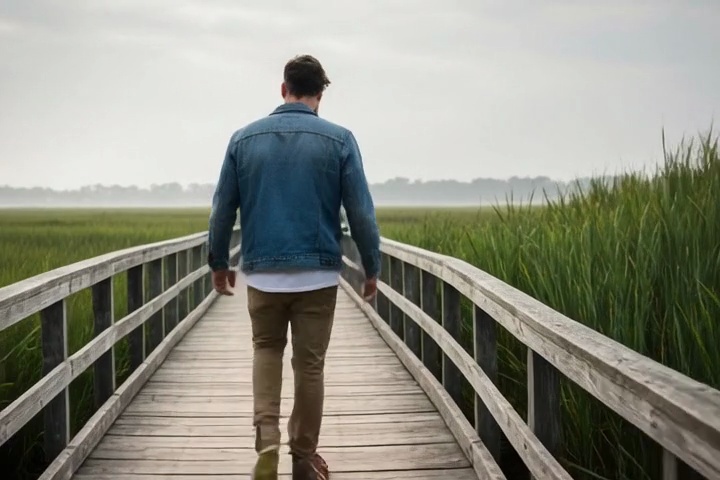} &
    \includegraphics[width=\imW]{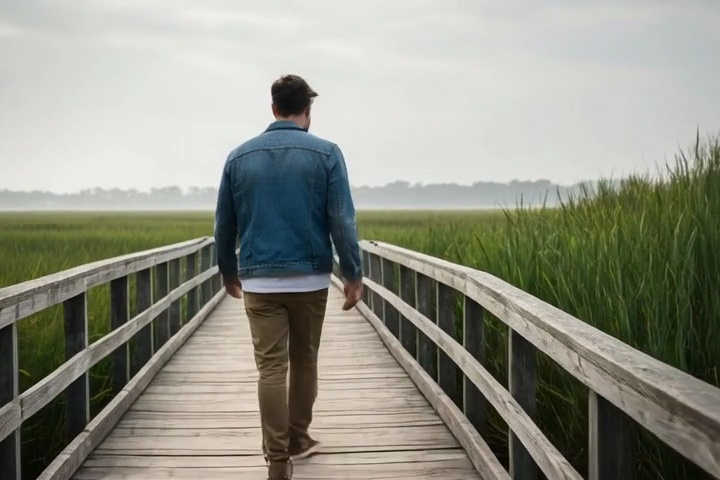} &
    \includegraphics[width=\imW]{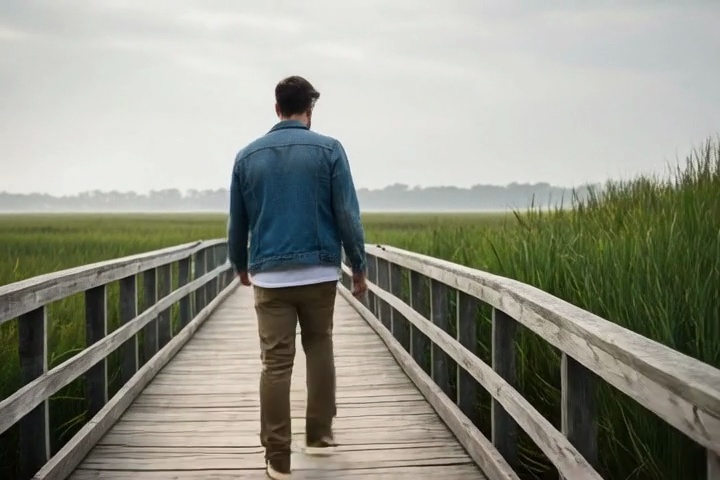} &
    \includegraphics[width=\imW]{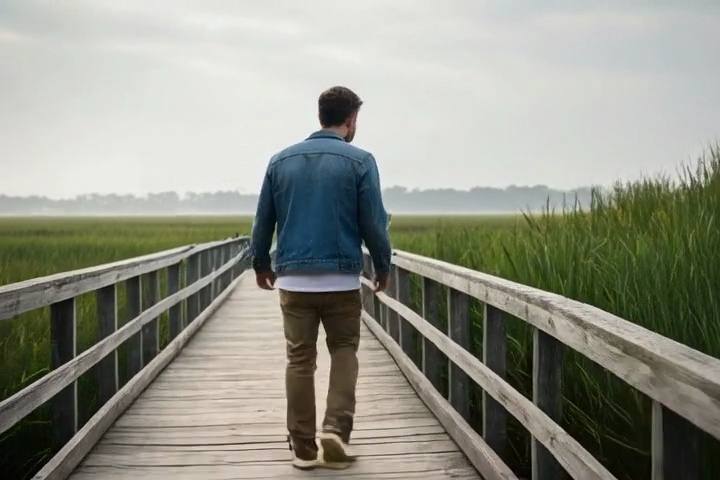} \\[5pt]

    \includegraphics[width=\imW]{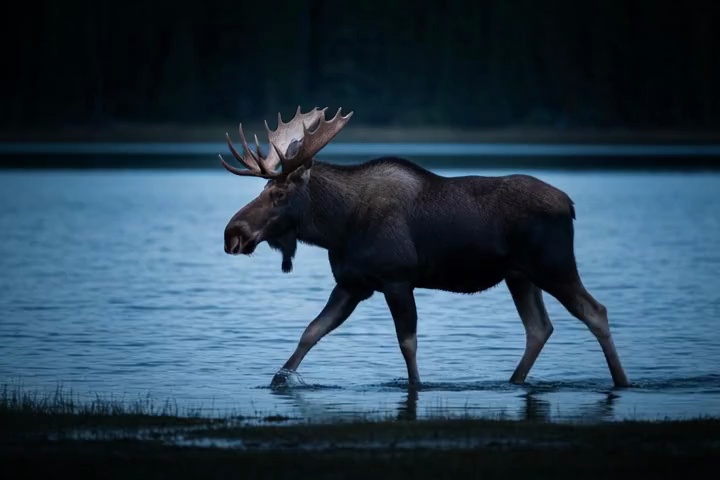} &
    \includegraphics[width=\imW]{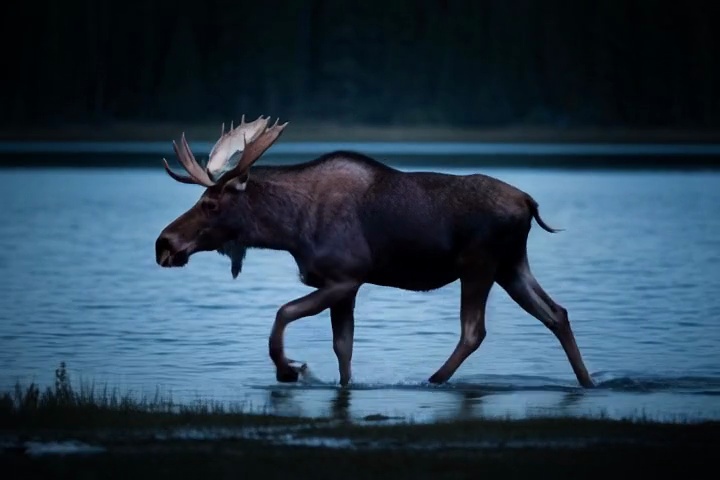} &
    \includegraphics[width=\imW]{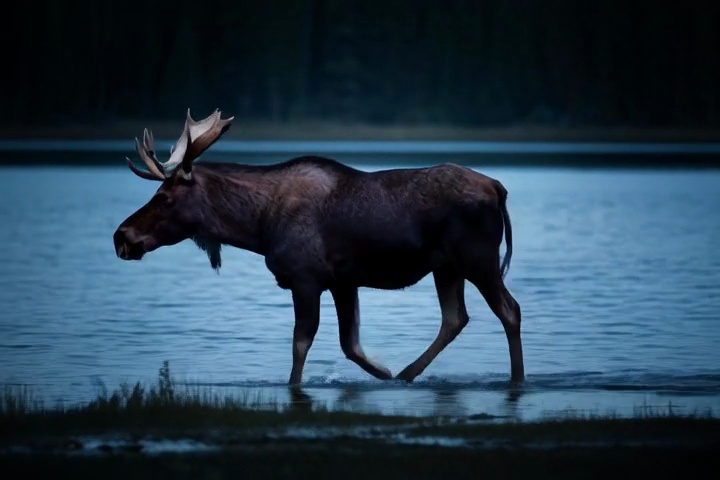} &
    \includegraphics[width=\imW]{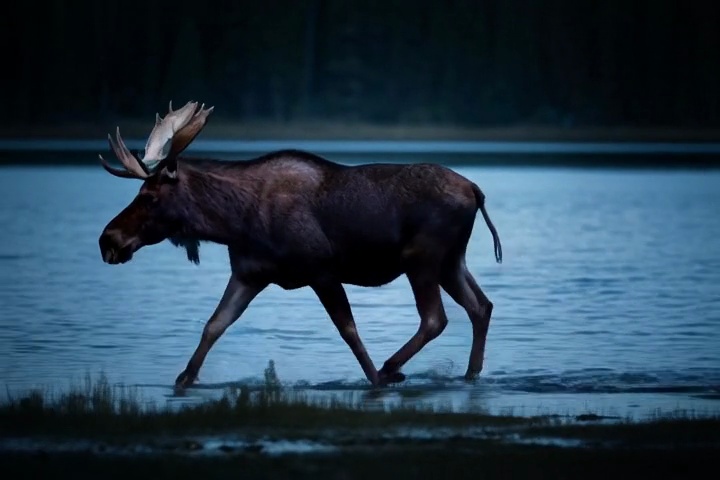} &
    \includegraphics[width=\imW]{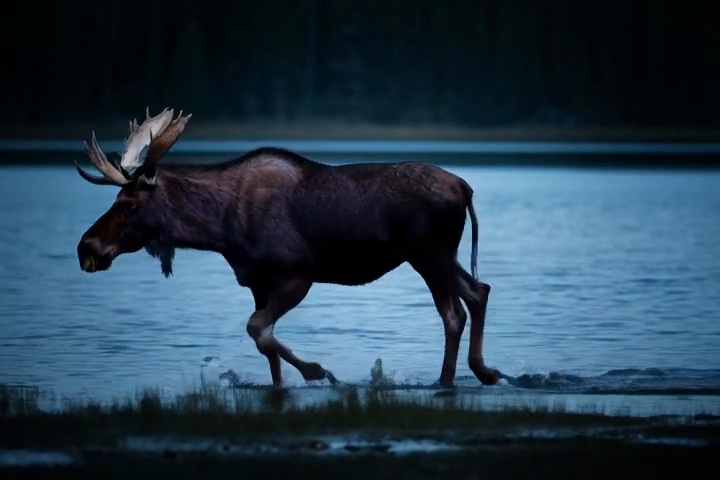} \\[5pt]

    \includegraphics[width=\imW]{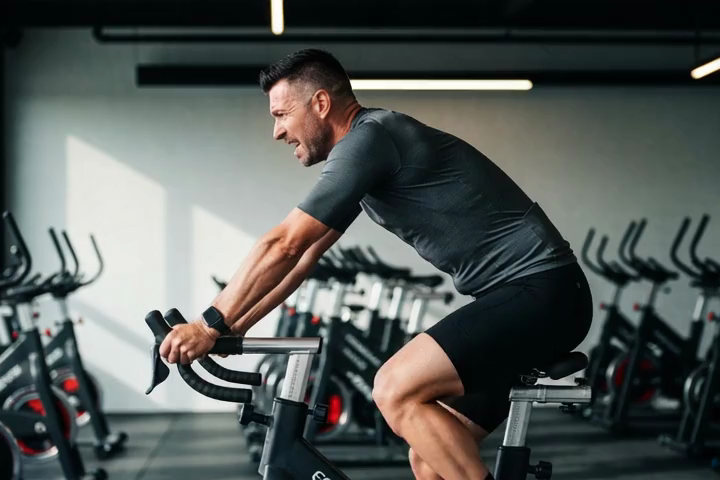} &
    \includegraphics[width=\imW]{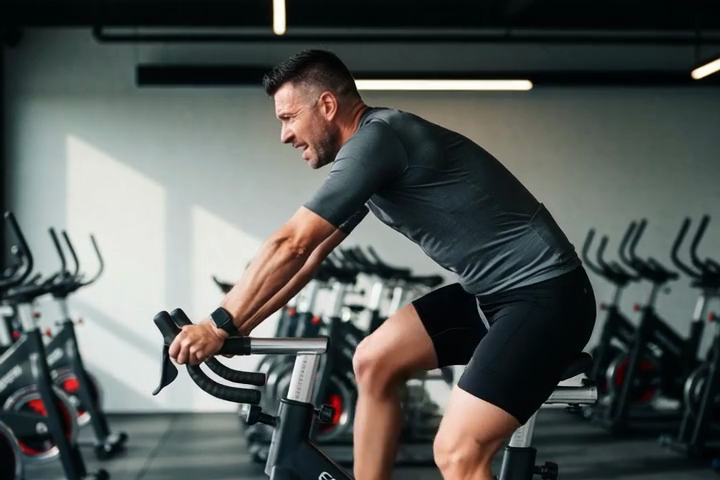} &
    \includegraphics[width=\imW]{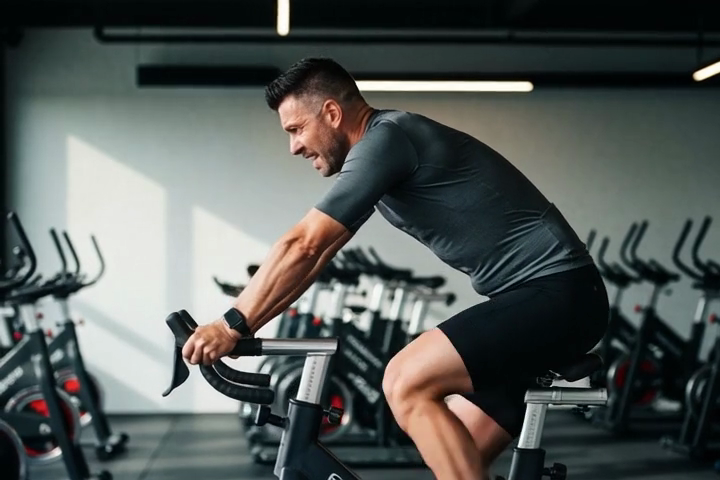} &
    \includegraphics[width=\imW]{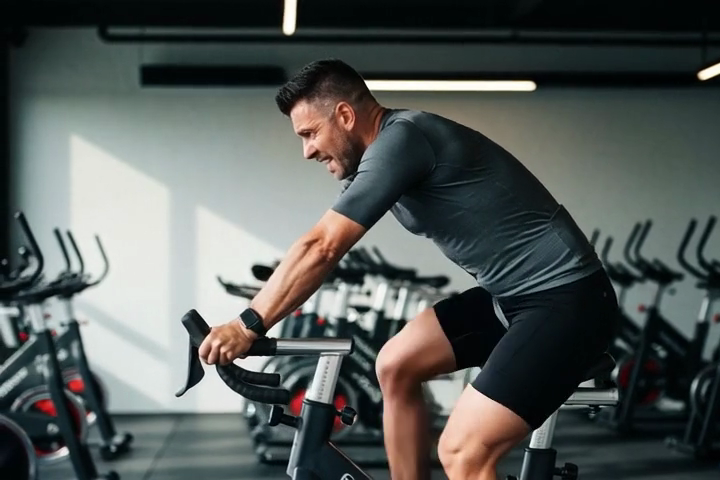} &
    \includegraphics[width=\imW]{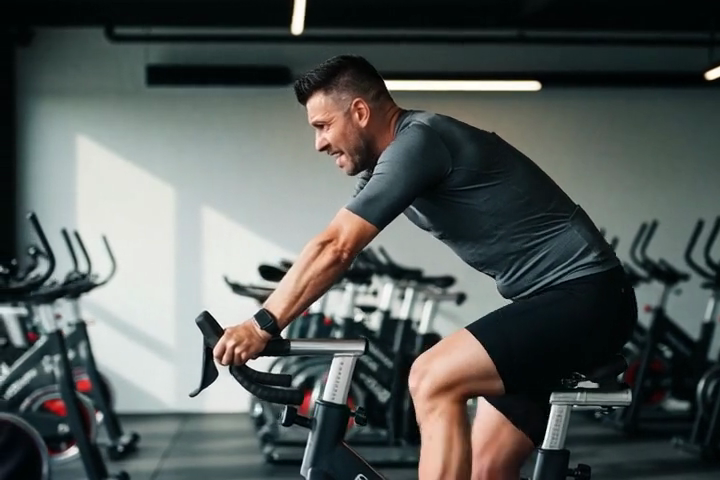} \\[5pt]

    \includegraphics[width=\imW]{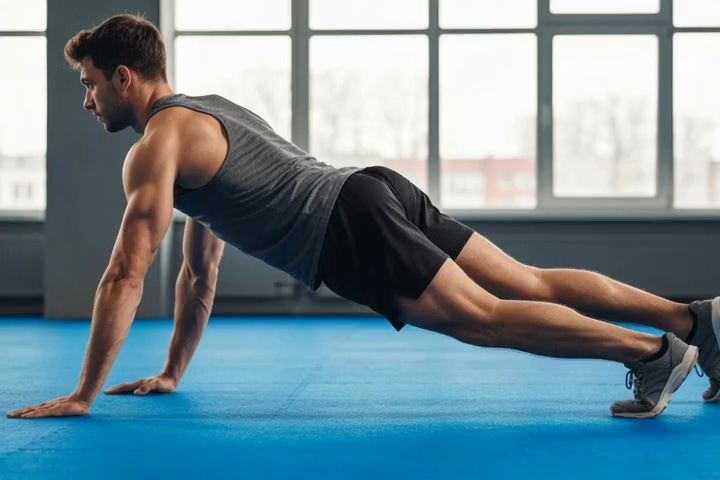} &
    \includegraphics[width=\imW]{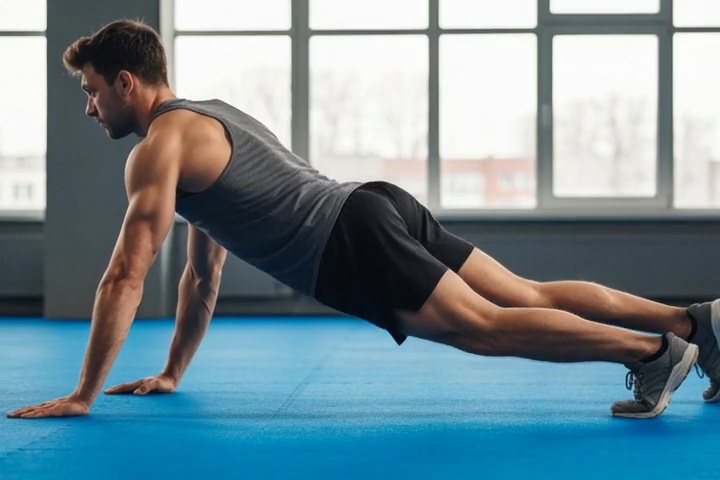} &
    \includegraphics[width=\imW]{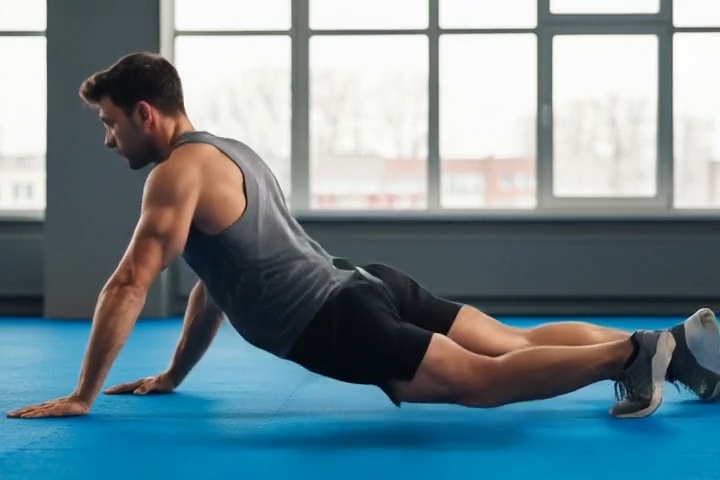} &
    \includegraphics[width=\imW]{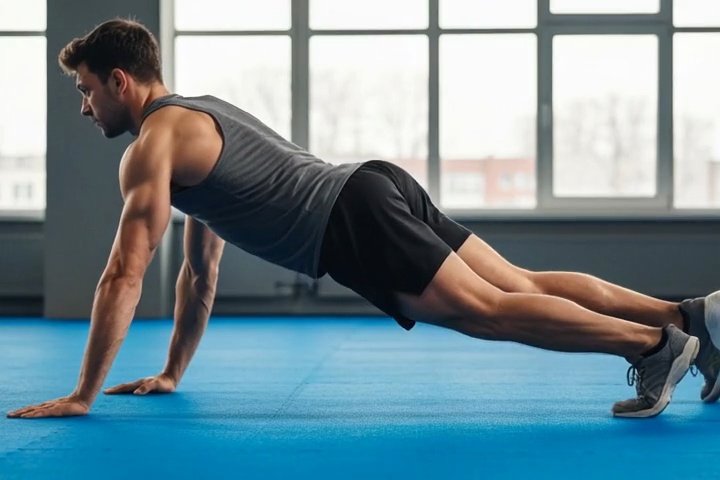} &
    \includegraphics[width=\imW]{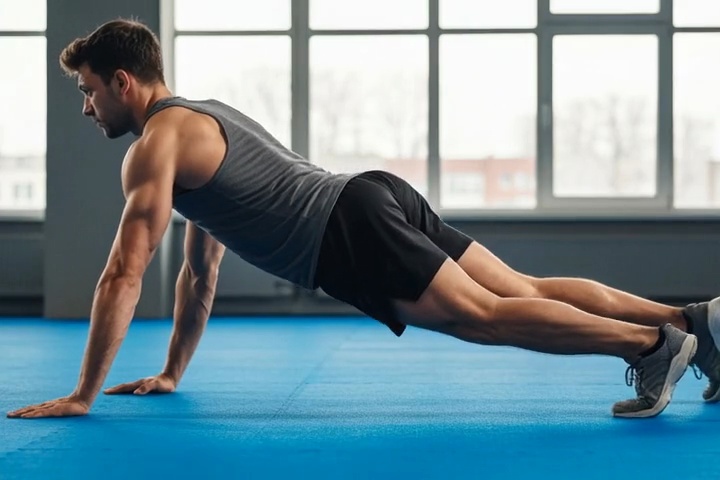} \\

\end{tabular}
\caption{\textbf{Additional Qualitative Results.} Our \model{} maintains robust structural consistency across a diverse range of motion dynamics.}
\end{figure*}

\section {Theoretical Analysis of Feature Fusion}

\noindent\textbf{Definition of $\softmax(\cdot)$.}
As referenced in Sec.~\ref{sec:loss}, $\softmax(\cdot)$ denotes a temperature-scaled softmax operation applied to each similarity vector, normalizing the $7 \times 7$ similarity scores into a probability distribution.
Given similarities $\{z_i\}_{i=1}^{K}$, we compute
\[
p_i = \frac{\exp\left(z_i / T\right)}
           {\sum_{j=1}^{K} \exp\left(z_j / T\right)},
\]
and in all experiments we set the temperature to $T = 0.1$.

\vspace{0.2cm}

\noindent \textbf{Interference in Feature-Space Fusion.} Let $a = F_\mathrm{mem,t}^{\mathrm{fwd}}$, $b = F_\mathrm{mem,t}^{\mathrm{bwd}}$, $c = F_\mathrm{mem,t+1}^{\mathrm{fwd}}$, and $d = F_\mathrm{mem,t+1}^{\mathrm{bwd}}$ represent directional SAM2 memory features across two consecutive frames.
The ideal local Gram similarities for the forward and backward teachers are $a \cdot c$ and $b \cdot d$, respectively.
If we first fuse forward and backward features in the feature space,
\[
f_t = k a + (1-k)b, \quad
f_{t+1} = k c + (1-k)d ,
\]
and then compute the local Gram, we obtain
\[
g_{\text{feat}} = f_t \cdot f_{t+1}
= k^2 (a \cdot c)
+ (1-k)^2 (b \cdot d)
+ k(1-k)\big(a \cdot d + b \cdot c\big).
\]
The final term introduces cross-correlations ($a \cdot d$ and $b \cdot c$) that couple forward features at frame $t$ with backward features at frame $t{+}1$.
These cross terms lack a counterpart in the valid teacher similarity matrix (neither purely forward nor backward), thereby mixing incompatible temporal contexts.
Consequently, the student model is supervised to learn spurious correlations absent in the teacher's distribution, leading to temporal inconsistencies.

In contrast, our LGF fusion computes local Gram similarities for each direction separately and fuses them only in the LGF space:
\[
g_{\text{lgf}} = k (a \cdot c) + (1-k)(b \cdot d).
\]
This formulation eliminates cross terms, yielding a convex combination of two consistent teacher signals and circumventing the interference inherent in feature-space fusion.

\section{Limitations and Future Work}

\noindent \textbf{Limitations.} We acknowledge that our model's performance is effectively bounded by the inherent capabilities of the underlying backbone, CogVideoX. specifically, in scenarios involving high-dynamic or complex motion, such as fast-paced dancing or competitive sports, we observe that generation artifacts may remain pronounced. This suggests that while our method significantly improves structural alignment, the ultimate video quality in extreme motion cases relies on the generative capacity of the base model.

\vspace{0.2cm}

\noindent \textbf{Future Work.} Our current pipeline relies on SAM 2 for video object segmentation and tracking. While the model demonstrates robust performance for single objects, we observe performance degradation in multi-object scenarios. Specifically, when prompted with visual cues (e.g., bounding boxes or points) for multiple distinct entities, the tracker struggles to maintain consistent identities across temporal sequences. Consequently, exploring effective feature representations for multi-object video generation remains an open and promising direction for future research.